\documentclass[twocolumn]{article}
\usepackage{fullpage}

\usepackage[utf8]{inputenc} % allow utf-8 input
\usepackage[T1]{fontenc}    % use 8-bit T1 fonts

\usepackage{authblk}

\usepackage{graphicx}
\usepackage{mathptmx}
\usepackage[unicode,colorlinks,citecolor={blue},filecolor={blue},urlcolor={blue},anchorcolor={blue}, linkcolor={blue}, breaklinks]{hyperref}       % hyperlinks
\usepackage{url}            % simple URL typesetting
\usepackage{booktabs}       % professional-quality tables
\usepackage{amsfonts}       % blackboard math symbols
\usepackage{bbm} % For bb 1
\usepackage{nicefrac}       % compact symbols for 1/2, etc.
\usepackage{microtype}      % microtypography
\usepackage{graphicx}
\usepackage{subcaption} % For subfigures and subcaptions
\usepackage{multirow}
\usepackage{multicol}
\usepackage{tablefootnote}
\usepackage{amsmath}
\usepackage{lipsum}
\usepackage{xcolor}
\usepackage{listings}
\usepackage{enumerate}

\newcommand{\best}[1]{\textbf{#1}}

\definecolor{brca-normal}{rgb}{0.2980392156862745, 0.4470588235294118, 0.6901960784313725}
\definecolor{brca-luma}{rgb}{0.867, 0.518, 0.322}
\definecolor{brca-lumb}{rgb}{0.333, 0.659, 0.408}
\definecolor{brca-basal}{rgb}{0.769, 0.306, 0.322}
\definecolor{brca-her2}{rgb}{0.506, 0.447, 0.702}

\newcommand{\GBKS}{GLYCOSAMINOGLYCAN\_BIOSYNTHESIS\_KERATAN\_SULFATE}
\newcommand{\SM}{SPHINGOLOPID\_METABOLISM}
\newcommand{\VLID}{VALINE\_LEUCINE\_AND\_ISOLEUCINE\_DEGRADATION}
\newcommand{\PSP}{P53\_SIGNALING\_PATHWAY}
\newcommand{\PC}{PANCREATIC\_CANCER}

\date{}

\title{Incorporating Prior Knowledge in Deep Learning Models via Pathway Activity Autoencoders}

\author[1,2,*,id]{\href{https://orcid.org/0000-0002-0347-7002}{Pedro~Henrique da~Costa~Avelar}}
\author[2,*,id]{\href{https://orcid.org/0000-0003-0977-3600}{Min Wu}}
\author[1,*,id]{\href{https://orcid.org/0000-0001-8403-1282}{Sophia Tsoka}}

\affil[1]{Department of Informatics, Faculty of Natural, Mathematical and Engineering Sciences, King’s College London, United Kingdom}
\affil[2]{Institute for Infocomm Research, A*STAR, Singapore}
\affil[$\ast$]{Emails \{pedro\_henrique.da\_costa\_avelar,sophia.tsoka\}@kcl.ac.uk, wumin@i2r.a-star.edu.sg}
\affil[id]{Author's name is a link to author's Orcid profile}

\begin{document}

\twocolumn[
\begin{@twocolumnfalse}
\maketitle
\begin{abstract}
\textbf{Motivation:} Despite advances in the computational analysis of high-throughput molecular profiling assays (e.g. transcriptomics), a dichotomy exists between methods that are simple and interpretable, and ones that are complex but with lower degree of interpretability. Furthermore, very few methods deal with trying to translate interpretability in biologically relevant terms, such as known pathway cascades.
Biological pathways reflecting signalling events or metabolic conversions are Small improvements or modifications of existing algorithms will generally not be suitable, unless novel biological results have been predicted and verified. Determining which pathways are implicated in disease and incorporating such pathway data as prior knowledge may enhance predictive modelling and personalised strategies for diagnosis, treatment and prevention of disease. 
\\\textbf{Results:} We propose a novel prior-knowledge-based deep auto-encoding framework, PAAE, together with its accompanying generative variant, PAVAE, for RNA-seq data in cancer.
Through comprehensive comparisons among various learning models, we show that, despite having access to a smaller set of features, our PAAE and PAVAE models achieve better out-of-set reconstruction results compared to common methodologies.
Furthermore, we compare our model with equivalent baselines on a classification task and show that they achieve better results than models which have access to the full input gene set.
Another result is that using vanilla variational frameworks might negatively impact both reconstruction outputs as well as classification performance.
Finally, our work directly contributes by providing comprehensive interpretability analyses on our models on top of improving prognostication for translational medicine. 
\\\textbf{Availability and implementation:}
    The source code will be publicly available at \url{https://github.com/phcavelar/pathwayae}. All of the data used on this work is publicly available and was acquired through XenaBrowser \url{https://xenabrowser.net/datapages/}.
\end{abstract}
\end{@twocolumnfalse}
]

\section{Introduction}

Incorporating medical or biological prior knowledge into predictive methods is important in developing models that are interpretable, and have configurable priors that are clinically relevant. This is especially important in the context of translational medicine, where data can be expensive to acquire and therefore scarce, a factor that limits the application of many machine learning algorithms \cite{von_rueden_informed_2021}. Incorporating biological priors in the form of biochemical pathways was shown to improve disease classification \cite{yang_pathway_2014}, with Knowledge Graphs have also been used as source of prior knowledge in various models \cite{lin_kgnn_2020,long_pre-training_2022}. 

Autoencoders (AE), originally proposed as a nonlinear generalisation of Principal Component Analysis \cite{hinton_reducing_2006}, are a family of deep learning architectures that focus on learning representations of inputs through mostly unsupervised methods, and have many variations which generally address some specific relational inductive bias required by the problem at hand. Key examples include (i) Variational Autoencoders (VAE) \cite{kingma_auto-encoding_2014} to enable generative modelling and a more interpretable latent representation; (ii) $\beta$-VAEs \cite{higgins_beta-vae_2017} which allow for disentangled factor learning on top of the traditional VAE; (iii) Concrete Autoencoders \cite{balin_concrete_2019} which allow end-to-end learnable feature selection, making the latent variables more interpretable among others. This family of deep learning architectures have recently been used to interpret survival-based bulk multi-omics cancer datasets, with two initial papers reporting the use of both standard Autoencoders \cite{chaudhary_deep_2018} as well as a Variational Autoencoders \cite{ronen_evaluation_2019}.

Specific pieces of work \cite{chaudhary_deep_2018,ronen_evaluation_2019} have been seminal and based their work on traditional Autoencoders, using them to learn latent joint representations of multi-omics features, which are then input to a traditional survival analysis pipeline to identify clusters of patients and explore how well these latent features separate patients into risk groups \cite{chaudhary_deep_2018,zhang_deep_2018,asada_uncovering_2020,lee_incorporating_2020,poirion_deepprog_2021}. The other main body dealing with unsupervised Autoencoders has as its most salient difference the use of Variational Autoencoders \cite{ronen_evaluation_2019,uyar_multi-omics_2021}, which help constrain the learned embeddings and allow for a more thorough exploration of the latent space \cite{higgins_beta-vae_2017} as well as paving the way for generative modelling \cite{kingma_auto-encoding_2014}, which was not used in any of the above-mentioned works.

Recent work has addressed learning representations for survival-based multi-omics cancer data in bulk tissue and in a supervised fashion \cite{tong_deep_2020,wissel_hierarchical_2022}, building on previous work on deep learning-based survival prediction methods \cite{katzman_deepsurv_2018,ching_cox-nnet_2018,huang_salmon_2019}. In \cite{tong_deep_2020} late-integration (i.e. analysing each omics layer separately) is reported, whereas \cite{wissel_hierarchical_2022} builds their analyses  on top of the late-integration method in a hierachical fashion, generating a two-step autoencoding architecture that provides both single-omics as well as multi-omics latent representations which are built to be relevant for survival.

Most of the above-mentioned work is either purely unsupervised or supervised on small cohorts and with low sample-to-feature ratios, with the number of samples as low as a few hundred and the number of features almost as high as one hundred thousand.  Given that most deep learning algorithms work best when there is a wealth of training data, low sample-to-feature ration might cause overfitting or insuficient generalisation. One possible way forward would be to apply informed machine learning \cite{von_rueden_informed_2021}, constraining the possible hypothesis spaces for the model and facilitating meaningful learning on the model. A wealth of research papers exist that attempt to apply some form of prior knowledge constraint in omics data analysis. Examples are applying latent-space constraints that enforce the feature space to approximate a tree hierarchical structure \cite{garrido_visualizing_2022}, as well as a latent space that enforces invariance to extrinsic confounding factors such as batch effects \cite{cao_sailer_2021}, building multiple latent representations to represent different pathways \cite{gut_pmvae_2021}, soft-enforcing in pathway sets and/or sparse and dense factors onto a linear decoder inside an Autoencoder \cite{rybakov_learning_2020}, as well as using pathway information to build multi-omics Autoencoders \cite{lemsara_pathme_2020}.

Here, we report two prior-knowledge based frameworks that use pathway information to constrain Autoencoder models into learning latent representations. This is done by building data-driven nonlinear pathway activities scores, which reduces the number of parameters needed with respect to a dense Autoencoder and provides a different feature space. As this space is based on prior knowledge in the form of interactions encoded in biochemical pathways, such space may be informative to clinical application. Our work differs from \cite{gut_pmvae_2021} in that we learn a single end-to-end model that jointly optimises pathways while still being constrained into producing one pathway activity score for each pathway, which is then used to build a richer latent space, instead of producing one multidimensional latent space per pathway. Furthermore, our work differs from \cite{lemsara_pathme_2020} in that our Autoencoder framework jointly optimises all the pathway activity scores, instead of training a different feature set per pathway, as well as using this space directly to produce a latent space that is decodable into the entire gene set, instead of only decoding the genes that are present in each pathway. We also show that our proposed method is more efficient than dense Autoencoders, which have access to more information and are commonly used in omics analyses \cite{chaudhary_deep_2018,zhang_deep_2018,asada_uncovering_2020,lee_incorporating_2020,poirion_deepprog_2021,ronen_evaluation_2019,uyar_multi-omics_2021}, in building a representation that is good at representing the input of unseen samples.

Our main contributions with this work are:
\begin{enumerate}
    \item Developing a framework to incorporate pathway prior knowledge into Autoencoder models, including Variational Autoencoder;
    \item Showing that such models can perform better despite using less parameters and less nonlinear layers, than traditional Autoencoder models with external validation on a classification task;
    \item Demonstrating ways to interpret the internal values of our Pathway Activity models, showing that as they provide values that are associated with Pathways, they are easier to interpret.
\end{enumerate}

\section{Materials and Methods}

This section provides details of data and two proposed Pathway Activity Autoecoder (PAAE) and Pathway Activity Variational Autoencoder (PAVAE) models.

\subsection{Datasets and Preprocessing}

GDC TCGA gene expression RNAseq HTSeq - FPKM dataset for Breast Cancer (BRCA) \cite{the_cancer_genome_atlas_network_comprehensive_2012} were acquired from xenabrowser (\url{https://xenabrowser.net/datapages/}), and the mRNA expression (Illumina HT-12 v3 microarray) Metabric dataset \cite{metabric_group_genomic_2012} acquired from cBioPortal (\url{https://www.cbioportal.org/}). The TCGA and Metabric datasets have $n_{BRCA}=1217$ and $n_{Metabric}=1756$ samples in total, respectively, and contains $60483$, and $20592$ measurements per sample, respectively, which were mapped from Ensembl gene IDs to gene names, from which those that appear on both datasets were taken. Merging repeated Gene IDs with the average of the values resulted in datasets with $d_{x}=56867$ features per sample. Whenever we take the average between multiple symbols, we first map from their original $log2+1$ values to a non-log space before taking the average and then back to their original log representation. Clinical data were collected from xenabrowser and cBioPortal (i.e.GDC TCGA BRCA ``Phenotypes'' and the metabric ``Clinical Data''). PAM50 subtype \cite{the_cancer_genome_atlas_network_comprehensive_2012,parker_supervised_2009} information contained in the \texttt{PAM50Call\_RNAseq} for the TCGA dataset and the \texttt{CLAUDIN\_SUBTYPE} column for the Metabric dataset, from which we drop empty values as well as samples with the \texttt{claudin-low} and \texttt{NC} values. Pathway sets were obtained from GSEA Human MSigDB Collections (\url{https://www.gsea-msigdb.org/}), where \texttt{c2.cp.kegg.v7.5.1} contains 186 pathways (referred to to as ``KEGG'')  and \texttt{h.all.v7.5.1.json} with 50 pathways (``Hallmark Genes'' or ``HG''). A full description of datasets and pathway sets can be found in the Supplementary material.

\subsection{Problem Statement}

Autoencoder models are often used as a dimensionality reduction method before using the factors for further analysis \cite{chaudhary_deep_2018,zhang_deep_2018,asada_uncovering_2020,lee_incorporating_2020,poirion_deepprog_2021,ronen_evaluation_2019,uyar_multi-omics_2021}, therefore it is of high importance that the latent vectors produced by the models not only contain enough information about the input so as to rebuild the input models, but also that the representations learned by Autoencoders contain meaningful information about features of interest. Such features have been used before for prognostication of survival chances \cite{chaudhary_deep_2018} as well as identifying predefined subtypes \cite{ronen_evaluation_2019,uyar_multi-omics_2021}. In this paper we focus on cancer subtype classification using the PAM50 subtypes for Breast Cancer.

To perform the classification tests: (1) unsupervised Autoencoder models were trained; (2) these models were used to compress the input to their latent representation; (3) the compressed latent representation was used as input to a classifier. For the AE and PAAE models we use $z$ as the latent representation, and the posterior means $\mu$ for the VAE and PAVAE models. Three commonly-used classifiers were used: Logistic Regression (LR), Support Vector Machine (SVM) classifier, and a Random Forest (RF) classifier.

\subsection{Our PAAE model}

\begin{figure}
    \centering
    \subcaptionbox{\label{fig:diagrams-ae:sub:ae}}{\includegraphics[scale=.94]{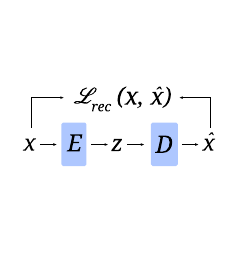}}
    \subcaptionbox[scale=.94]{\label{fig:diagrams-ae:sub:paae}}{\includegraphics{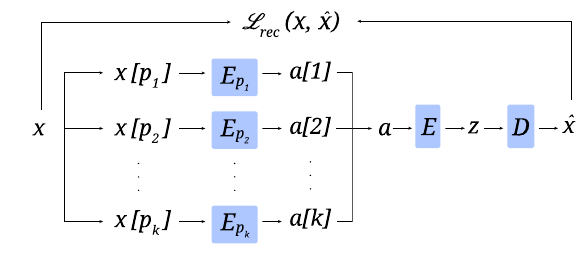}}
    \caption{A diagrammatic representation of a (\subref{fig:diagrams-ae:sub:ae}) traditional Autoencoder (AE) and a (\subref{fig:diagrams-ae:sub:paae}) Pathway Activity Autoencoder (PAAE). Note that the union of the inputs available to the PAAE $\bigcup_{p_{j} \in P}{p_{j}}$ is often smaller than the entire gene set that the model has to reconstruct at $\hat{x}$, thus being more constrained and having less information available than the traditional AE.}
    \label{fig:diagram-ae}
\end{figure}

Here we propose a new framework (Pathway Activity Autoencoder PAAE), where instead of building an Autoencoder through a dense encoder and decoder, we use prior knowledge in the form of biochemical pathways to build \emph{pathway activity encoders} $E_{p_{j}}$ for each pathway $p_{j}$ inside a predetermined set of pathways $P$, where each pathway is comprised of a set of genes $\{g_1, \ldots, g_{|p_{j}|}\}$. Each of these pathway encoders is also a $|p_{j}| \rightarrow 1$ learnable function that maps components of that pathway into a single score, which we then concatenate into a pathway activity vector $a$, as in Eq. (\ref{eq:paae}). 

\begin{align}
\footnotesize
    a_{j} &= E_{p_{j}}(x_{:,p_{j}}) \label{eq:paae-single-activity} \\
    a &= \parallel_{p_{j} \in P} a_{j} \label{eq:paae}
\end{align}

This pathway activity vector is then further compressed through a final encoder $E$ into an encoding $z = E(a)$, which is finally used to reconstruct the input through $\hat{x} = D(z)$. Visualisation of this architecture and comparison with a traditional Autoencoder is shown in Fig.~\ref{fig:diagram-ae}.

Finally, the model is trained to minimise a reconstruction loss $\mathcal{L}_{rec}$, which commonly takes the form of either the Mean Squared Error (MSE) between the input $x$ and its reconstruction $\hat{x}$, or the Binary Cross Entropy, in the case that the inputs represent binary classes or values scaled between 0 and 1. Since we are dealing with data that is not necessarily 0-1 normalised, the final loss, using MSE, would be as in Eqs. (\ref{eq:mse-ae} and \ref{eq:mse-ae-de}). 

\begin{align}
\footnotesize
    \mathcal{L}_{rec} &= \frac{1}{d} \sum_{i=1}^{d} (x_{:,i}-\hat{x}_{:,i})^2 \label{eq:mse-ae} \\
     &= \frac{1}{d} \sum_{i=1}^{d} (x_{:,i}-D(E(x))_{:,i})^2 .\label{eq:mse-ae-de}
\end{align}

It is noted that this novel, prior-knowledge-based construction of an Autoencoder allows us to have both a pathway-centric data-driven set of features (considered to be highly interpretable as they directly encode meaning through its prior-knowledge constraints), as well as another more compressed representation that would be more similar to a traditional Autoencoder.

\subsection{PAVAE -- Pathway-Constrained Generative Modelling} \label{ssec:vae}

\begin{figure*}
    \centering
    \hfill \subcaptionbox{\label{fig:diagrams-vae:sub:vae}}{\includegraphics{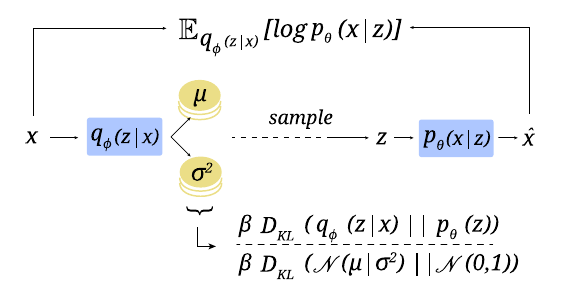}}
    \hfill \subcaptionbox{\label{fig:diagrams-vae:sub:pavae}}{\includegraphics{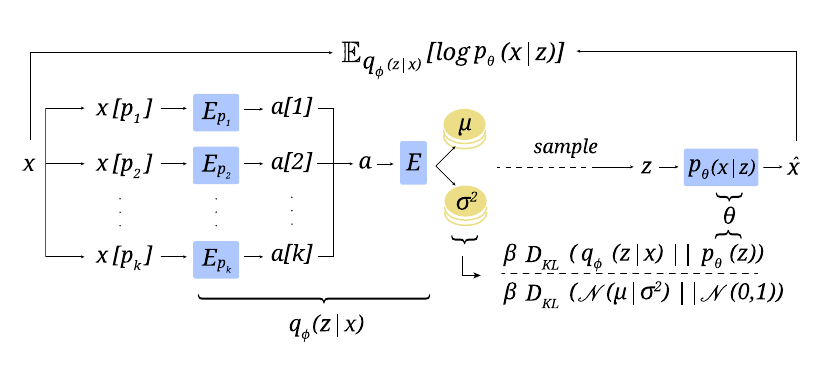}} \hfill
    \caption{A diagrammatic representation of a (\subref{fig:diagrams-vae:sub:vae}) Variational Autoencoder (VAE) and (\subref{fig:diagrams-vae:sub:pavae}) a Pathway Activity Variational Autoencoder (PAVAE). Note that apart from what is mentioned in Fig.~\ref{fig:diagram-ae} about the Pathway Activity construction being more constrained and having less information than the traditional construction, the PAVAE enables $q_{\phi}(z|x)$ to be composed of two different parts, i.e. the pathway activity encoders which build up the pathway activity vector $a$, and the variational encoder $q_{\phi}(z|a)$ which will encode a posterior density $p_{\theta}(z|a)$ that is based on pathway activities observed in the sample.}
    \label{fig:diagrams-vae}
\end{figure*}

Traditional Autoencoders are known to be insufficient for learning meaningful representations, usually requiring some sort of regularization technique \cite{bengio_representation_2013,kingma_auto-encoding_2014}. To address this issue we also extend our PAAE to the Variational Autoencoder (VAE) framework, as shown in Fig.~\ref{fig:diagrams-vae}. This model includes a variational bound, which acts in lieu of a regularization term \cite{kingma_auto-encoding_2014}. This is achieved through learning an encoder that outputs a probability distribution, from which values are then sampled and decoded into the final reconstructed input. In more precise terms, considering our input dataset $x$, composed of $n$ samples and $d$ gene expression features, we assume that this data is generated by some random process, involving an unobserved continuous random variable $z$, which we assume is generated from a prior distribution $p_{\theta\star}(z)$, and that values from $x$ are generated by a \emph{conditional} distribution $p_{\theta\star}(x|z)$, and we assume that the prior and the likelihood respectively come from a parametric family of distributions $p_{\theta}(z)$ and  $p_{\theta}(x|z)$.

To navigate the intractability of finding the true marginal likelihood $p_{\theta}(x)$ and the true posterior density $p_{\theta}(z|x)$ through normal means (e.g., through the Expectation Maximization algorithm), which commonly occurs when the likelihood $p_{\theta}(x|z)$ is modelled by a somewhat complex likelihood function (e.g., a neural network), we introduce a model $q_{\phi}(z|x)$ that approximates the true posterior density $p_{\theta}(z|x)$. We further constrain the model $q_{\phi}(z|x)$ to be composed of two corresponding stages: first a set of pathway activity encoders that produce the learned pathway activity vector $a = E_{p_{j}}(x)$ that is then used as an input to another module that computes $q_{\phi}(z|a)$. This constraint allows the learned generative process posterior to be dependent on pathway activities $p_{\theta}(z|a)$. With this \emph{encoder} $q_{\phi}(z|x) = q_{\phi}(z|a)$, we can produce a distribution given a datapoint $x$ from which one can sample \emph{codes} $z$ (i.e. latent representations), from which $x$ could have been generated through the probabilistic decoding process of the \emph{decoder} $p_{\theta}(x|z)$. This process can be visualised in Fig.~\ref{fig:diagrams-vae:sub:pavae} and compared with the traditional VAE shown in Fig.~\ref{fig:diagrams-vae:sub:vae}.

Given these components, our optimisation objective is to jointly minimise an Estimated Lower Bound (ELBO) loss w.r.t. $\theta$ and $\phi$, $\mathcal{L}(\theta,\phi,x) = \mathbb{E}_{q_{\phi}(z|x)}[log p_{\theta}(x|z)] - D_{KL}(q_{\phi}(z|x) || p_{\theta}(z)) +$, where $D_{KL}$ is the KL divergence between the two distributions. This formulation has also been expanded to learn disentangled representations of individual generative factors by having a constraint that $D_{KL}(q_{\phi}(z|x) || p_{\theta}(z)) < \varepsilon$, which ultimately leads to the altered loss shown in Eq. (\ref{eq:bvae}).

\begin{equation} \label{eq:bvae}
\footnotesize
    \mathcal{L}(\theta,\phi,x) = \mathbb{E}_{q_{\phi}(z|x)}[log p_{\theta}(x|z)] - \beta D_{KL}(q_{\phi}(z|x) || p_{\theta}(z)).
\end{equation}

This allows us to analyse which degree of disentanglement is best so that the learned latent factors are informative for omics data analysis, since 
by having $\beta>1$ there is a stronger constraint bottleneck that limits the capacity of $z$, which should encourage the model to learn the most efficient representation of the data. Furthermore, to the best of our knowledge, an issue specific to training a VAE for omics data analysis, relates to what was encountered in \cite{cao_sailer_2021}, where the model is trained first without enforcing the Kullback-Leibner divergence, possibly due to the instabilities of calculating $D_{KL}$ in the early stages of training. Here, we also model this into the loss by having the loss be controlled by a time parameter $t$ that influences either a step function $\chi_{t>T}(t)$ that transitions from 0 to 1 when $t\geq T$, or a smooth function $S_{T_{s},T_{e}}(t) = \frac{1}{1+e^{-\frac{10}{e-s}(t-s)}}$ that is parameterized by an annealing start time $T_{s}$ and end time $T_{e}$, between which it smoothly progresses from a value $S_{T_{s},T_{e}}(T_{s}) \approx 0$ to $S_{T_{s},T_{e}}(T_{e}) \approx 1$. These functions modulate the degree to which the $\beta$ constraint is enforced alongside the training of the model, effectively avoiding the instabilities found in \cite{cao_sailer_2021} and in our work, giving us the final equations \ref{eq:bvae-step} and \ref{eq:bvae-smooth}, for the step and smooth functions respectively.

\begin{equation} \label{eq:bvae-step}
\footnotesize
    \mathcal{L}(\theta,\phi,x,t) = \mathbb{E}_{q_{\phi}(z|x)}[log p_{\theta}(x|z)] - \chi_{t>T}(t) \beta D_{KL}(q_{\phi}(z|x) || p_{\theta}(z)).
\end{equation}

\begin{equation} \label{eq:bvae-smooth}
\footnotesize
    \mathcal{L}(\theta,\phi,x,t) = \mathbb{E}_{q_{\phi}(z|x)}[log p_{\theta}(x|z)] - S_{T_{s},T_{e}}(t) \beta D_{KL}(q_{\phi}(z|x) || p_{\theta}(z)).
\end{equation}

\section{Results}

\subsection{Baselines}

We compare our proposed Pathway Activity Autoencoder (PAAE) and Pathway Activity with two baselines that have been frequently used in the literature: Autoencoders (AE) and Variational Autoencoders (VAE). Our AE baseline can be seen as roughtly equivalent to \cite{lee_incorporating_2020} as a main representative of works that use standard Autoencoders for early integration \cite{chaudhary_deep_2018,zhang_deep_2018,asada_uncovering_2020,lee_incorporating_2020}. Furthermore, our VAE baseline can also be seen as a representative of models that use Variational Autoencoders as their kernel for early multi-omics integration \cite{ronen_evaluation_2019,uyar_multi-omics_2021}. We have used our own implementation of the VAE framework to allow for more fine-tuned comparisons and hyperparameter tuning.

\subsection{Training and Testing Environment}

We train all models using batch training for 1024 epochs with a learning rate of $10^{-4}$ using the Adam optimiser under the \texttt{pytorch} v1.12.0 and the \texttt{skorch} v0.11.0 frameworks, all other components of our pipeline are built using \texttt{scikit-learn} v1.0.2. All models are trained with a dropout rate of $50\%$ and use rectified linear units as nonlinearities after each hidden layer except for the last layer in each module. Variational models have loss scheduling/warm-up iterations and do so with $T_{s} = 32$ and whenever there is some form of annealing, the annealing end time is set to 128 epochs after the start $T_{e} = T_{s}+128$.

For all pipelines, we first do one internal validation test on the LUSC TCGA dataset using 4-fold cross validation to choose the best hyperparameters for each pipeline using grid search on the values in Table~\ref{tab:gridsearch}, using the hyperparameter combination with highest ROC AUC (using the hidden latent representation $z$ or $\mu$) for external validation. We do 16 repeats of our external validation loop, which includes training our pipeline on the full LUSC TCGA dataset, re-normalising the input of the Metabric dataset before it is used as input to the rest of the pipeline.

\subsection{External Validation on Metabric}

\begin{figure}
    \centering
    \includegraphics[width=.45\textwidth]{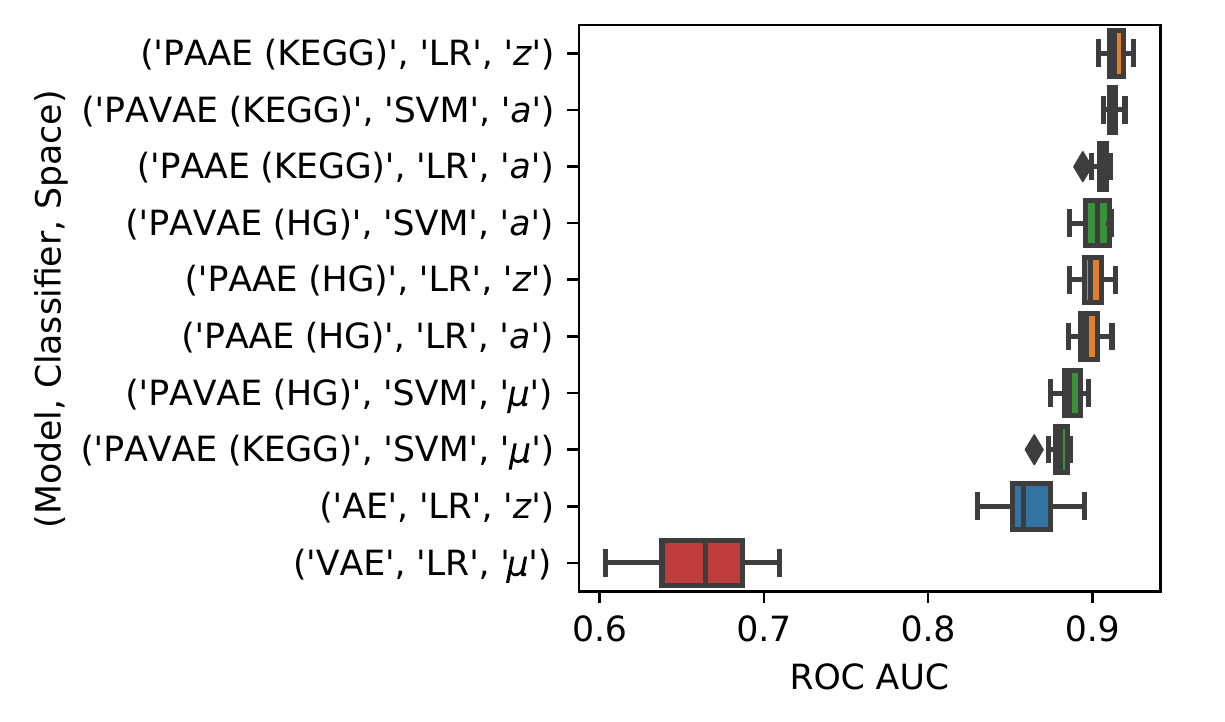}
    \caption{ROC AUC of a the pipeline tested on the Metabric dataset after being trained the TCGA dataset and tested.}
    \label{fig:clfbrcaext-results}
\end{figure}

Both our PAVAE and PAAE models significantly outperform ($p<10^{-3}$) our baselines in terms of the Area Under the Curve of the Receiver Operating Curve (ROC AUC), as can be seen in both Fig.~\ref{fig:clfbrcaext-results} and Tab.~\ref{tab:clf}, independently of if we used either the pathway activity space $a$ or the internal representation $z$ or $\mu$. 

Our PAVAE models both achieved a higher performance by using their pathway activity vectors $a$ ($p<10^{-3}$) instead of their hidden representation $\mu$, which was the value optimised during internal validation and hyperparameter tuning. In fact, our best combination of PAVAE model and representation was not significantly different to our best PAAE model and representation combination, showing that we achieved similar performance results on both frameworks, while the VAE baseline was still considered to be significantly worse ($p<10^{-3}$) than the AE baseline.

Furthermore, our proposed smooth $\beta$ scheduling function, $S_{T_{s},T_{e}}(t)$ was chosen as through the internal validation and hyperparameter tuning and, although significantly worse for the $\mu$ space when compared with the step function $\chi_{t>T}(t)$ we had previously seen in the literature \cite{cao_sailer_2021}, was tied for the best representation when using the $a$ space among all models.

\begin{table*}
    \centering
    \scriptsize
    \caption{Median (and IQR) results. Best value in \best{bold} ($p\leq0.05$). Clf stand for Classifier, LR for Logistic Regression and SVM for Support Vector Machine.}\label{tab:clf}
    \begin{tabular}{lllllllllll}
\toprule
Model & $\beta$ sched. & S & Clf. & \#Param & Test MSE & Accuracy & Precision & Recall & F1 & ROC AUC \\
\midrule
AE & & $z$ & LR & 4795105 & 33.758 (9.822) & 0.625 (0.050) & 0.663 (0.044) & 0.518 (0.030) & 0.525 (0.031) & 0.858 (0.023) \\
PAAE (HG) & & $z$ & LR & \textbf{1444883} & 25.730 (1.131) & 0.670 (0.025) & 0.666 (0.023) & 0.614 (0.034) & 0.626 (0.040) & 0.899 (0.011) \\
PAAE (HG) & & $a$ & LR & \textbf{1444883} & 25.730 (1.131) & 0.673 (0.017) & 0.693 (0.011) & 0.599 (0.025) & 0.622 (0.029) & 0.896 (0.010) \\
PAAE (KEGG) & & $z$ & LR & 1628955 & \textbf{19.611 (1.293)} & \textbf{0.692 (0.014)} & 0.665 (0.019) & \textbf{0.650 (0.017)} & \textbf{0.647 (0.021)} & \textbf{0.914 (0.009)} \\
PAAE (KEGG) & & $a$ & LR & 1628955 & \textbf{19.611 (1.293)} & 0.673 (0.022) & 0.705 (0.027) & 0.603 (0.028) & 0.616 (0.027) & 0.907 (0.004) \\
\midrule
VAE & $\chi_{t>T}(t)$ & $\mu$ & LR & 4803361 & 42009 (209076) & 0.473 (0.055) & 0.300 (0.067) & 0.368 (0.038) & 0.242 (0.036) & 0.664 (0.049) \\ %VAE-$\mathbbm{1}$-[128, 64] & $z$ & 4803361 & 42009.199 (209076.647) & 0.473 (0.055) & 0.300 (0.067) & 0.368 (0.038) & 0.242 (0.036) & 0.664 (0.049) \\
PAVAE (HG) & $\chi_{t>T}(t)$ & $\mu$ & SVM & 2437089 & 29.274 (2.209) & 0.640 (0.015) & 0.647 (0.029) & 0.574 (0.020) & 0.585 (0.018) & 0.886 (0.010) \\
PAVAE (HG) & $\chi_{t>T}(t)$ & $a$ & SVM & 2437089 & 29.274 (2.209) & 0.654 (0.023) & 0.696 (0.036) & 0.571 (0.021) & 0.582 (0.033) & 0.903 (0.014) \\
PAVAE (KEGG) & $S_{T_{s},T_{e}}(t)$ & $\mu$ & SVM & 2459837 & 24.603 (0.474) & 0.631 (0.013) & 0.620 (0.028) & 0.578 (0.017) & 0.570 (0.022) & 0.880 (0.007) \\
PAVAE (KEGG) & $S_{T_{s},T_{e}}(t)$ & $a$ & SVM & 2459837 & 24.603 (0.474) & 0.644 (0.023) & \textbf{0.730 (0.036)} & 0.552 (0.033) & 0.553 (0.044) & \textbf{0.912 (0.004)} \\
\bottomrule
\end{tabular}

%AE-[128, 64]
%{PAAE-[32]-[64] (HG)}
%{PAAE-[32]-[64] (KEGG)}
%VAE-$\mathbbm{1}$-[128, 64]
%{PAVAE-$\sigma$-[]-[128,64] (KEGG)}
%{PAVAE-$\mathbbm{1}$-[]-[128,64] (HG)}

\end{table*}

\subsection{Model Analysis}

\subsubsection{Parameter and Layer Efficiency}

\begin{figure}
    \centering
    \includegraphics[width=.45\textwidth]{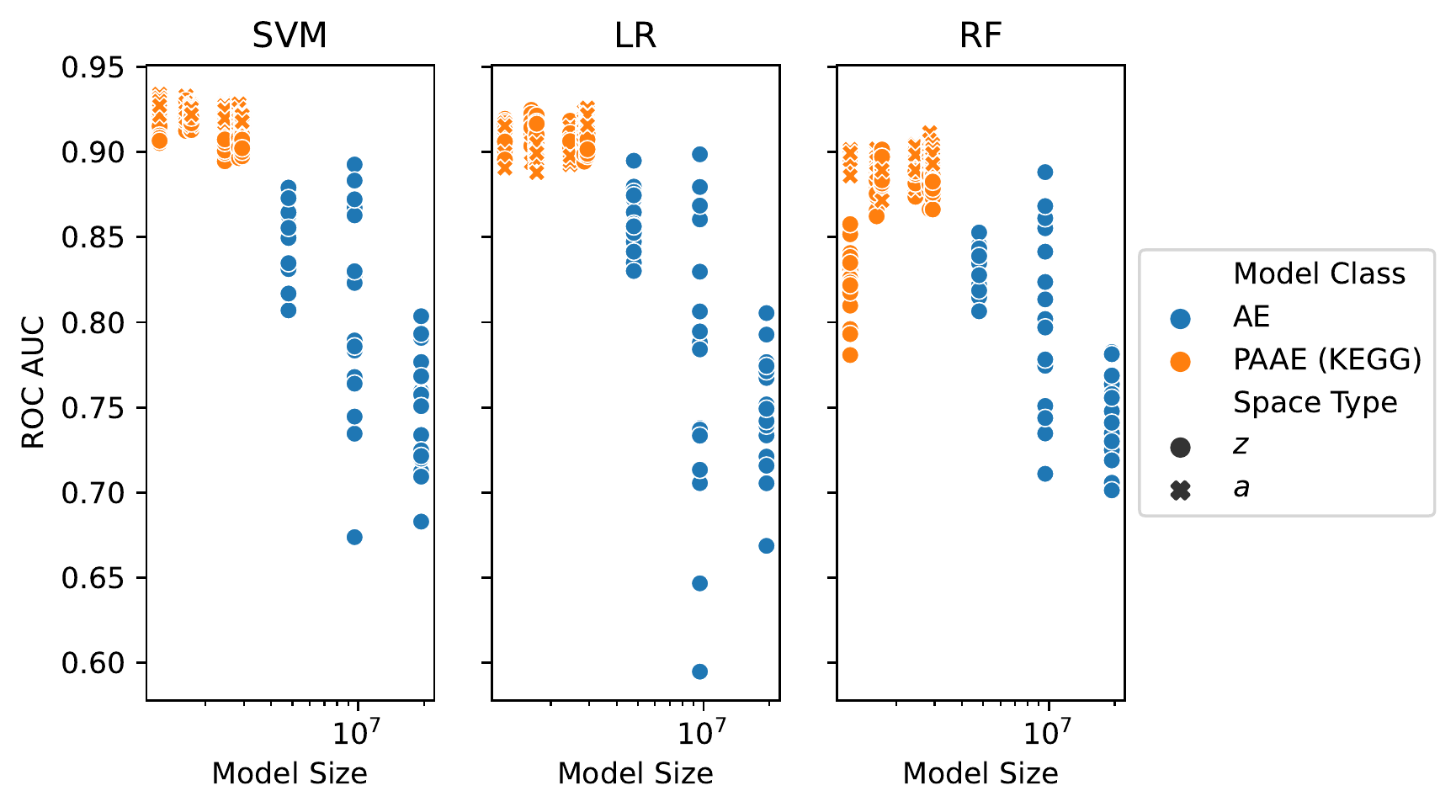}
    \caption{Parameter efficiency plot comparing our best model (PAAE (KEGG)) against our best baseline (AE). We show performances for both $z$ and $a$ spaces of our model.}
    \label{fig:brca-ext-ablation-paae-parameters}
\end{figure}

To justify our model architecture choices, we performed multiple ablation studies to evaluate their impact. In Figures~\ref{fig:brca-ext-ablation-paae-layers} and \ref{fig:brca-ext-ablation-paae-parameters} we can see that our PAAE model has a higher parameter efficiency than the AE model, achieving better or similar performance with less parameters irrespective of representation used. For both Logistic Regression and Support Vector Machines, both the pathway activity space and the hidden latent representation provided better performance than our baseline. However, for the Random Forest classifier this was valid only when our model had nonlinear layers (Fig.~\ref{fig:brca-ext-ablation-paae-layers}).

The effect of $\beta$ term of VAE on reconstruction as well as classification performance was investigated, and we found that it did not improve either. This had already been described in \cite{kingma_auto-encoding_2014}, which pointed to the information loss in the latent space due to normalisation. In fact, when training some of the VAE models without any $\beta$ scheduling technique, many model configurations diverged due to the same problems faced by \cite{cao_sailer_2021}, and depending on other factors such as input normalization (i.e., z-scoring instead of percentile-scoring) also fail despite our efforts to mitigate these using warm-up sequences. We also performed tests by having a smooth transition from the cold-start of the VAE to the point where the KL divergence was being applied to the VAE to the maximum $\beta$ value for the model, as well as a step transition, both of which yielded similar results, with the step functions performing slightly better for larger amounts of $\beta$, but being less stable and causing the models to diverge more often.

\subsubsection{Feature Space Topology and Interpretability}

Feature spaces generated by our models were explored in order to establish that: (1) our models provide a meaningful feature space and (2) that our models are interpretable with respect to providing insight to the contribution of pathways in clinical phenotype prediction and therefore prognosis. 

In terms of the feature space, our models provide a more meaningful feature space (see Fig.~\ref{fig:featspace}) and better separation between each of the classes is achieved compared to the relevant baselines. In fact, our models already provide a meaningful representation in the pathway space, with the further transformation on our models generally improving separation and providing a slightly different, but still recognizable, feature space. In contrast, the AE and VAE models seem to fail to provide a good separation altogether.

\begin{figure*}
    \centering
    \subcaptionbox{TCGA (train)\label{fig:clustermap:sub:tcga}}{\includegraphics[width=.45\linewidth]{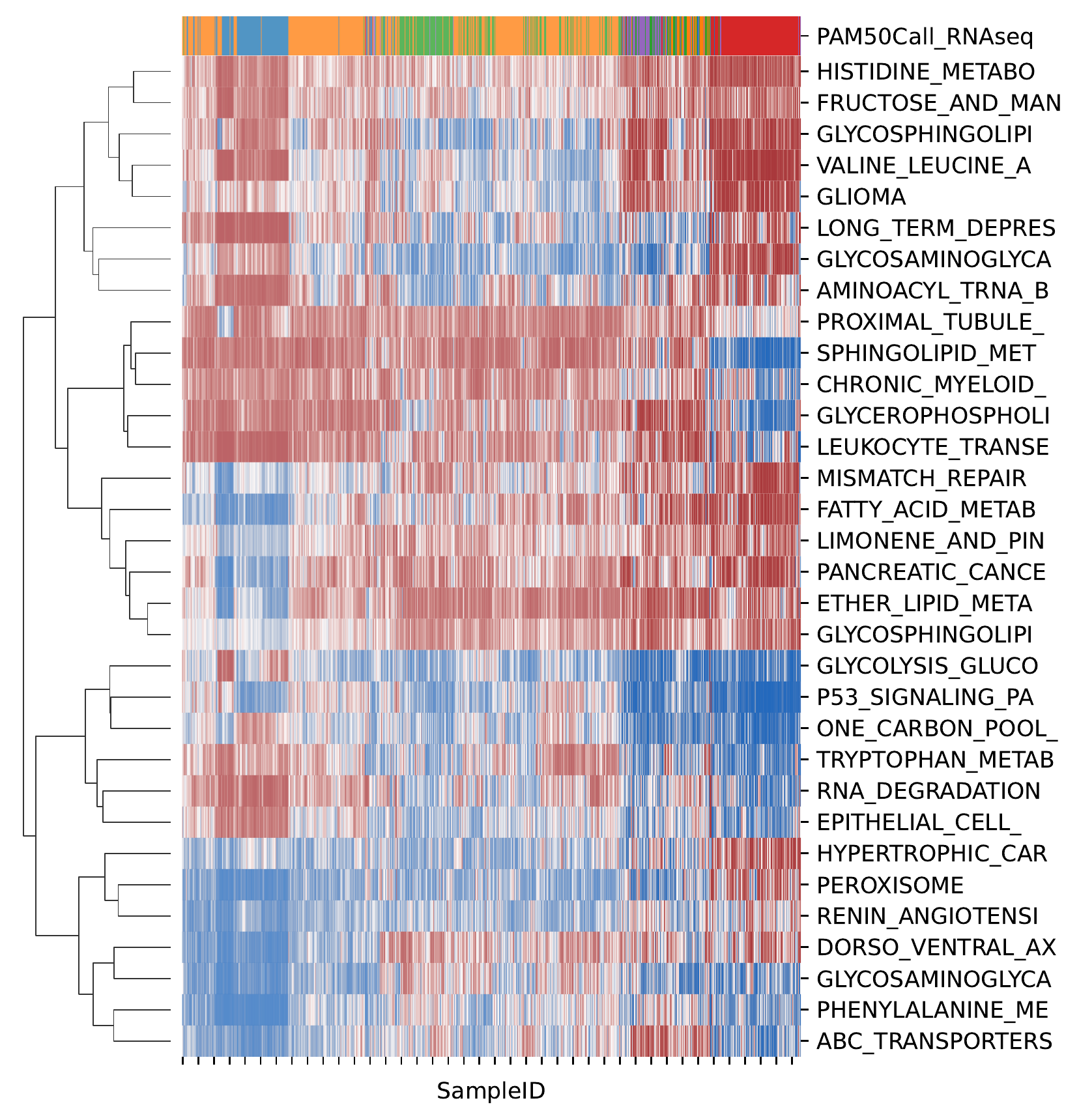}}
    \hfill
    \subcaptionbox{Metabric (test)\label{fig:clustermap:sub:meta}}{\includegraphics[width=.45\linewidth]{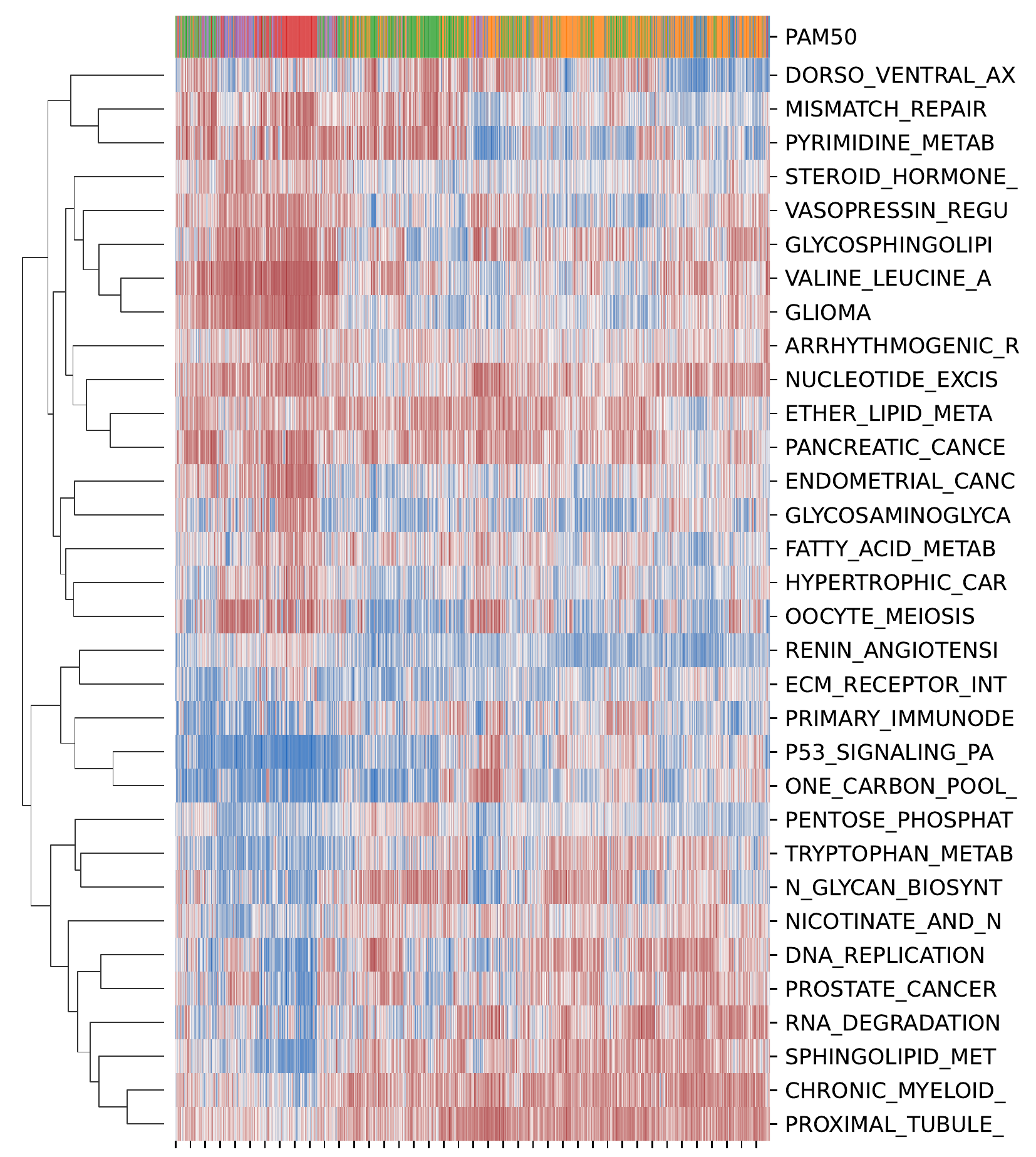}}
    
    \caption{Clustermap based on the cosine distance between sample inferred pathway activity vectors for KEGG PAAE's pathway activity space. Colours indicate clinical phenotypes for BRCA ({\color{brca-normal}Normal in blue}, {\color{brca-luma}Luminal A in orange}, {\color{brca-lumb}Luminal B in green}, {\color{brca-basal}Basal in red}, {\color{brca-her2}Her2 in purple}). For legibility we only provide the 32 pathways with highest mutual information with respect to the classes. Fig.~\ref{fig:clustermap-hallmark} shows similar results for the Hallmark Genes pathway set across all of its 50 pathways}
    \label{fig:clustermap}
\end{figure*}

With regards to interpretability, as shown in Fig.~\ref{fig:clustermap} that clustering samples through the cosine distance of their pathway activity vectors provides separation of their clinical labels, even though our model is trained in an entirely unsupervised fashion, illustrating that augmentation via pathway prior knowledge benefits our model in both interpretability and predictive power. Furthermore, the majority of pathways that were among the 32 with most mutual information w.r.t. the classification target in our training set were also considered to be among the most significant in the test dataset, showing that the features learned by our model are consistent in their informational content even across datasets.

\begin{figure*}
    \centering
    \subcaptionbox{\label{fig:featuremap:sub:class}}{\includegraphics[width=.3\linewidth]{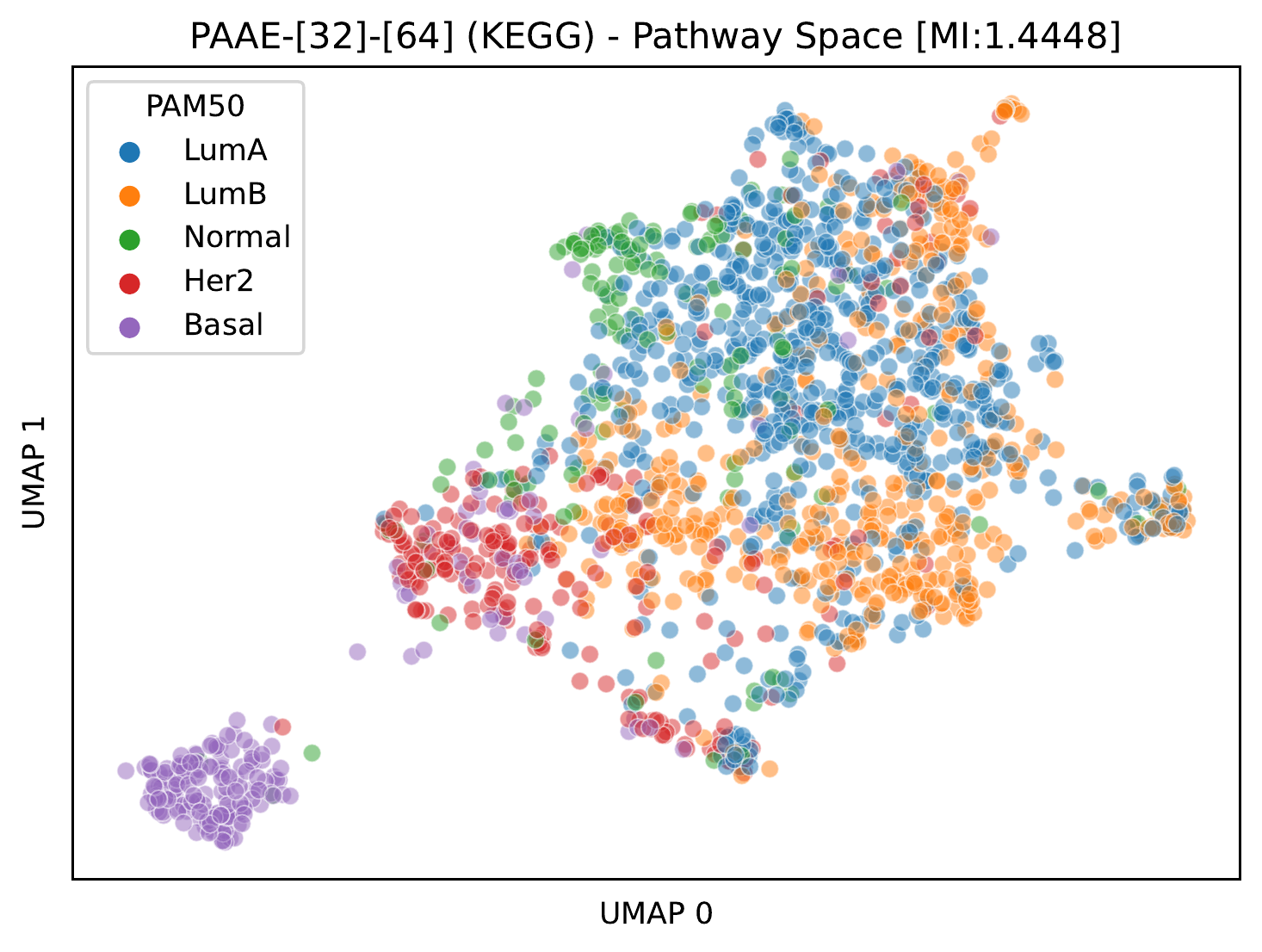}}
    \hfill
    \subcaptionbox{\tiny VALINE\_LEUCINE\_AND\_ISOLEUCINE\_DEGRADATION \label{fig:featuremap:sub:VALINE_LEUCINE_AND_ISOLEUCINE_DEGRADATION}}{\includegraphics[width=.3\linewidth]{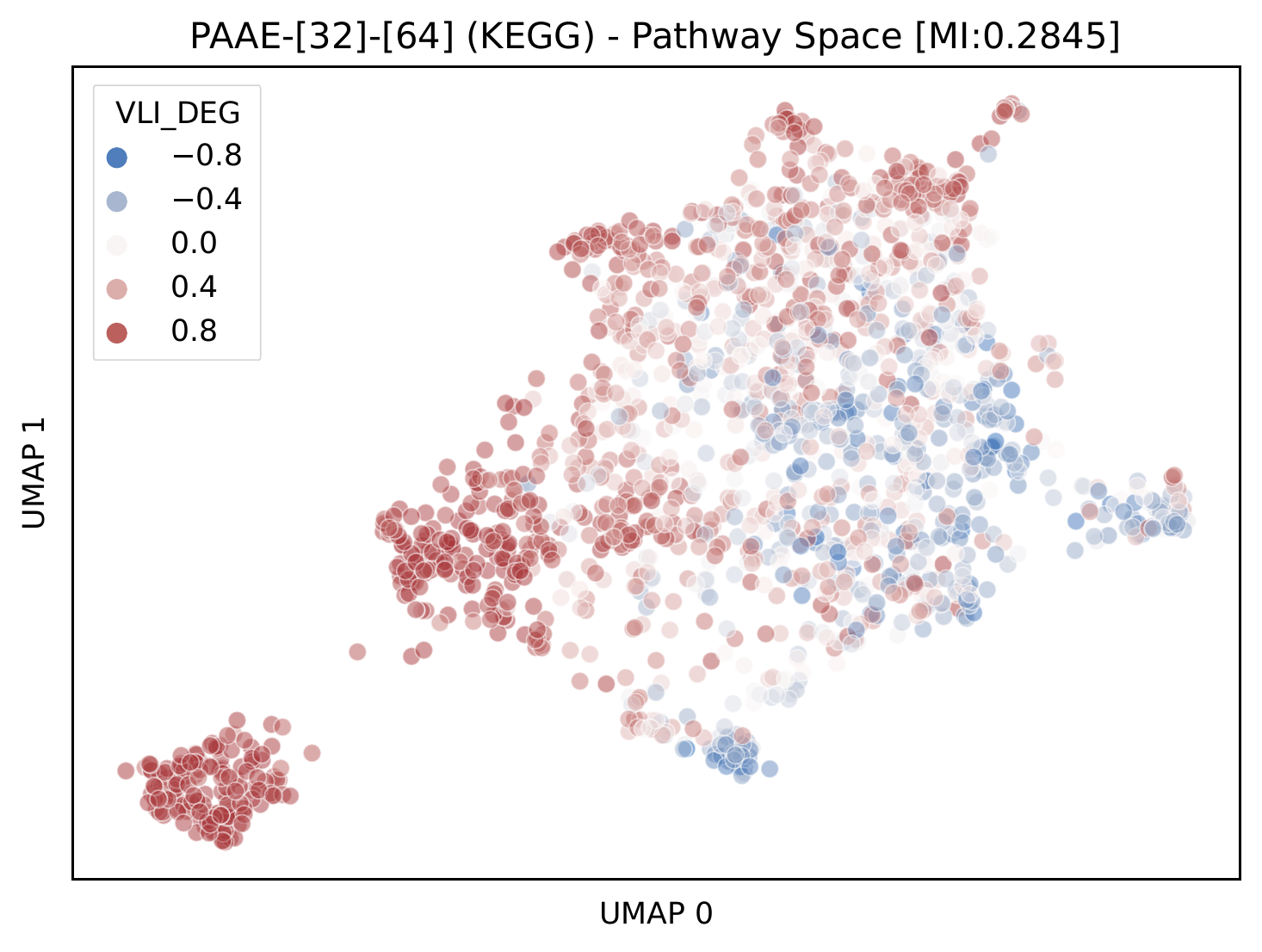}}
    \hfill
    \subcaptionbox{\tiny P53\_SIGNALING\_PATHWAY \label{fig:featuremap:sub:P53_SIGNALING_PATHWAY}}{\includegraphics[width=.3\linewidth]{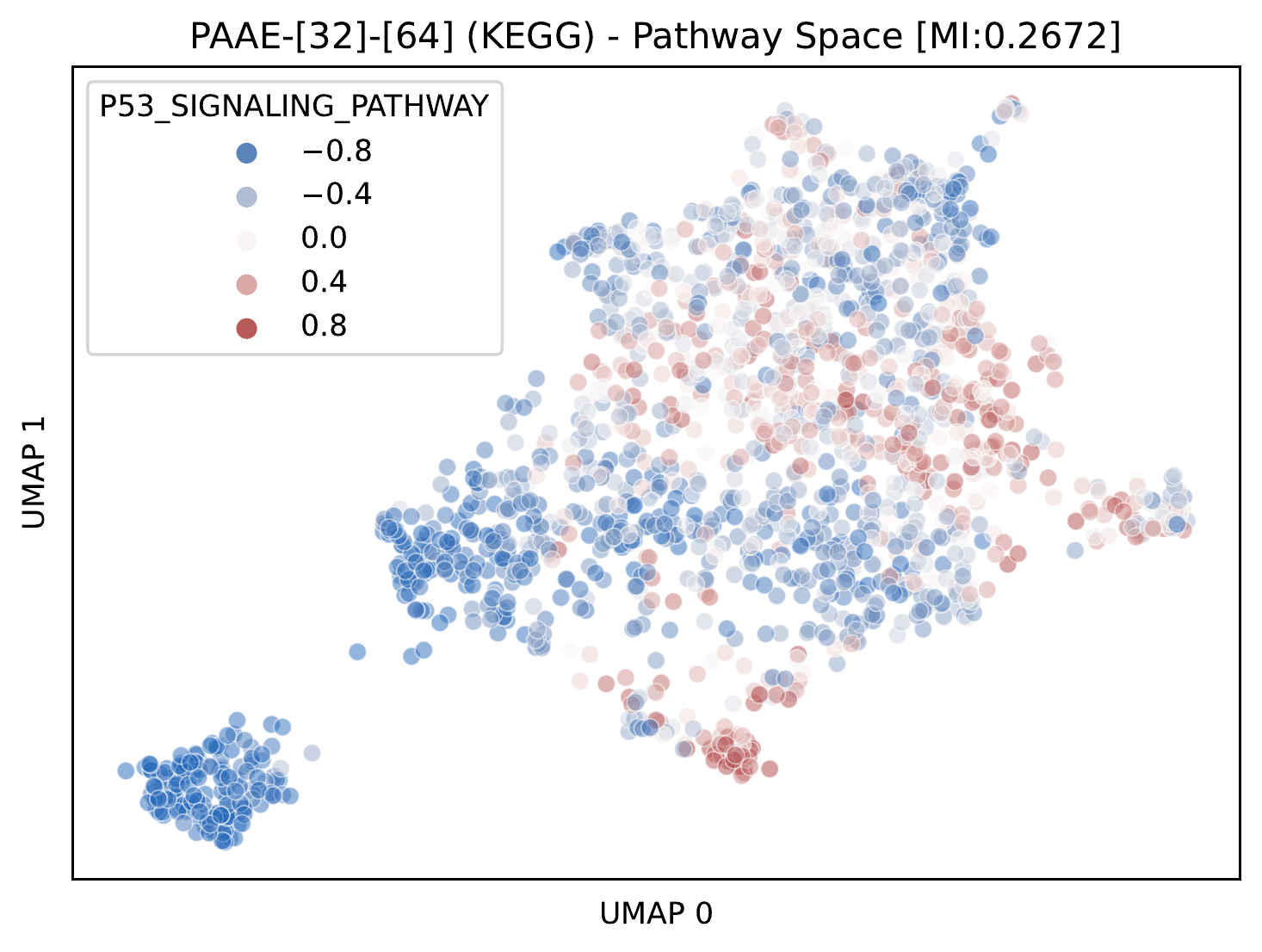}}
    \\
    \subcaptionbox{\tiny GLIOMA \label{fig:featuremap:sub:GLIOMA}}{\includegraphics[width=.3\linewidth]{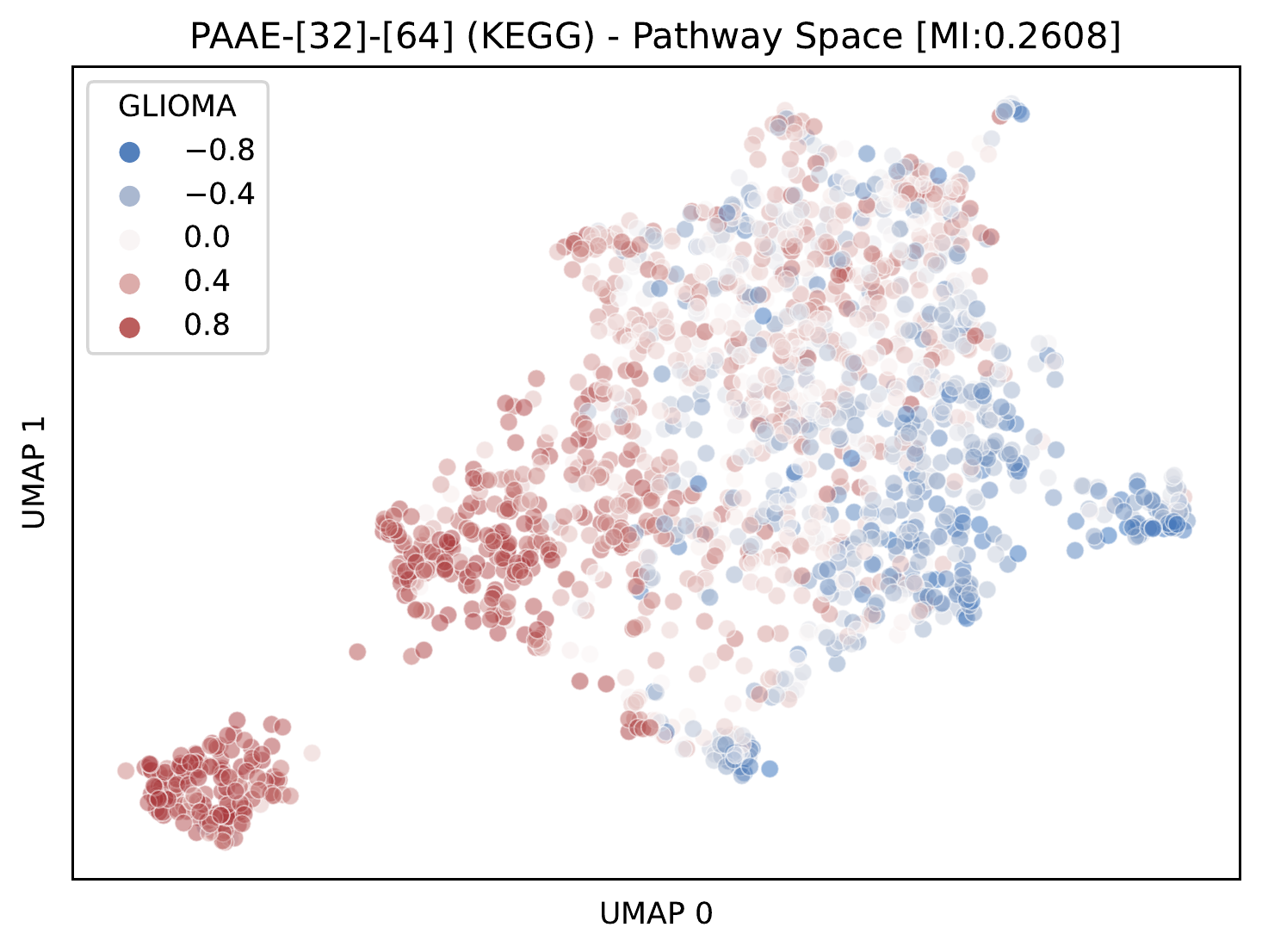}}
    \hfill
    \subcaptionbox{\tiny SHPINGOLIPID\_METABOLISM \label{fig:featuremap:sub:SHPINGOLIPID_METABOLISM}}{\includegraphics[width=.3\linewidth]{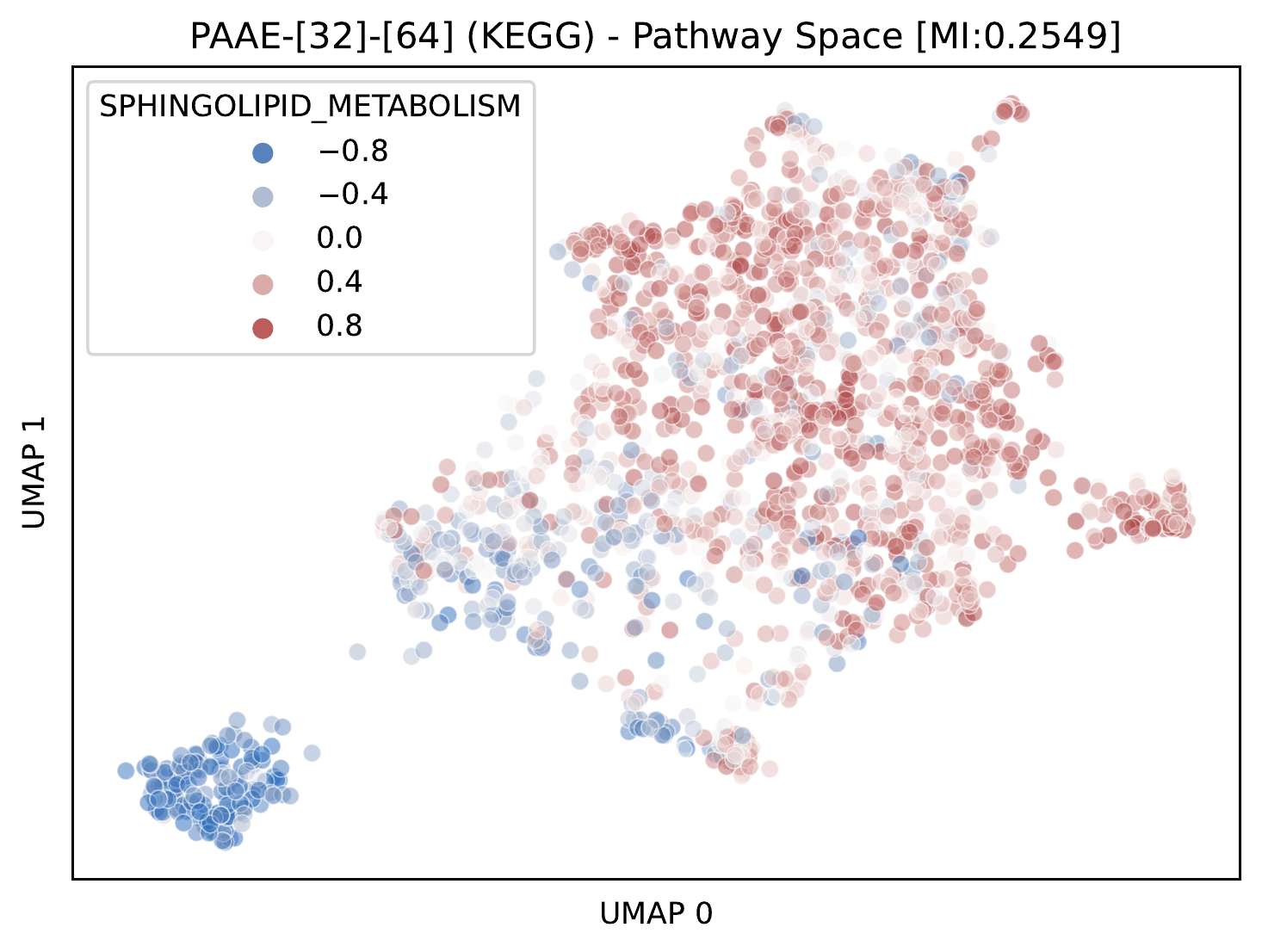}}
    \hfill
    \subcaptionbox{\tiny MISMATCH\_REPAIR \label{fig:featuremap:sub:MISMATCH_REPAIR}}{\includegraphics[width=.3\linewidth]{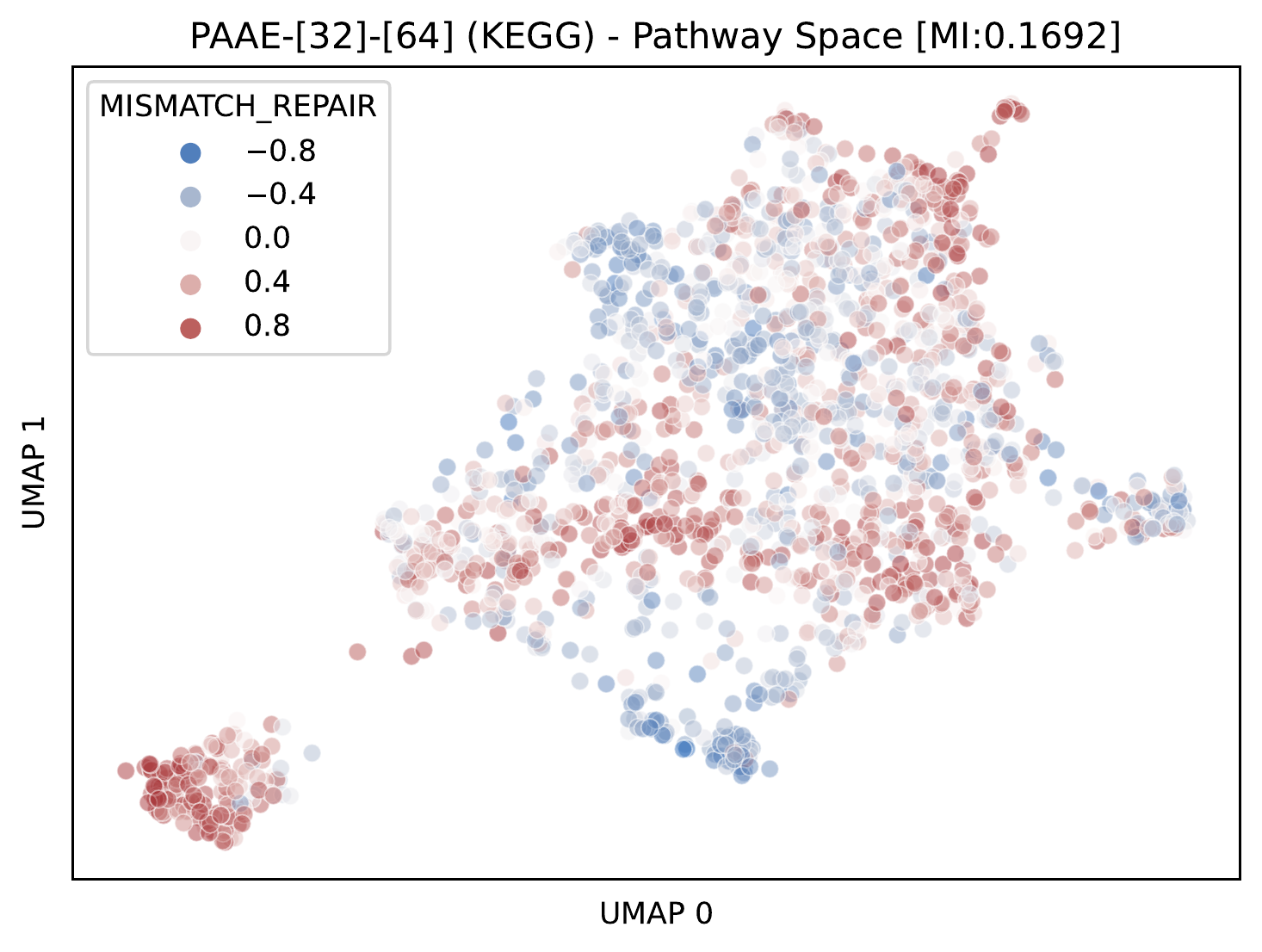}}
    
    \caption{The featuremap showing the intensity of each sample's inferred pathway activity vectors for KEGG PAAE's pathway activity space in the Metabric dataset, as well as the class distribution overlayed on top of the 2-dimensional UMAP reduction. For conciseness we only show here the 5 pathways that have the most mutual information w.r.t. the classes. See Figs.~\ref{fig:featuremap-tcga} and \ref{fig:featuremap-meta} for the same plots on the TCGA and Metabric datasets showing pathways that are among the 5 most relevant on either dataset as well as the same plots for the Hallmark Genes pathway set in Figs.~\ref{fig:featuremap-tcga-hallmark} and \ref{fig:featuremap-meta-hallmark}.}
    \label{fig:featuremap}
\end{figure*}

We can also provide a visualisation that is both informative of the topology of the pathway space as well as show how individual features affect the space. We extend the concept of a feature map \cite{zeiler_visualizing_2014} from a Convolutional Neural Network's to the 2d reduction of our Pathway Activity space, as shown in Fig.~\ref{fig:featuremap}, with a visualisation of both the reduced space with the classes as well as the activations of each pathway activity score. We also provide further interpretability studies on the appendix, see Figs.~\ref{fig:clustermap-kegg-importantgenes} and \ref{fig:clustermap-hallmark-importantgenes}, and Tabs.~\ref{tab:important-genes-kegg} and \ref{tab:important-genes-hallmark} for interpretability analyses going up to individual features, and Figs.~\ref{fig:survival-kegg-importantgenes} and \ref{fig:survival-hallmark-importantgenes} for survival analyses of some of these individually-discovered features.

\section{Discussions and Conclusions}

In this paper we propose both a non-generative and generative, unsupervised, prior-knowledge-based, auto-encoding deep learning framework to improve patient stratification and prognosis via analysis of gene expression data. Our results show that our proposed pathway activity Autoencoder frameworks offer promising means of using biomedical prior knowledge to facilitate unsupervised deep learning methods for omics data analysis. Although we focus mainly on unsupervised dimensionality reduction for classification, results translate into better latent representations, while also providing an intermediate latent representation that is directly translatable into a concept relevant to medical practitioners.

Furthermore, our results so far show that the commonly applied technique of using VAEs (which are generative models in nature) to generate latent representations might possibly be harmful for analyses, something that also seemed to be corroborated by our previous study where the variational model performed worse than our proposed methods \cite{da_costa_avelar_multi-omic_2023}. Despite this worse overall performance, we showed that our proposed generative framework, although still generally worse than our non-generative models, were generally more robust than the other generative models while also providing better reconstruction results (Fig.~\ref{fig:clfbrcaext-reconstruction-results}).

Future work can target extension of our framework to allow data integration of multiple omics levels by mapping pathways that describe complex molecular interactions, and we found that this first foray into pathway-activity-augmented neural networks serves as demonstration of the potential that exists in analysis of biomedical systems through informed machine learning \cite{von_rueden_informed_2021}.

\section*{Acknowledgements} The Authors would like to thank Roman Laddach for help with data acquisition, Roman Laddach and Wong Wai Yee for fruitful discussions, and João Nuno Beleza Oliveira Vidal Lourenço for designing the diagrams. The Authors acknowledge funding by King's College London and the A*STAR Research Attachment Programme (ARAP) to PHCA. The results shown here are in whole or part based upon data generated by the TCGA Research Network: \url{https://www.cancer.gov/tcga}. WM acknowledges funding by the AI, Analytics and Informatics (AI3) Horizontal Technology Programme Office (HTPO) seed grant (grant no: C211118015) from A*STAR, Singapore. 

\bibliographystyle{plain}
\bibliography{paae}

\clearpage
\onecolumn

\section*{Supplementary Material}
\setcounter{section}{1} \renewcommand\thesection{S}
\setcounter{table}{0} \renewcommand\thetable{S\arabic{table}}
\setcounter{figure}{0} \renewcommand\thefigure{S\arabic{figure}}
\setcounter{equation}{0} \renewcommand\theequation{S\arabic{equation}}

\subsection{Materials and Methods}

\subsubsection{Dataset links}

\begin{itemize}
    \item BRCA Gene Expression \url{https://xenabrowser.net/datapages/?dataset=TCGA-BRCA.htseq_fpkm.tsv&host=https%3A%2F%2Fgdc.xenahubs.net&removeHub=https%3A%2F%2Fxena.treehouse.gi.ucsc.edu%3A443}
    \item BRCA Subtypes \url{https://xenabrowser.net/datapages/?dataset=TCGA.BRCA.sampleMap%2FBRCA_clinicalMatrix&host=https%3A%2F%2Ftcga.xenahubs.net&removeHub=https%3A%2F%2Fxena.treehouse.gi.ucsc.edu%3A443}
    \item Ensembl Gene ID to Gene name mapping: \url{https://emckclac-my.sharepoint.com/:u:/g/personal/k21126169_kcl_ac_uk/EdTc-1e_cYJJk65a66L9nR8BY7_JmZksS5Am2xDHJJOjaw?e=5a2WgJ}
    \item Metabric GEx \url{https://www.cbioportal.org/study/summary?id=brca_metabric}
    \item Metabric Subtypes \url{https://www.cbioportal.org/study/clinicalData?id=brca_metabric}
    \item OncoKB Cancer gene search \url{https://www.oncokb.org/cancerGenes}
\end{itemize}

\subsubsection{Autoencoder Architecture and Training Environment}

We train all models using batch training for 1024 epochs with a learning rate of $10^{-4}$ using the Adam optimiser under the \texttt{pytorch} v1.12.0 and the \texttt{skorch} v0.11.0 frameworks, all other components of our pipeline are built using \texttt{scikit-learn} v1.0.2. All models are trained with a dropout rate of 50\%.

All the models tested use rectified linear units as nonlinearities after each hidden layer except for the last layer in each module, i.e., the last layer in the pathway encoders, and the latent space encoder and decoder are all simple linear layers without any nonlinearity. We do this to allow our model to output any real value, and, due to our network initialization method, to produce roughly normal outputs.

For the variational models we use the loss scheduling/warm-up procedure described in the main paper, using $T_{s}=32$ and with annealing duration $128$ for $T_{e} = T_{s} + 128$ in our annealed scheduling function $S_{T_{s},T_{e}}(t)$.

For all pipelines, we first perform an internal 4-fold cross validation for hyperparameter tuning, selecting the model with highest ROC AUC from a grid search in the parameters shown in Tab.~\ref{tab:gridsearch}, where the classifier was applied to the hidden representation vector (i.e., $z$ or $\mu$), which resulted in the following hyperparameters being chosen:

\begin{itemize}
    \item Autoencoder: Encoder with 2 layers with 128, and 64 units; using a Logistic Regression classifier.
    \item Pathway Activity Autoencoders: Pathway Activity Encoders with 2 layers with a 32-unit hidden layer; an Encoder with only a single 64-unit layer; and using using a Logistic Regression classifier.
    \item Variational Autoencoder: Encoder with 2 layers with 128, and $2\times64$ units (64 for $\mu$ and 64 for $\sigma$); using a Logistic Regression classifier.
    \item Pathway Activity Variational Autoencoders: A linear Pathway Activity Encoders with only a single linear layer; Encoder with 2 layers with 128, and $2\times64$ units (64 for $\mu$ and 64 for $\sigma$); and using using a Support Vector Machine classifier.
\end{itemize}

For all Variational models we use a gaussian distribution defined by $\mu$ and $\sigma$, from which the values of $z$ are sampled using the reparametrization trick to maintain differentiability \cite{kingma_auto-encoding_2014}.

We train the Logistic Regression models with l2 regularization with $C=1$, for a maximum of $100$ iterations using the lbfgs solver, and using the softmax function for the multi-class cases. For the Support Vector Machine classifiers we use $C=1$ as the l2 regularization parameter and a Radial Basis Function kernel. For the Random Forest Classifier we build 100 Trees using Gini impurity, considering $\sqrt{d_{input}}$ features for splitting, boostrapping samples when building trees.

\begin{table*}[bpht]
    \centering
    \scriptsize
    \caption{Hyperparameters used during the internal validation grid search.}
    \label{tab:gridsearch}
    \begin{tabular}{cccccc}
         \toprule
         Model & Encoder Layer Sizes & $\beta$ & $\beta$ Schedule & Pathway Module Hidden Layer Sizes & Classifier\\
         \midrule
         AE & [128,64], [256,128,64], [512,256,128,64] & N/A & N/A & N/A & LR,SVM,RF \\
         VAE & [128,64], [256,128,64], [512,256,128,64] & 1,5,10,50,100 & 
         $\chi_{t>32}(t)$, $S_{32,160}(t)$ & N/A & LR,SVM,RF \\
         PAAE & [64],[128,64] & N/A & N/A & [],[32],[32,16] & LR,SVM,RF \\
         PAVAE & [128,64], [256,128,64] & 1,5,10,50,100 & $\chi_{t>32}(t)$, $S_{32,160}(t)$ & [],[32],[32,16] & LR,SVM,RF \\
         \bottomrule
    \end{tabular}
\end{table*}

After the hyperparameters were selected, we train the models in a completely unsupervised fashion on our training dataset (TCGA) and then use the models to compress the input to their latent representation, using $z$ for the AE and PAAE models and the posterior means $\mu$ for the VAE and PAVAE models, as well as a separate compression of our input to its pathway activity score representation $a$ for Pathway Activity models, then using this compressed latent representation as the input for training the classifier.

After training, we then normalize the input of our test dataset using the same method as was used for the training dataset, since our test dataset is on a completely different scale and was produced by a different gene expresssion measuring technology. We then use this as input for the all models to compress the features, and use the compressed representation as input for the classifiers that produce predictions of each sample's cancer subtypes.

For all our tests we consider the Area Under the Receiver Operating Characteristic Curve (ROC AUC), macro-averaged, as the main result for each set and, unless otherwise stated, we apply the Wilcoxon Rank-Sum test for testing statistical significance between different medians. We also provide other metrics such as accuracy, precision, recall and F1 score. Note that our Pathway Activity methods can't access the full set of input genes when building their internal representation, while at the same time having to rebuild the entirety of the input.

\subsubsection{Methodological Details for Interpretability}

Finally, for the analyses involving raw input features, such as those in Figs.~\ref{fig:clustermap-kegg-importantgenes}, \ref{fig:clustermap-hallmark-importantgenes}, \ref{fig:survival-kegg-importantgenes} and \ref{fig:survival-hallmark-importantgenes}, we re-normalise the genes from $\log(FPKM+1)$ to $\log(TPM+1)$ in the TCGA dataset, and from $\log(Intensity+1)$ to $\log(IPM+1)$, where FPKM stands for Fragments Per Kilobase Million, TPM for Transcripts Per Million, Intensity is the raw intensity value measured in the Metabric dataset, and IPM is ``Intensity Per Million'', where the intensity is normalised per sample as if it was a raw count value with $IPM_{g} = \frac{I_{g}\times10^{6}}{\sum_{g}I_{g}}$.

For clustermaps, in Figs.~\ref{fig:clustermap-kegg} and \ref{fig:clustermap-hallmark}, we cluster along both rows and columns, using the cosine distance metric, as it is frequently used for deep learning embeddings. In the clustermaps that involve raw features (Figs.~\ref{fig:clustermap-kegg-importantgenes} and \ref{fig:clustermap-hallmark-importantgenes}), however, we only cluster by column, and use the Euclidean distance. For all clustermaps, we use \texttt{seaborn}'s \texttt{clustermap} function.

For all survival-related plots and statistical tests (Figs.~\ref{fig:survival-kegg-importantgenes} and \ref{fig:survival-hallmark-importantgenes}) we set a time window limit of 5 years ($365\times5=1825$ days) on the survival data. We use the \texttt{lifelines} library for both the tests and the plots. All of the analyses are done in an almost-unsupervised fashion, being that the only supervision we use is that we choose the top 5 pathways w.r.t. mutual information with the PAM50 subtypes, and some subtypes might have different survival outcomes.

\subsection{Results}

\begin{figure}
    \centering
    \includegraphics[width=.45\textwidth]{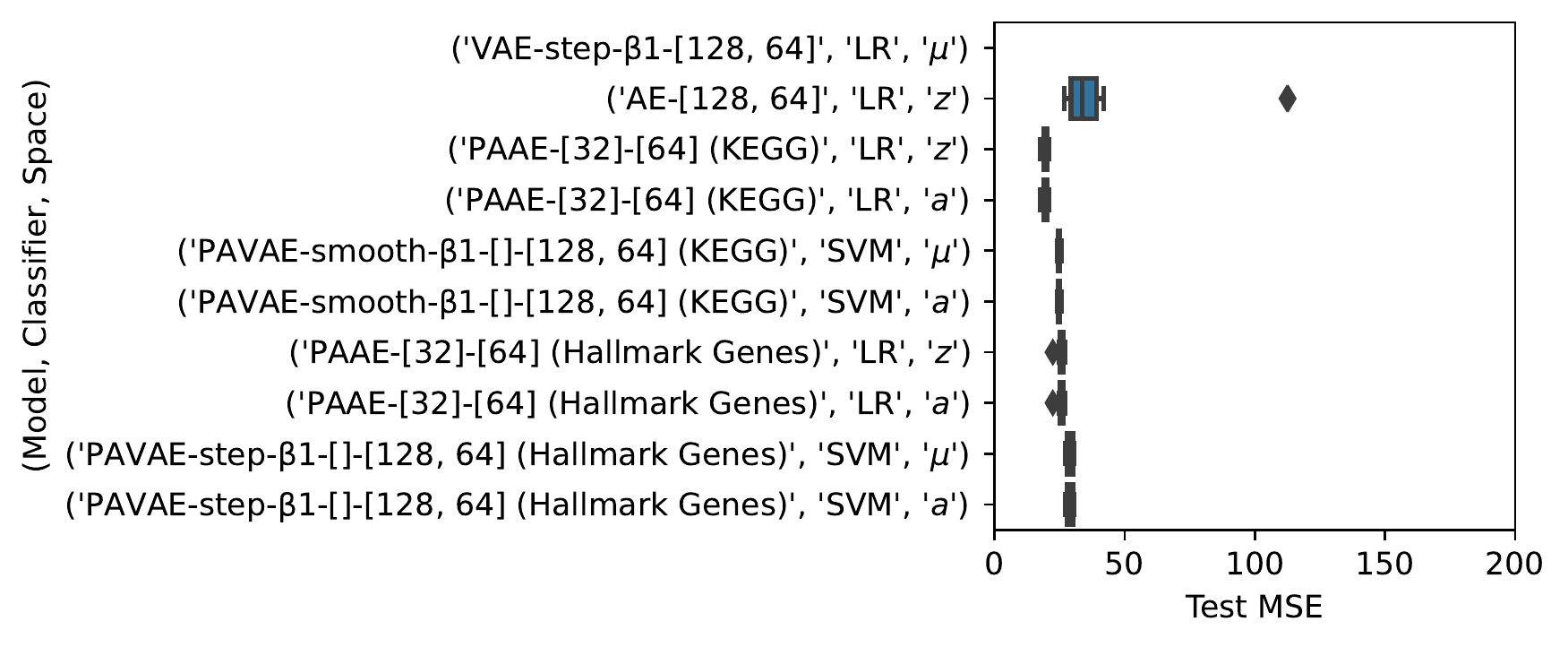}
    \caption{Test MSE for the $log_2(fpkm+10^{-3})$ normalisation. Note that the vanilla VAE results were ommitted for legibility.}
    \label{fig:clfbrcaext-reconstruction-results}
\end{figure}

Fig.~\ref{fig:clfbrcaext-reconstruction-results} shows the Test MSE values from the models trained in the TCGA dataset and tested on the Metabric dataset.

\subsection{Ablation Studies}

\begin{figure}
    \centering
    \includegraphics[width=.45\textwidth]{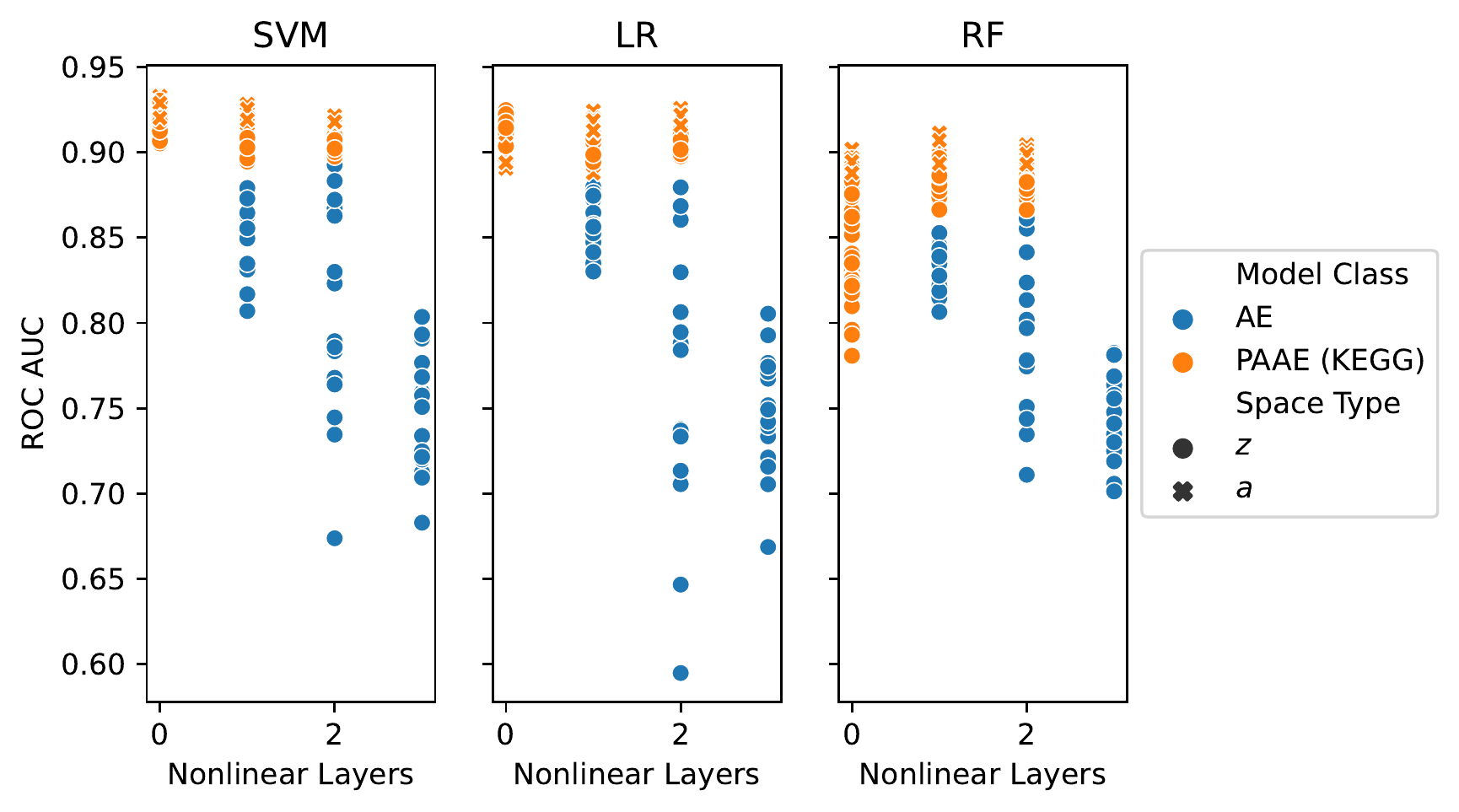}
    \caption{Layer efficiency plot comparing our best model (PAAE (KEGG)) against our best baseline (AE). We performances for both $z$ and $a$ spaces of our model.}
    \label{fig:brca-ext-ablation-paae-layers}
\end{figure}

Fig.~\ref{fig:brca-ext-ablation-paae-layers} shows that our PAAE model performs better than the traditional AE model regardless of the number of nonlinearities, and that it's more robust within this hyperparameter choice.

\subsection{Interpretability}

\begin{figure*}
    \centering
    \subcaptionbox{AE - $z$\label{fig:featspace:sub:ae}}{\includegraphics[width=.3\textwidth]{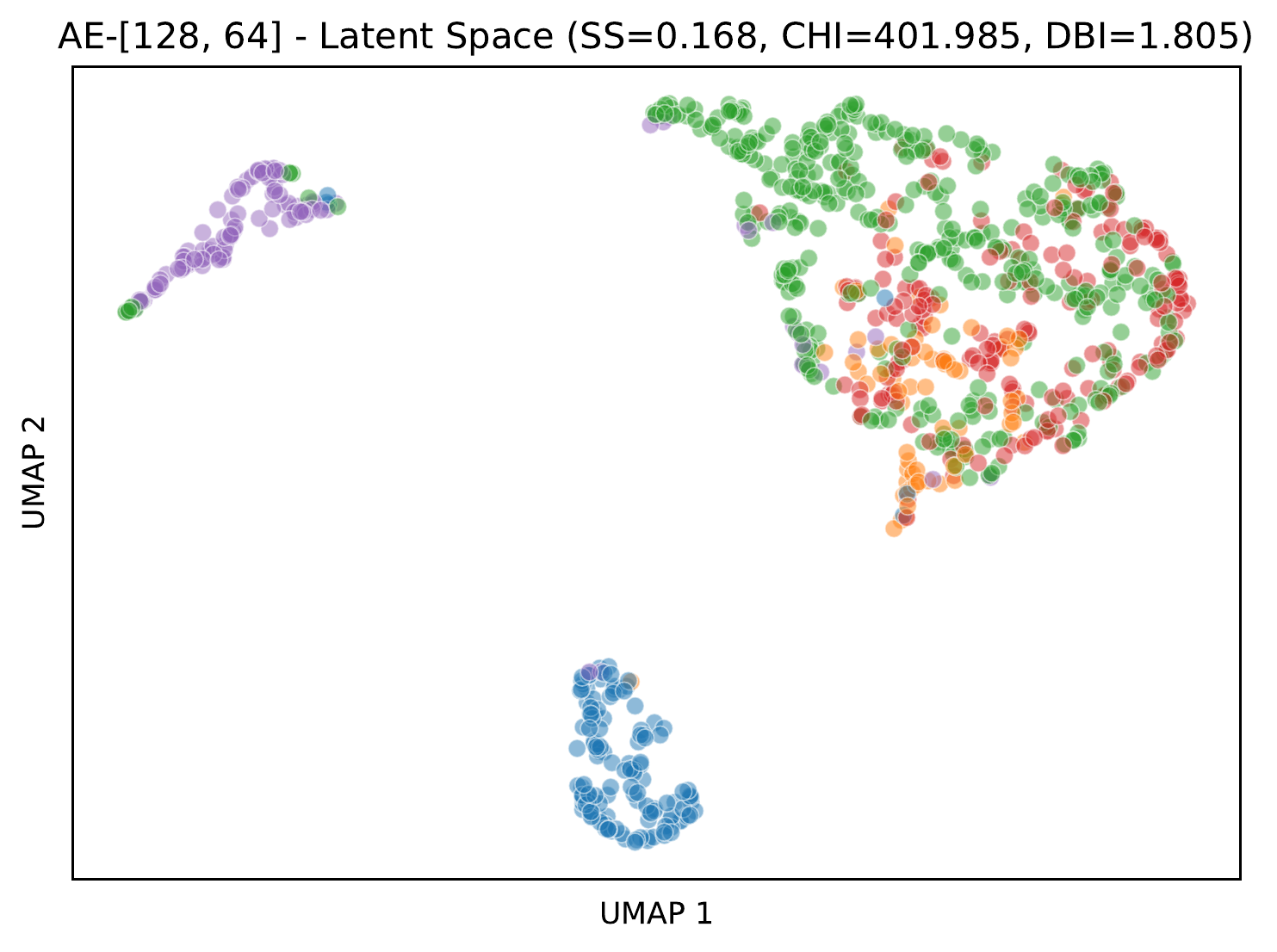}}
    \hfill
    \subcaptionbox{PAAE - $a$\label{fig:featspace:sub:paae-pathway}}{\includegraphics[width=.3\textwidth]{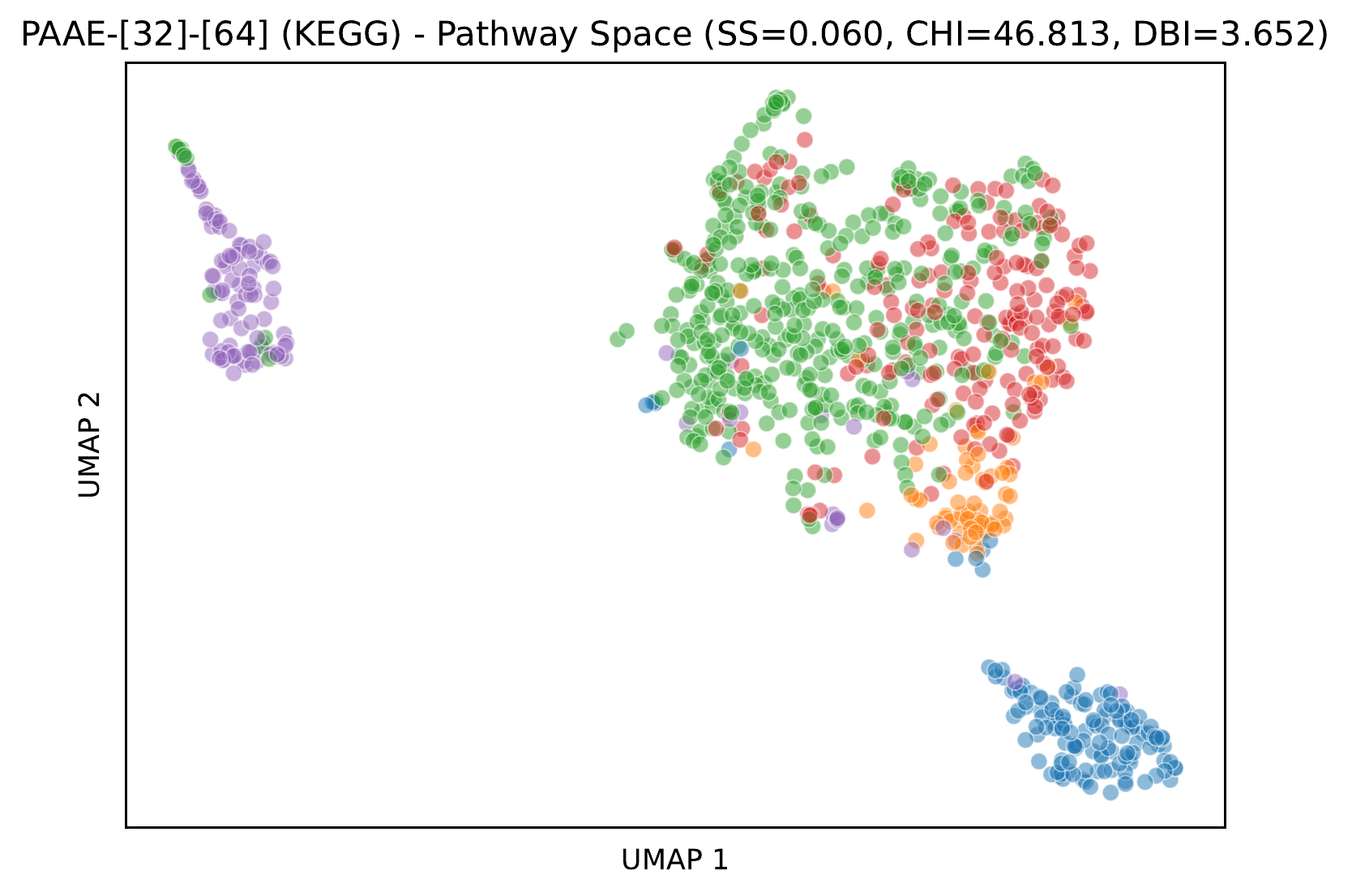}}
    \subcaptionbox{PAAE - $z$\label{fig:featspace:sub:paae-latent}}{\includegraphics[width=.3\textwidth]{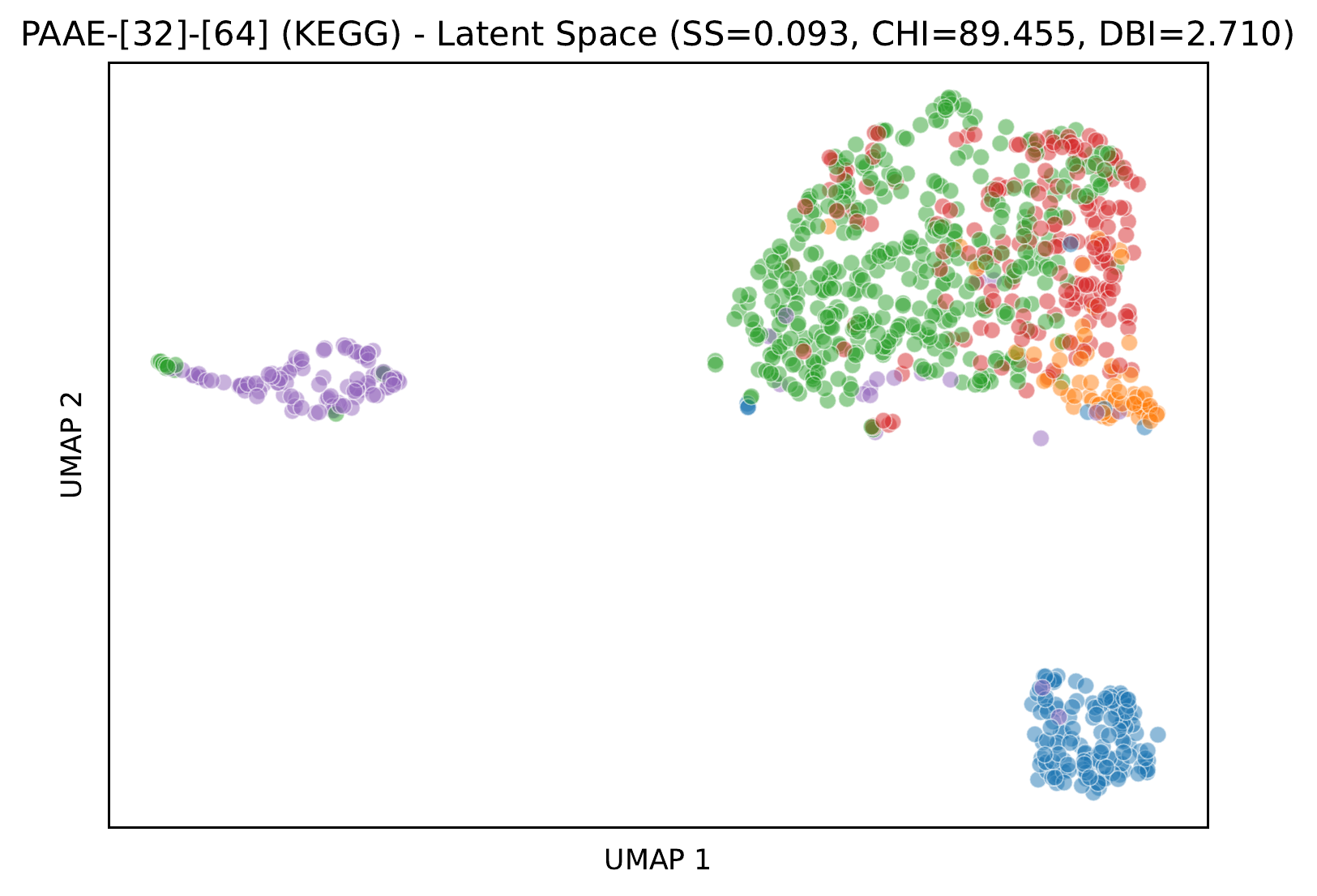}}
    \\
    \subcaptionbox{VAE - $\mu$\label{fig:featspace:sub:vae}}{\includegraphics[width=.3\textwidth]{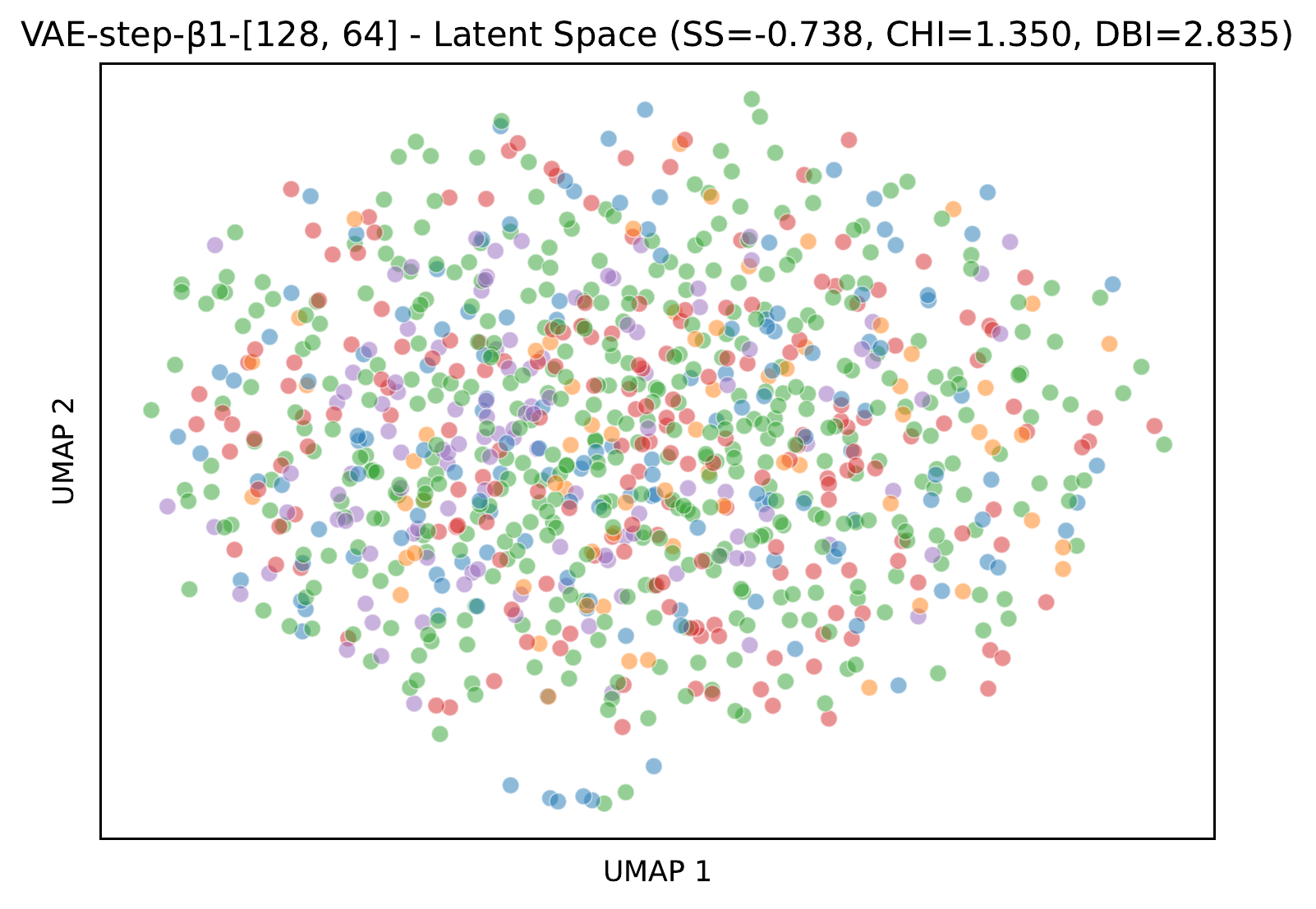}}
    \hfill
    \subcaptionbox{PAVAE - $a$\label{fig:featspace:sub:pavae-pathway}}{\includegraphics[width=.3\textwidth]{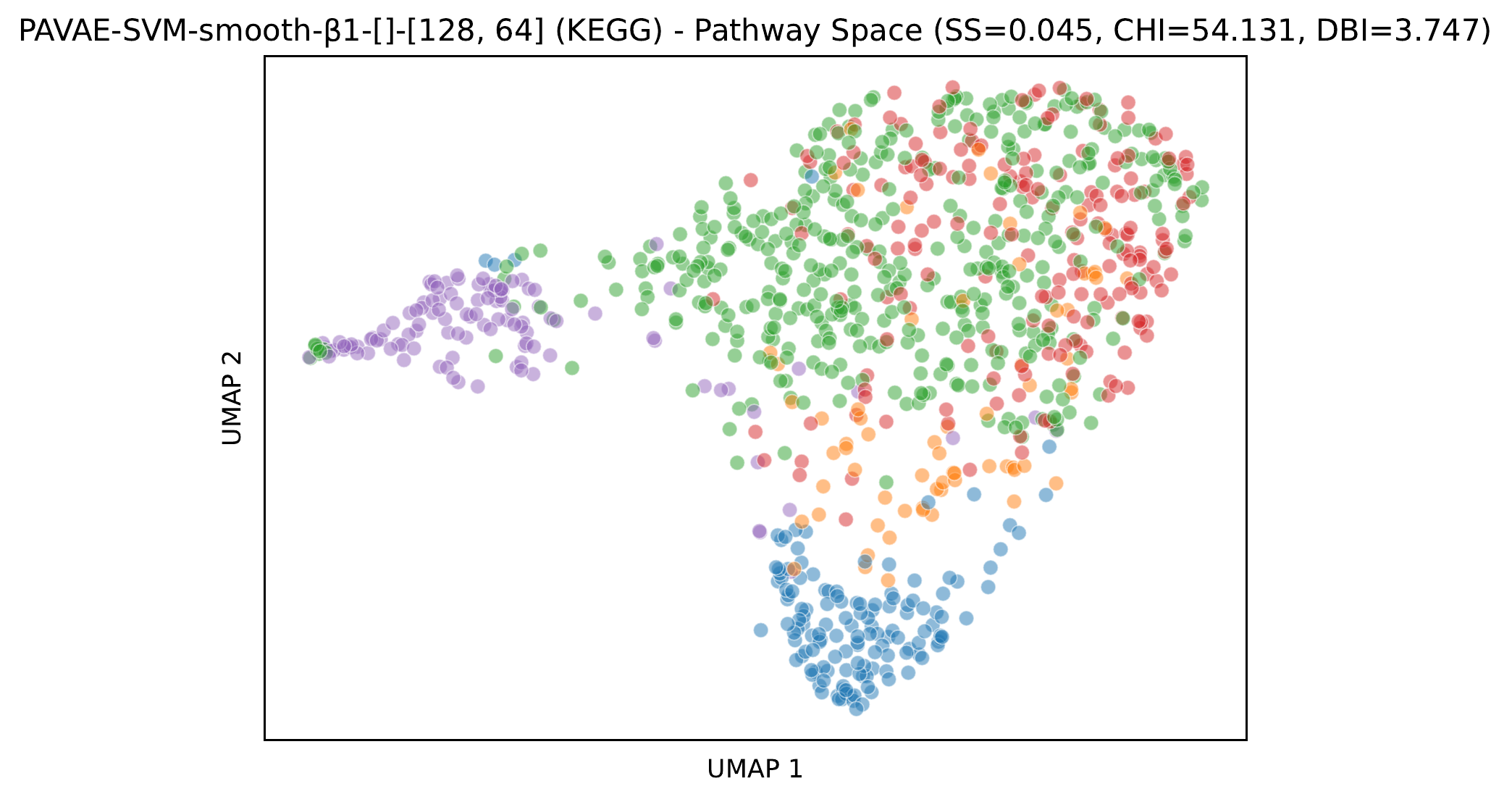}}
    \subcaptionbox{PAVAE - $\mu$\label{fig:featspace:sub:pavae-latent}}{\includegraphics[width=.3\textwidth]{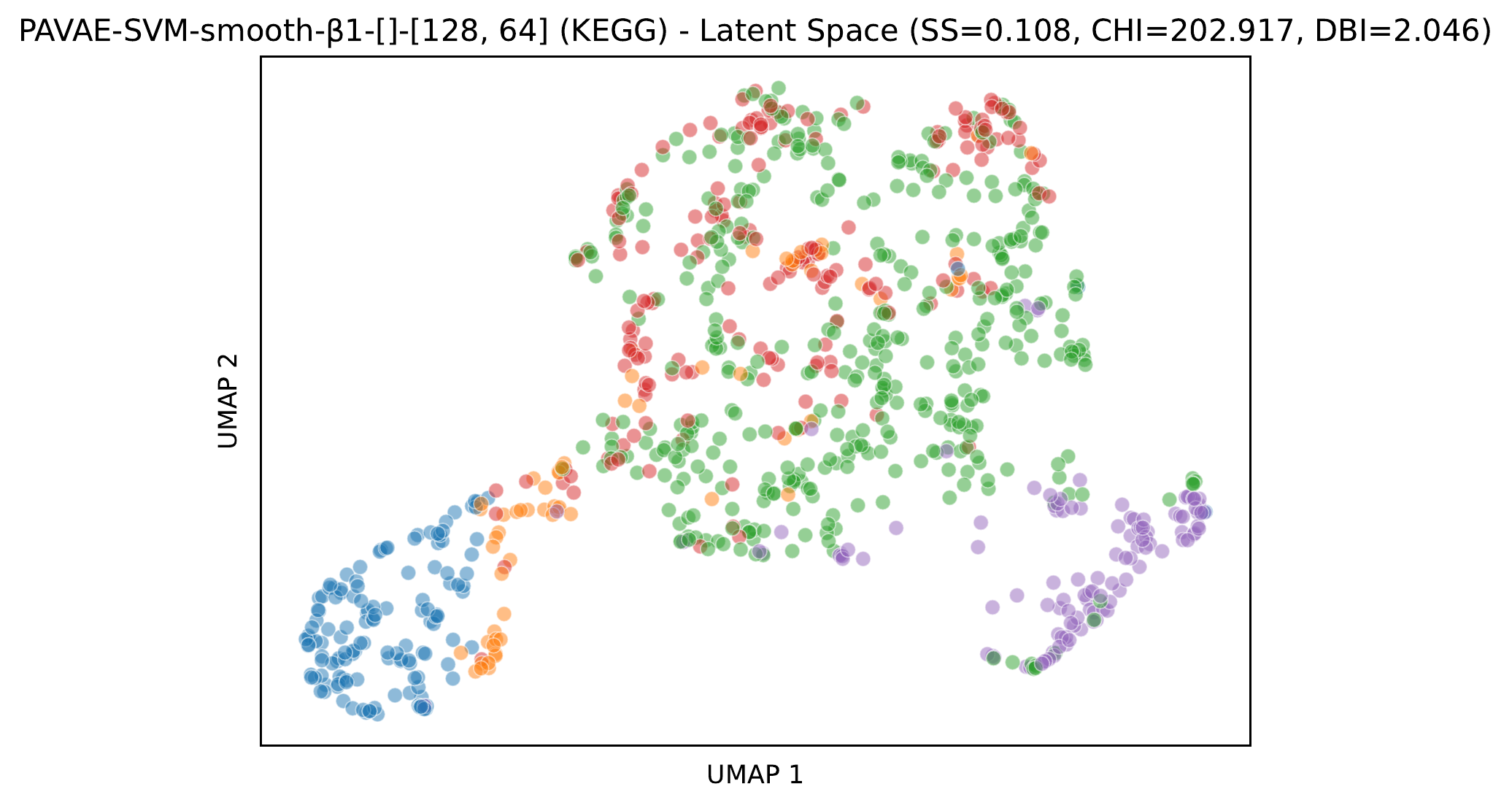}}
    \caption{The 2-dimensional UMAP reduction, for the TCGA-BRCA dataset, of (\subref{fig:featspace:sub:ae}) a standard AutoEncoder, the pathway activities inferred by (\subref{fig:featspace:sub:paae-pathway}) and (\subref{fig:featspace:sub:paae-latent}) latent space learned by our PAAE model , (\subref{fig:featspace:sub:vae}) the standard Variational AutoEncoder, and the pathway activities inferred by (\subref{fig:featspace:sub:pavae-pathway}) and (\subref{fig:featspace:sub:pavae-latent}) latent space learned by our PAVAE model.
    We used the Hallmark Genes pathway set for our pathway activity models in this plot.
    The colours mark each of the 5-classes.}
    \label{fig:featspace}
\end{figure*}

\subsubsection{Interpreting the Pathway Activity Scores}

As we can see in Eq.~\ref{eq:paae}, the pathway activity space that we visualise is built of the individual pathway activity scores of each pathway, described in Eq.~\ref{eq:paae-single-activity}. Each of the pathway activity scores is built by stacking fully connected layers using the normalised gene expression of the genes belonging to a specific pathway as an input. The encoder in Eq.~\ref{eq:paae-single-activity}, if it has $k$ layers, can be described as in Eq.~\ref{eq:paae-activity-encoder}:

\begin{equation}\label{eq:paae-activity-encoder}
\footnotesize
\begin{split}
    a_{j} = E_{p_{j}}(x) &= h_{j,k} \\
    h_{j,k} &= W_{j,k} \times h_{j,k-1} + b_{j,k} \\
    h_{j,i} &= f_{j,k}(W_{j,i} \times h_{j,i-1} + b_{j,i}) \\
    h_{j,1} &= f_{j,k}(W_{j,1} \times x_{:,p_{j}} + b_{j,1})
\end{split}
\end{equation}

Where $W_{j,i}$ is a $\Re^{d_{j,i-1} \times d_{j,i}}$ matrix, and $f_{j,k}$ is a nonlinear function, not present in the last layer so that our output is roughly normal-shaped.

Since the weights $W_{j,i}$ can be both positive and negative, one should not expect the final result $a_{j}$ to have a positive/negative directionality w.r.t. the pathway's input $x_{:,p_{j}}$, and one should always consider the pathway activity scores in terms of \textbf{contrasts} between samples.

\begin{figure*}
    \centering
    \subcaptionbox{TCGA (train)\label{fig:clustermap-kegg:sub:tcga}}{\includegraphics[width=.45\linewidth]{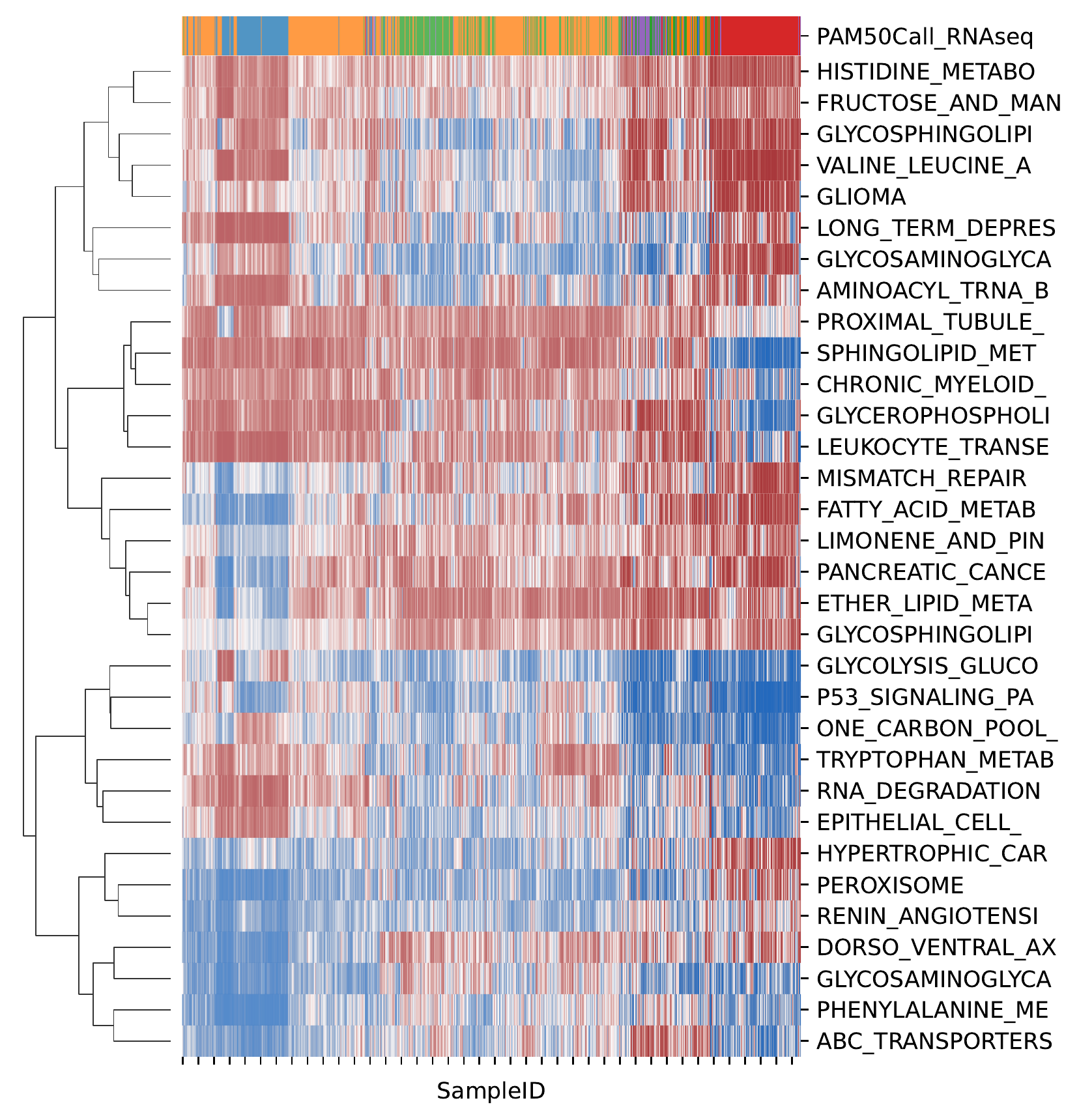}}
    \hfill
    \subcaptionbox{Metabric (test)\label{fig:clustermap-kegg:sub:meta}}{\includegraphics[width=.45\linewidth]{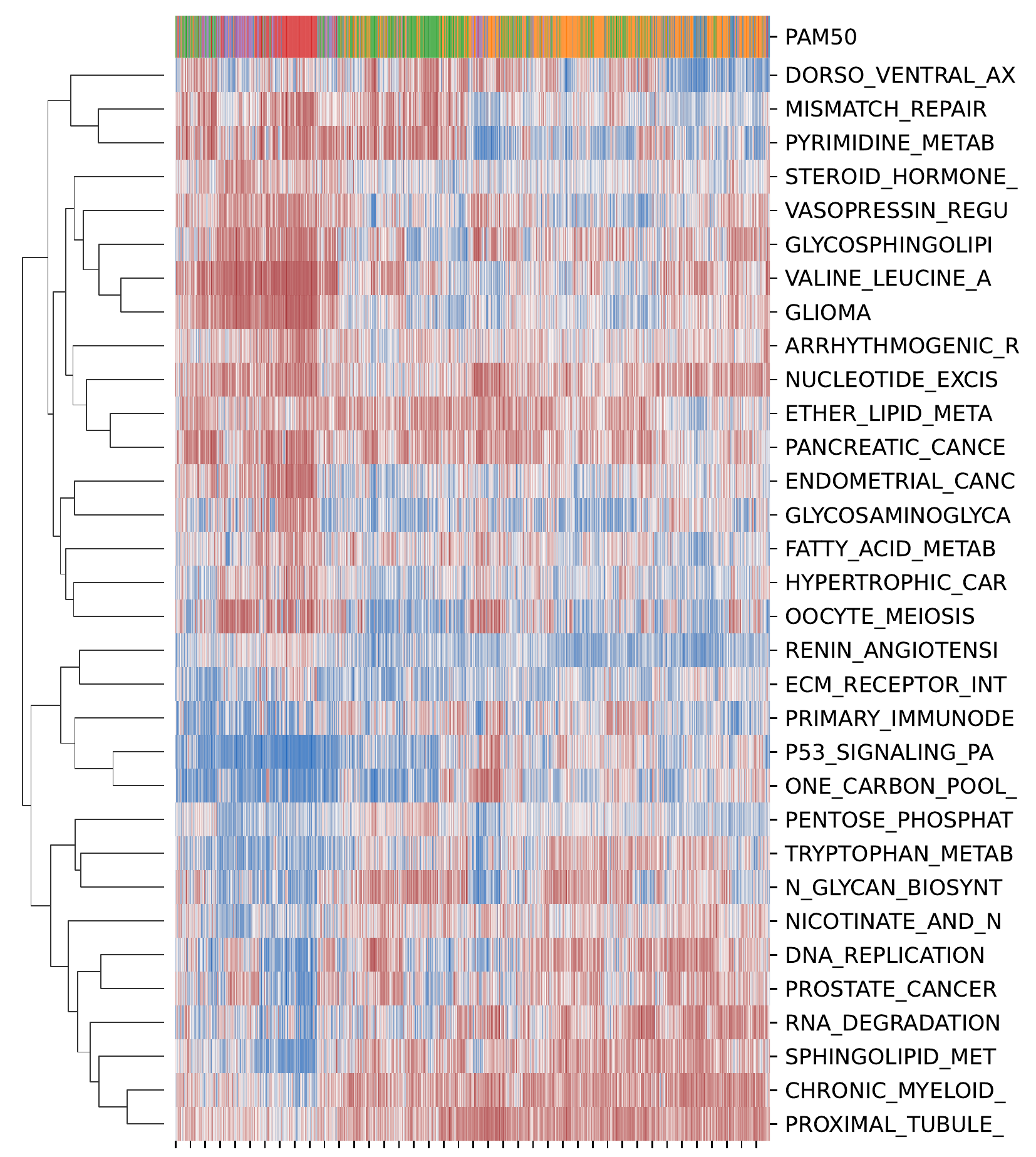}}
    
    \caption{The clustermap using the cosine distance between samples' inferred pathway activity vectors for KEGG PAAE's pathway activity space with the colours marking BRCA's 5-classes: {\color{brca-normal}Normal in blue}, {\color{brca-luma}Luminal A in orange}, {\color{brca-lumb}Luminal B in green}, {\color{brca-basal}Basal in red}, {\color{brca-her2}Her2 in purple}.}
    \label{fig:clustermap-kegg}
\end{figure*}

\begin{figure*}
    \centering
    \subcaptionbox{TCGA (train)\label{fig:clustermap-hallmark:sub:tcga}}{\includegraphics[width=.49\linewidth]{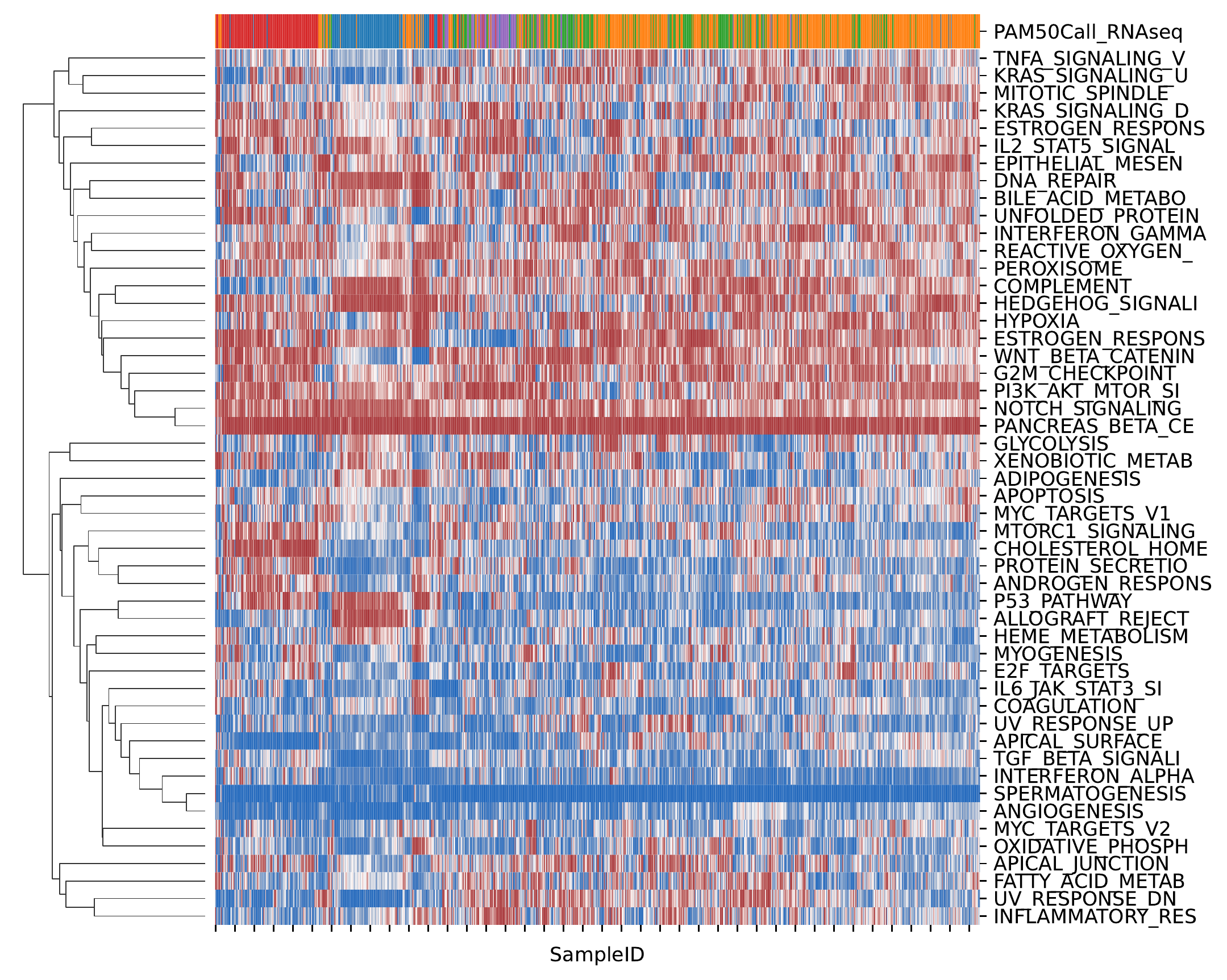}}
    \hfill
    \subcaptionbox{Metabric (test)\label{fig:clustermap-hallmark:sub:meta}}{\includegraphics[width=.49\linewidth]{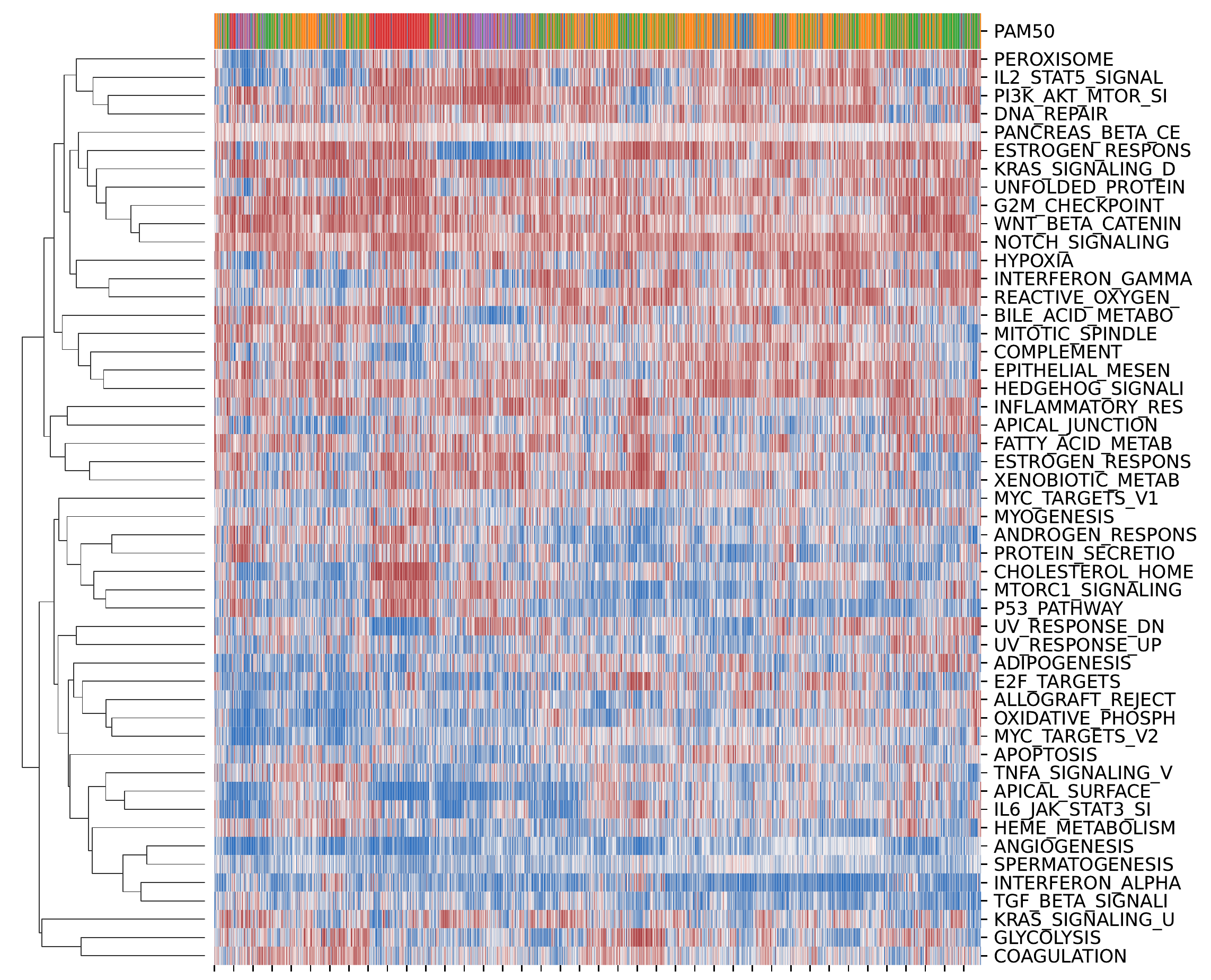}}
    
    \caption{The clustermap using the cosine distance between samples' inferred pathway activity vectors for Hallmark Genes PAAE's pathway activity space with the colours marking BRCA's 5-classes: {\color{brca-normal}Normal in blue}, {\color{brca-luma}Luminal A in orange}, {\color{brca-lumb}Luminal B in green}, {\color{brca-basal}Basal in red}, {\color{brca-her2}Her2 in purple}.}
    \label{fig:clustermap-hallmark}
\end{figure*}

\begin{figure*}
    \centering
    \subcaptionbox{TCGA (train) \label{fig:featuremap-tcga-hallmark:sub:class}}{\includegraphics[width=.3\linewidth]{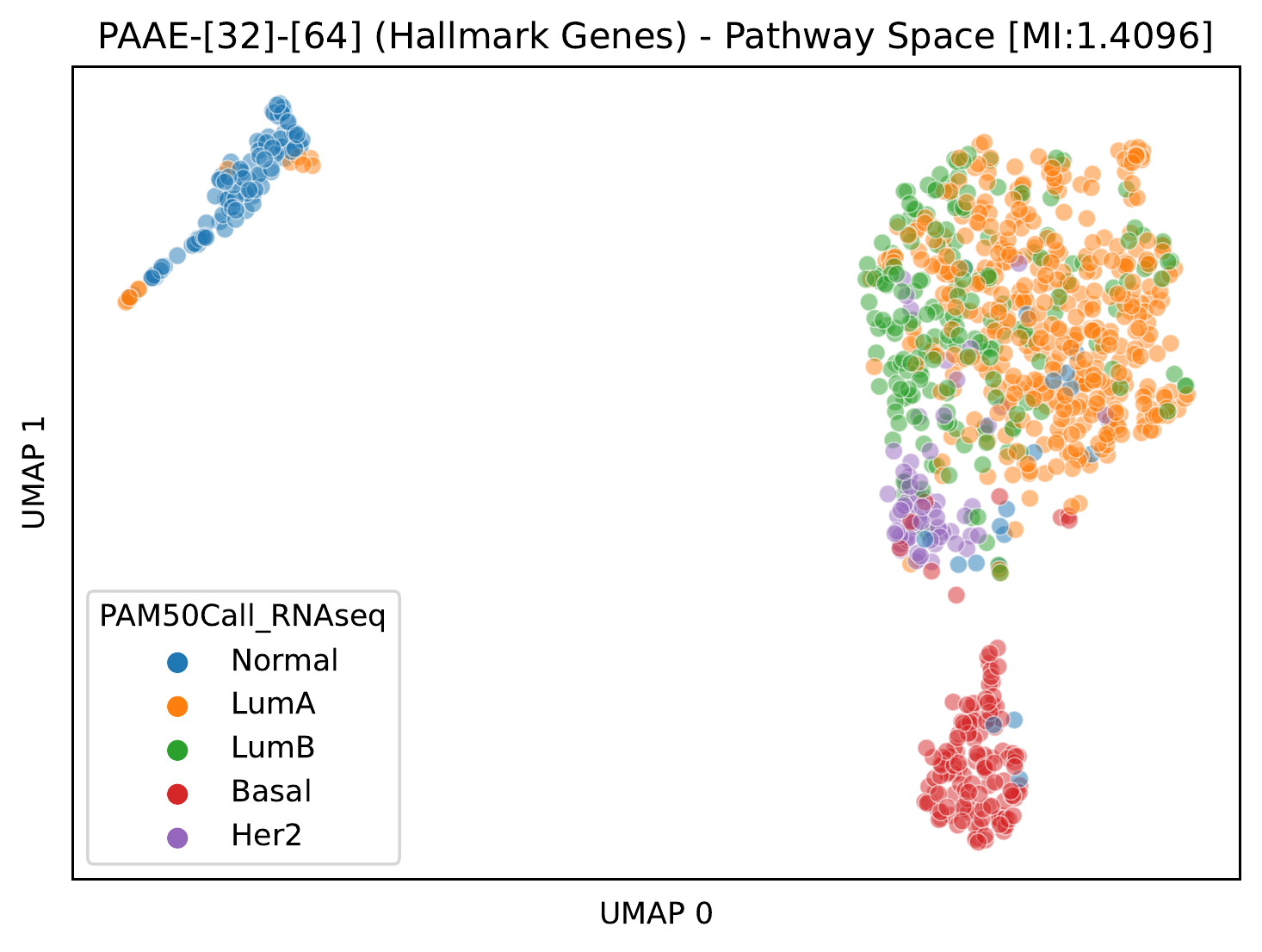}}
    \hfill
    \subcaptionbox{\tiny P53\_PATHWAY  \label{fig:featuremap-tcga-hallmark:sub:P53_PATHWAY}}{\includegraphics[width=.3\linewidth]{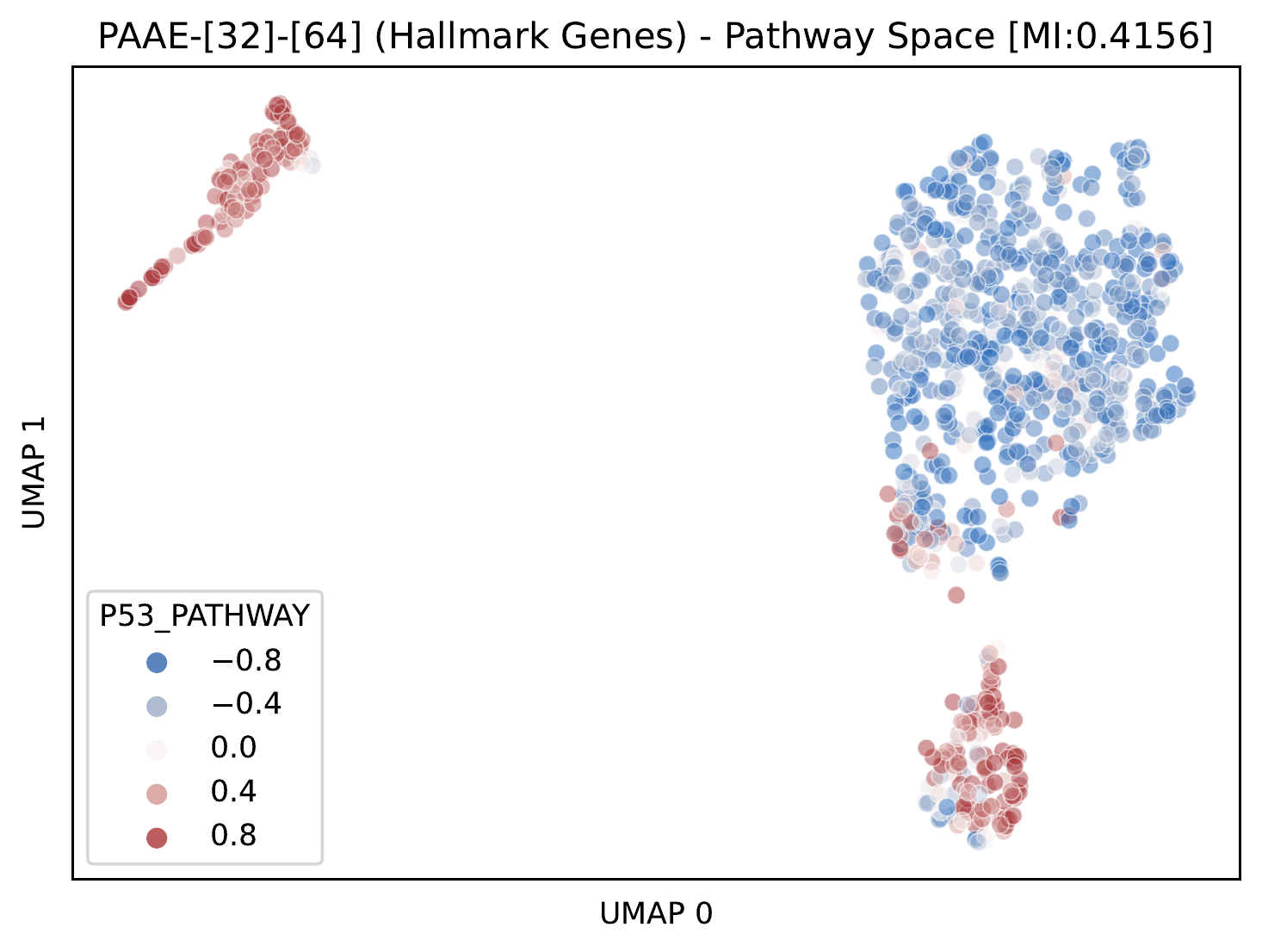}}
    \hfill
    \subcaptionbox{\tiny UV\_RESPONSE\_DN  \label{fig:featuremap-tcga-hallmark:sub:UV_RESPONSE_DN}}{\includegraphics[width=.3\linewidth]{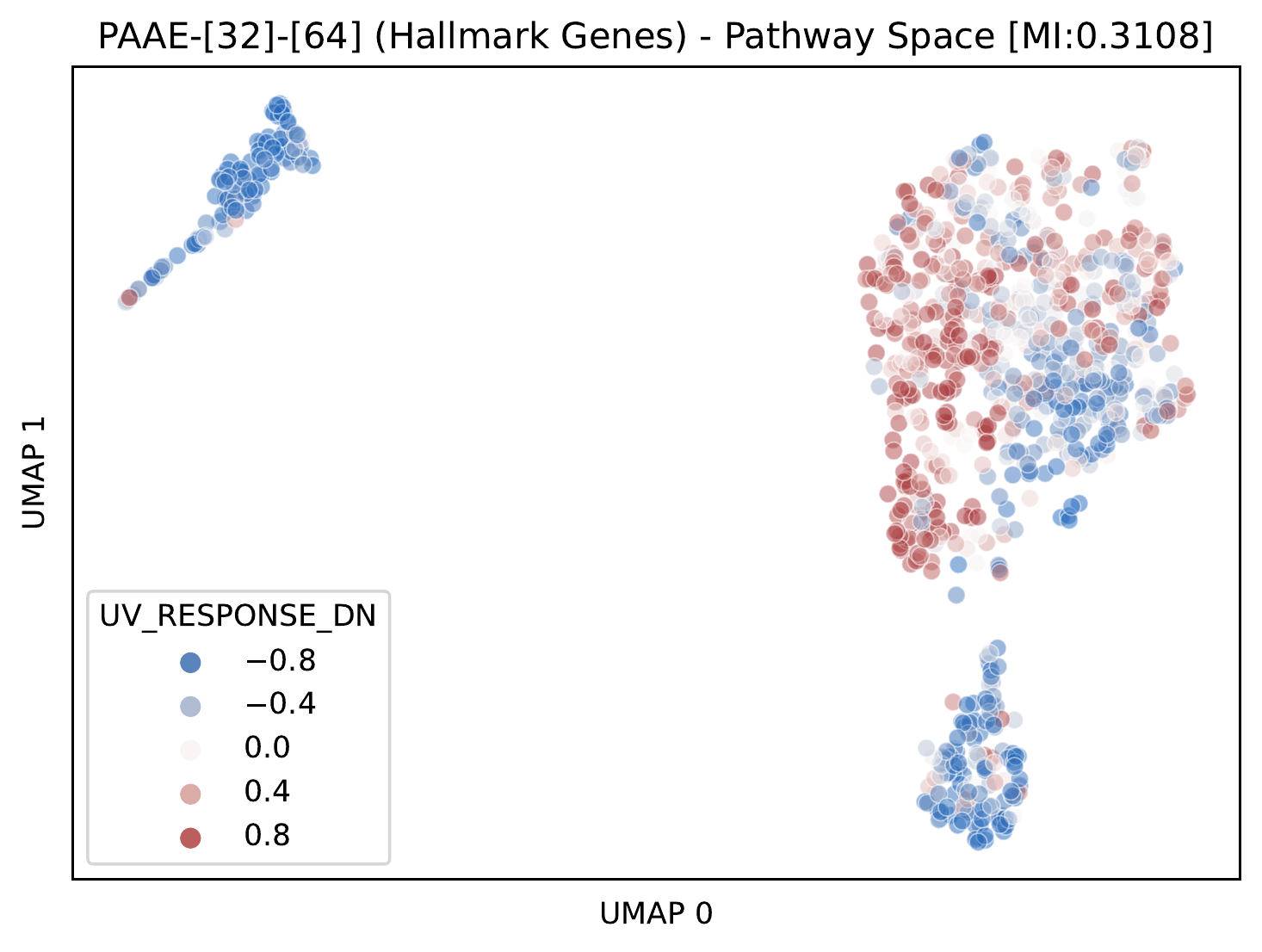}}
    \\
    \subcaptionbox{\tiny CHOLESTEROL\_HOMEOSTASIS  \label{fig:featuremap-tcga-hallmark:sub:CHOLESTEROL_HOMEOSTASIS}}{\includegraphics[width=.3\linewidth]{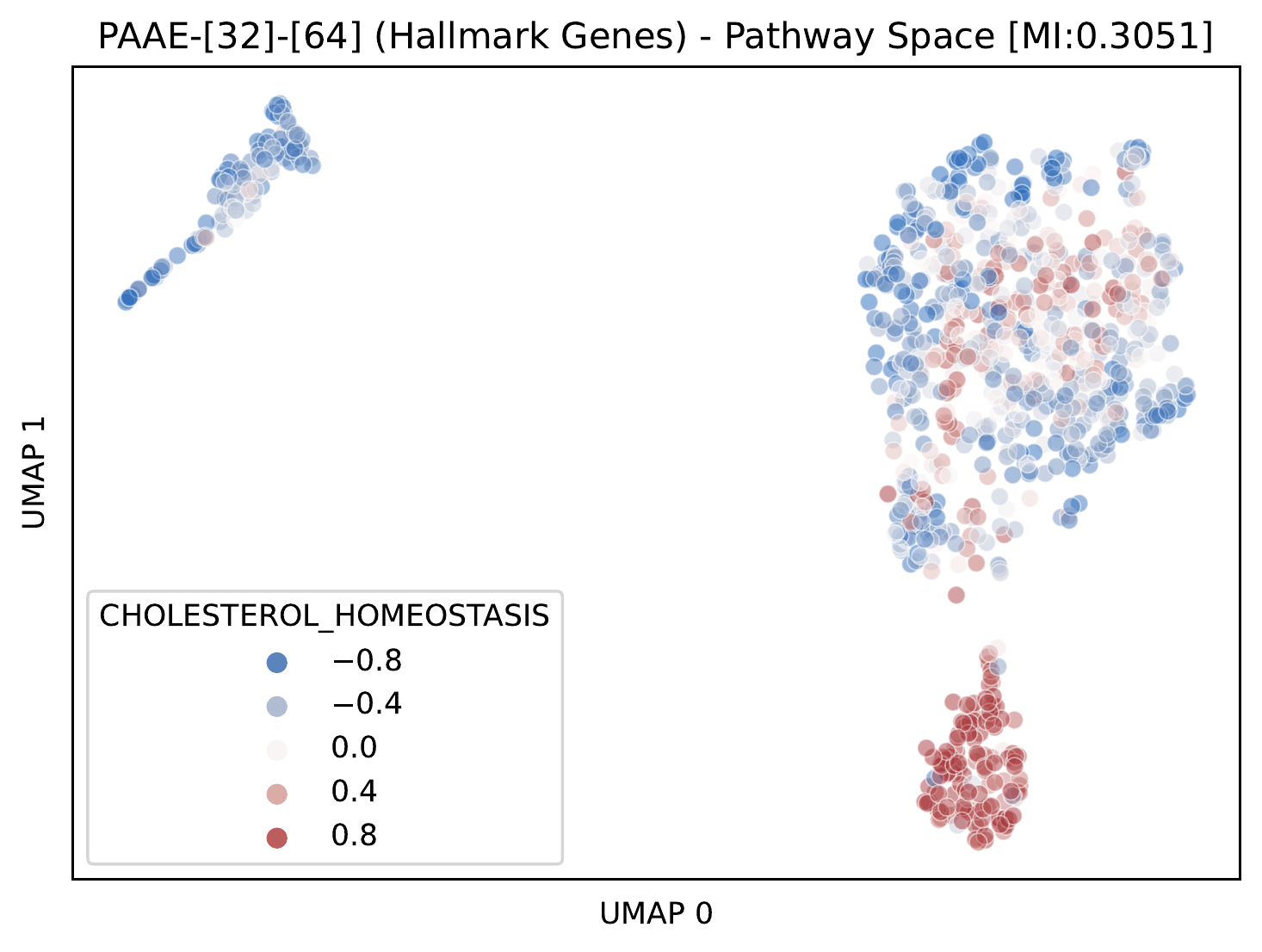}}
    \hfill
    \subcaptionbox{\tiny APICAL\_SURFACE  \label{fig:featuremap-tcga-hallmark:sub:APICAL_SURFACE}}{\includegraphics[width=.3\linewidth]{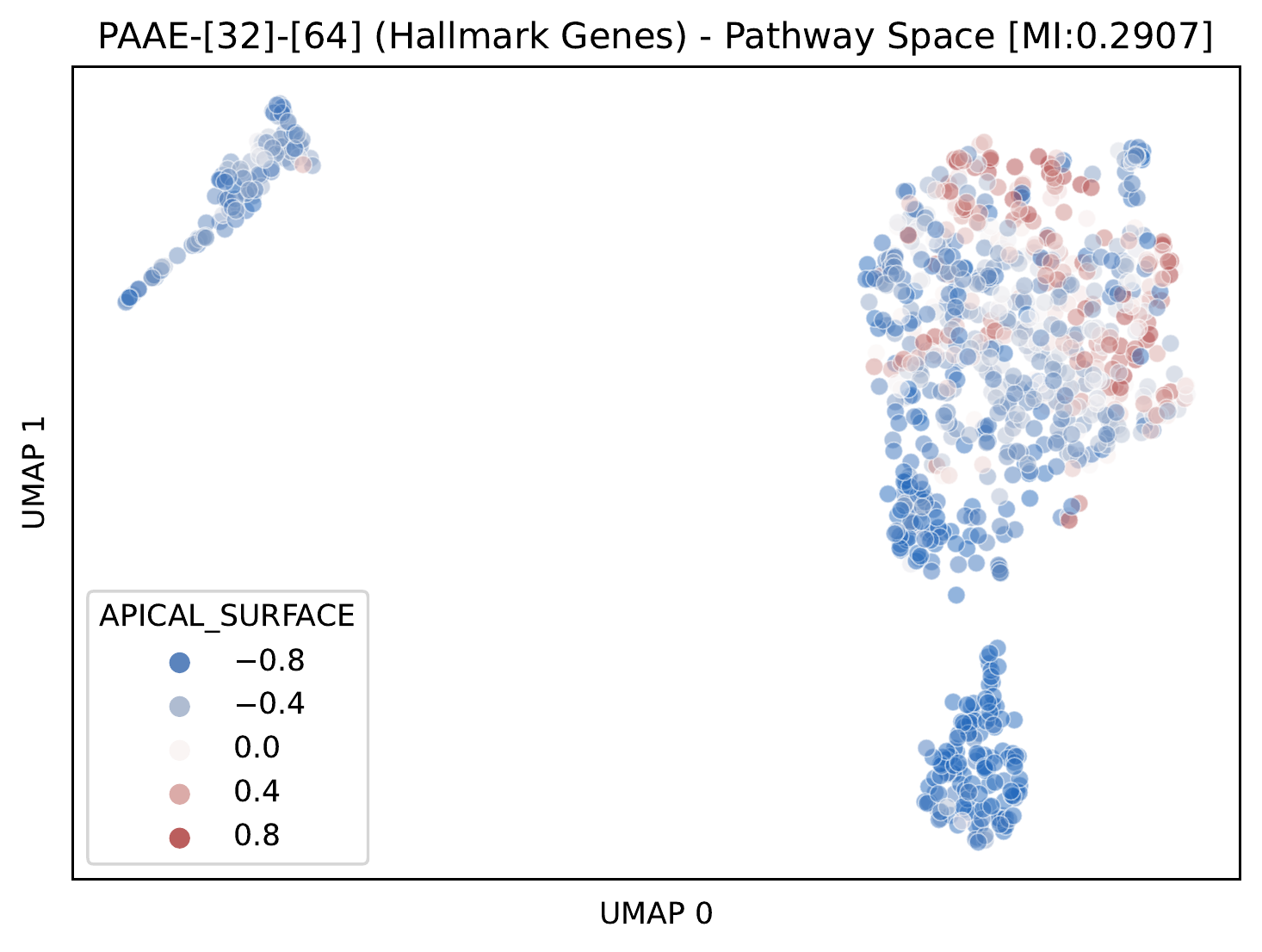}}
    \hfill
    \subcaptionbox{\tiny MTORC1\_SIGNALING  \label{fig:featuremap-tcga-hallmark:sub:MTORC1_SIGNALING}}{\includegraphics[width=.3\linewidth]{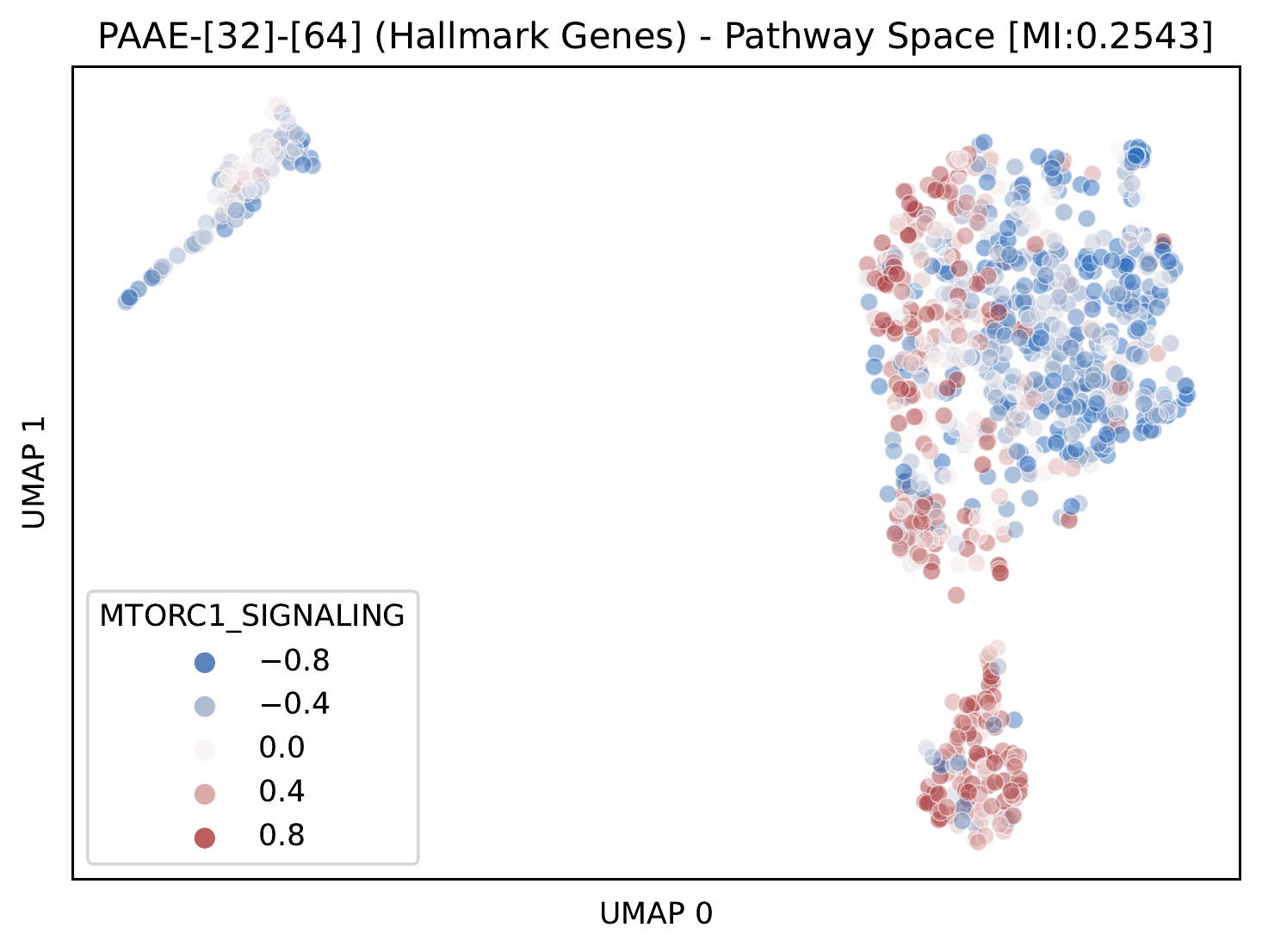}}
    
    \caption{The featuremap showing the intensity of each sample's inferred pathway activity vectors for Halmark Genes PAAE's pathway activity space in the TCGA dataset, as well as the class distribution overlayed on top of the 2-dimensional UMAP reduction: {\color{brca-normal}Normal in blue}, {\color{brca-luma}Luminal A in orange}, {\color{brca-lumb}Luminal B in green}, {\color{brca-basal}Basal in red}, {\color{brca-her2}Her2 in purple}. We only show here pathways that are among the top 5 with highest the most mutual information w.r.t. the classes on either TCGA or Metabric.}
    \label{fig:featuremap-tcga-hallmark}
\end{figure*}

\begin{figure*}
    \centering
    \subcaptionbox{Metabric (test) \label{fig:featuremap-meta-hallmark:sub:class}}{\includegraphics[width=.3\linewidth]{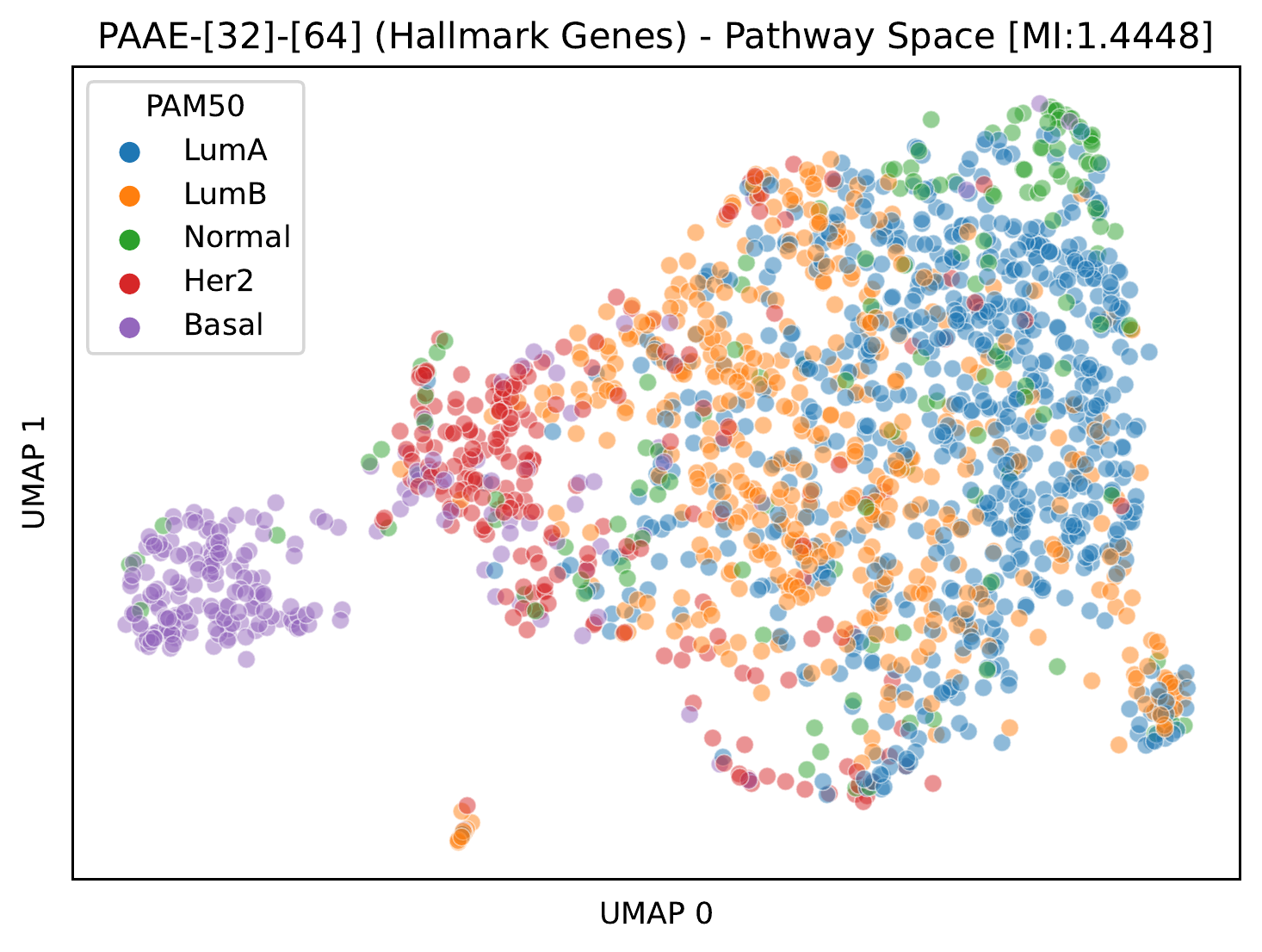}}
    \hfill
    \subcaptionbox{\tiny P53\_PATHWAY  \label{fig:featuremap-meta-hallmark:sub:P53_PATHWAY}}{\includegraphics[width=.3\linewidth]{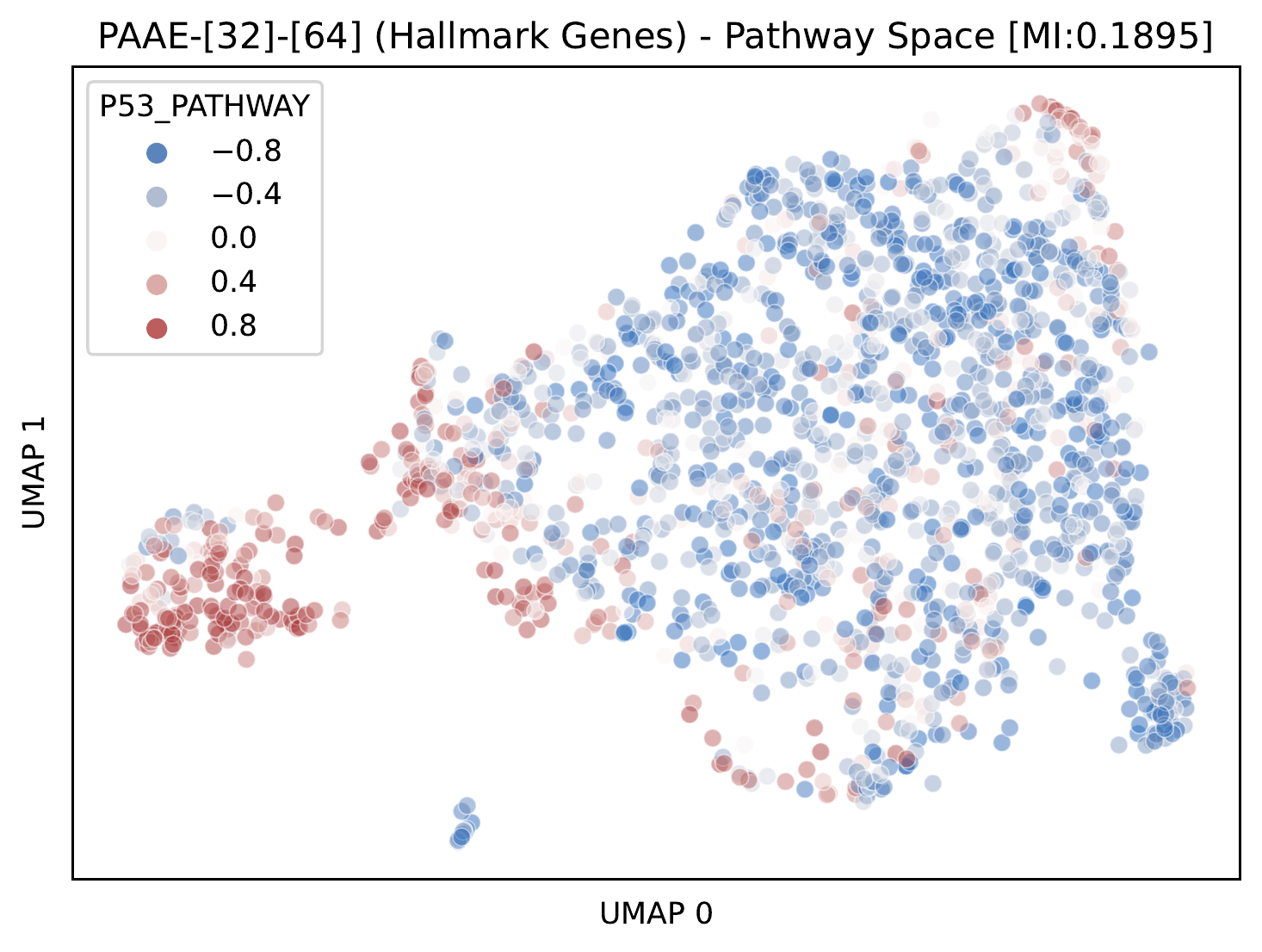}}
    \hfill
    \subcaptionbox{\tiny UV\_RESPONSE\_DN  \label{fig:featuremap-meta-hallmark:sub:UV_RESPONSE_DN}}{\includegraphics[width=.3\linewidth]{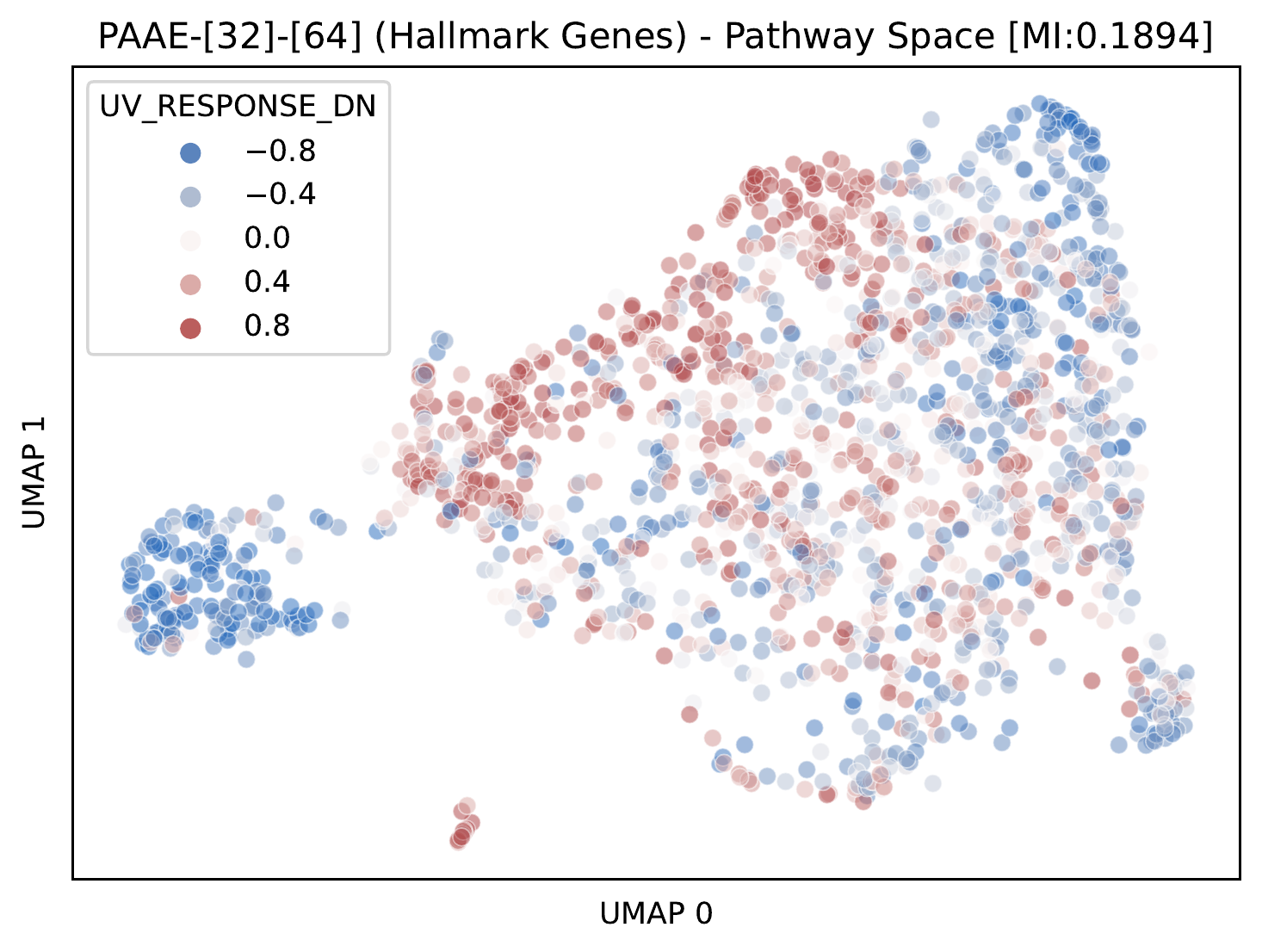}}
    \\
    \subcaptionbox{\tiny CHOLESTEROL\_HOMEOSTASIS  \label{fig:featuremap-meta-hallmark:sub:CHOLESTEROL_HOMEOSTASIS}}{\includegraphics[width=.3\linewidth]{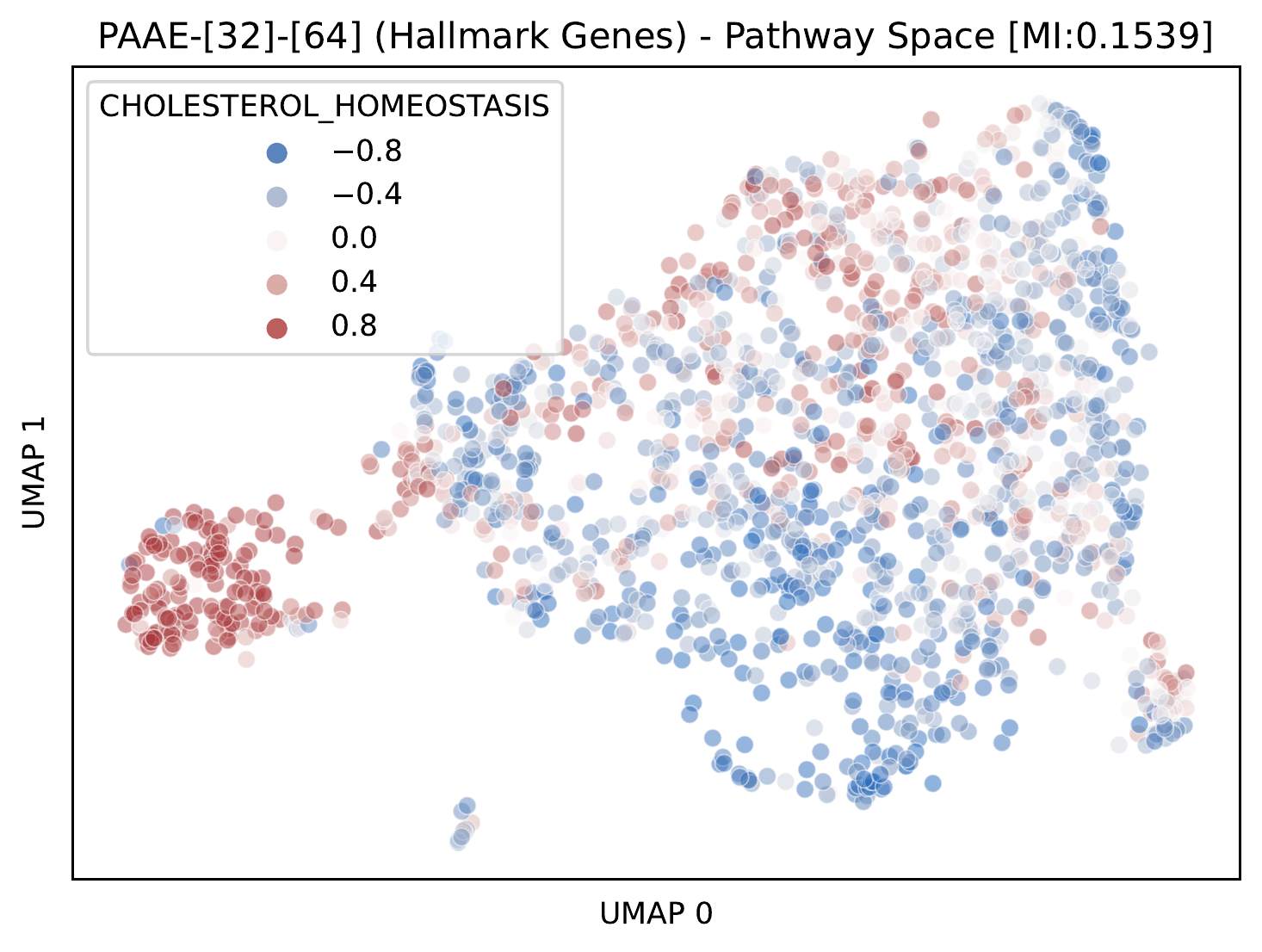}}
    \hfill
    \subcaptionbox{\tiny APICAL\_SURFACE  \label{fig:featuremap-meta-hallmark:sub:APICAL_SURFACE}}{\includegraphics[width=.3\linewidth]{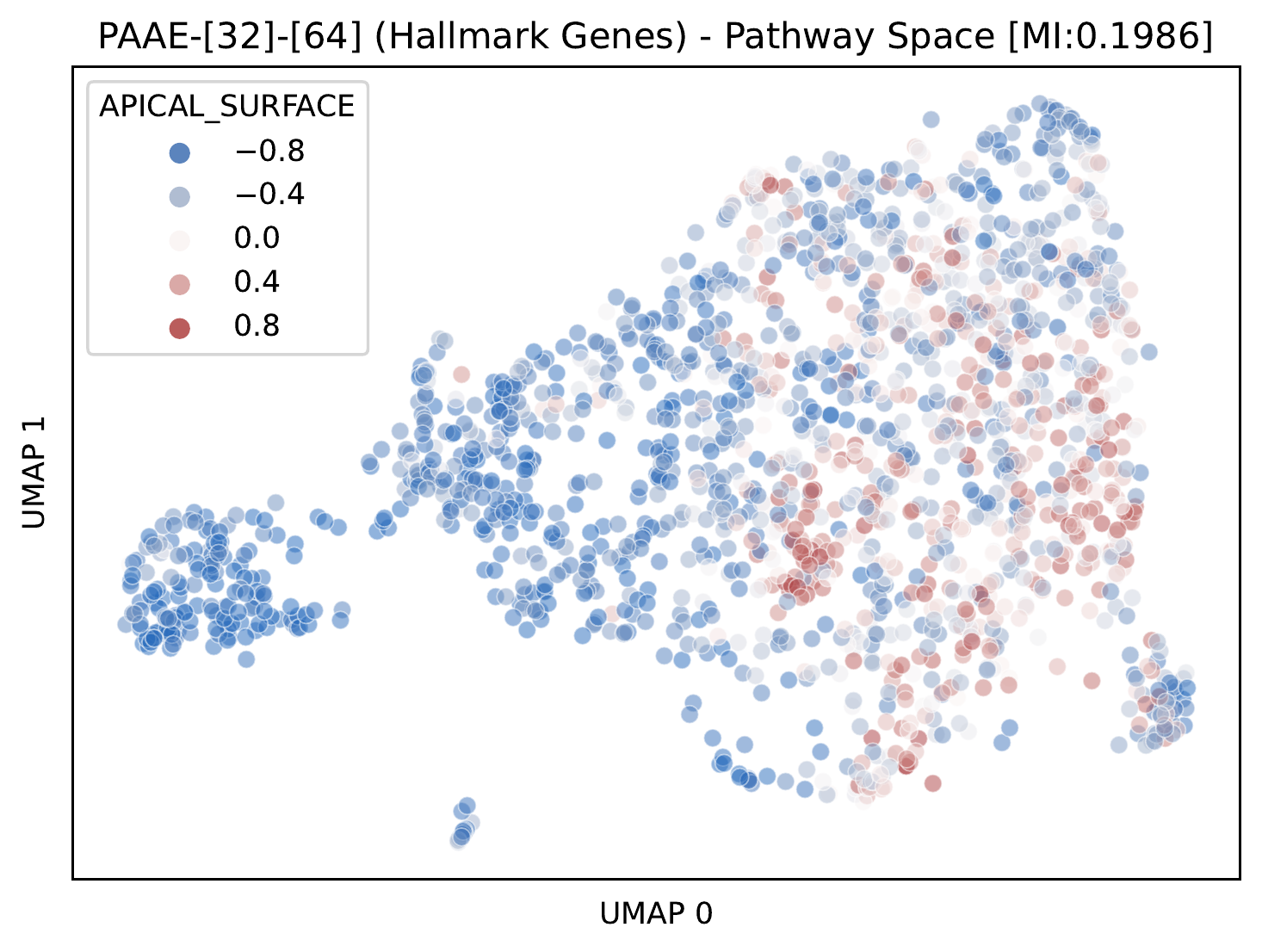}}
    \hfill
    \subcaptionbox{\tiny MTORC1\_SIGNALING  \label{fig:featuremap-meta-hallmark:sub:MTORC1_SIGNALING}}{\includegraphics[width=.3\linewidth]{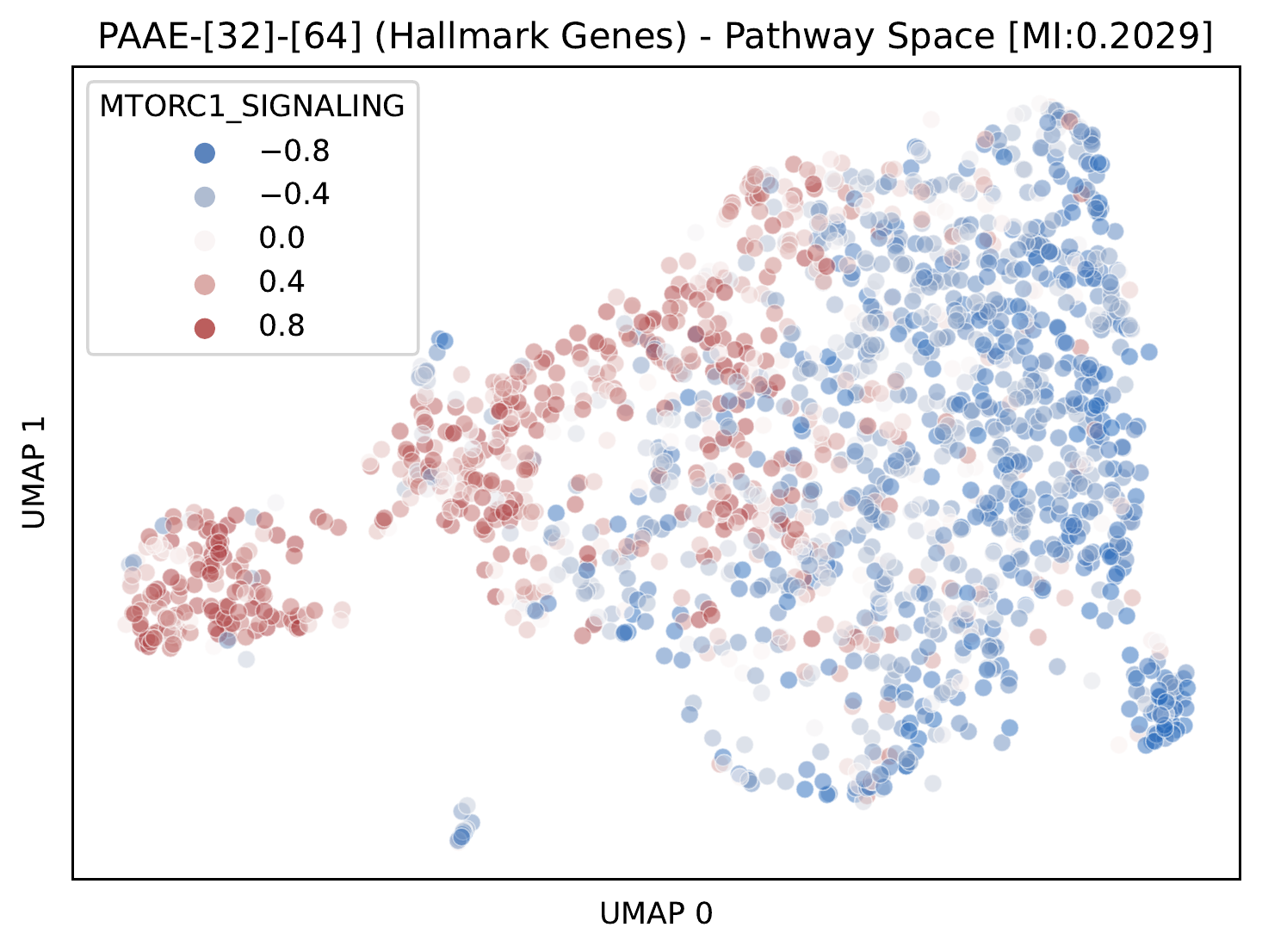}}
    
    \caption{The featuremap showing the intensity of each sample's inferred pathway activity vectors for Halmark Genes PAAE's pathway activity space in the Metabric dataset, as well as the class distribution overlayed on top of the 2-dimensional UMAP reduction: {\color{brca-lumb}Normal in green}, {\color{brca-normal}Luminal A in blue}, {\color{brca-luma}Luminal B in orange}, {\color{brca-her2}Basal in purple}, {\color{brca-basal}Her2 in red}. We only show here pathways that are among the top 5 with highest the most mutual information w.r.t. the classes on either TCGA or Metabric.}
    \label{fig:featuremap-meta-hallmark}
\end{figure*}

The pathway activity score $a_{j}$ is the value that we use in both our heatmap/clustermap (Figs.~\ref{fig:clustermap-kegg}, \ref{fig:clustermap-hallmark}) and featuremap (Figs.~\ref{fig:featuremap-tcga}, \ref{fig:featuremap-meta}, \ref{fig:featuremap-tcga-hallmark}, \ref{fig:featuremap-meta-hallmark}) visualisations, and thus both these visualisations should be interpreted with the caveat above. Furthermore, in cases where nonlinear layers are involved in the calculation of the pathway activity score $a_{j}$, one should not expect the values to behaviour linearly.

Furthermore, we can get the most important features for each pathway using either neural path weights (NPW, Eq.~\ref{eq:npw}) or absolute neural path weight (ANPW, Eq.~\ref{eq:anpw}) \cite{uyar_multi-omics_2021}.

\begin{equation}\label{eq:npw}
\footnotesize
    \operatorname{NPW}_{j} = \prod_{i=1}^{k} W_{j,i} ~,~ \operatorname{NPW}_{j} \in \Re^{|p_{j}|}
\end{equation}

\begin{equation}\label{eq:anpw}
\footnotesize
    \operatorname{ANPW}_{j}[g] = |\operatorname{NPW}_{j}[g]|
\end{equation}

Note, however, that since our models might compute the pathway activity score value through a combination of nonlinear functions, these values should not be interpreted as a linear influence on the value, and combinations with other values may suppress or excite the score more than simply increasing/decreasing the genes with highest ANPW values.

Also note that one can do this interpretability technique for any point in any neural network, but we only define this for our pathway activity score for simplicity sake.

\subsubsection{KEGG}

For a neural network we produced when training with the KEGG pathway set, we found that the 5 pathways with most mutual information w.r.t. the PAM50 subtype in the TCGA dataset were GLYCOSAMINOGLYCAN\_BIOSYNTHESIS\_KERATAN\_SULFATE, SPHINGOLIPID\_METABOLISM, VALINE\_LEUCINE\_AND\_ISOLEUCINE\_DEGRADATION, P53\_SIGNALING\_PATHWAY, and PANCREATIC\_CANCER, as well as GLIOMA and MISMATCH REPAIR, which were on the top 5 for the Metabric dataset, but ranked 8 and 15, respectively, on the TCGA dataset. For each of these pathways, we calculated the 10 genes with the highest ANPW, which can be seen in Tab. ~\ref{tab:important-genes-kegg}, where
$^{\ast}$ indicates that the gene was a match on OncoKB \cite{chakravarty_oncokb_2017}.

\begin{figure*}
    \centering
    \subcaptionbox{TCGA (train)\label{fig:clustermap-kegg-importantgenes:sub:tcga}}{\includegraphics[width=.49\linewidth]{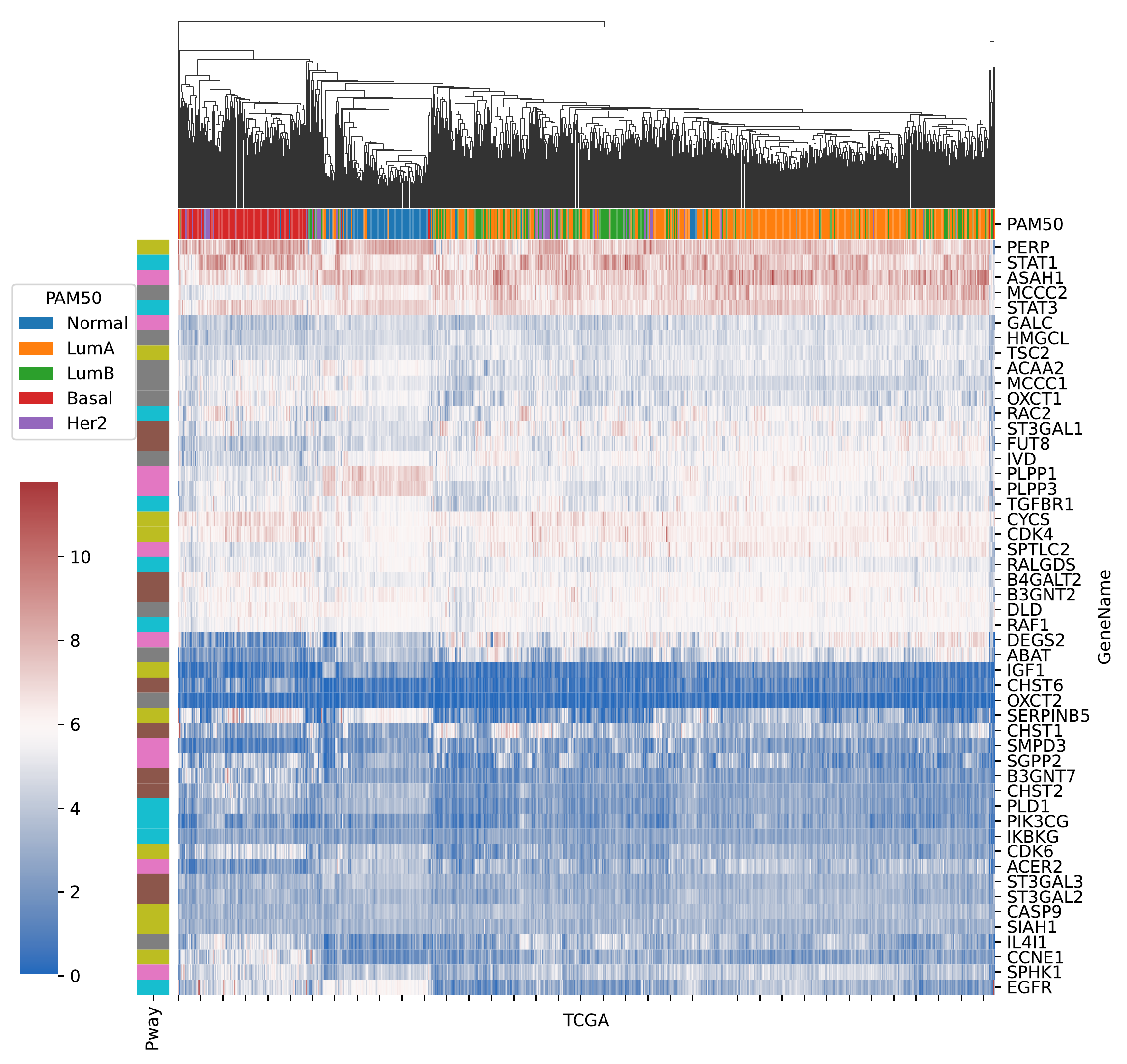}}
    \hfill
    \subcaptionbox{Metabric (test)\label{fig:clustermap-kegg-importantgenes:sub:meta}}{\includegraphics[width=.49\linewidth]{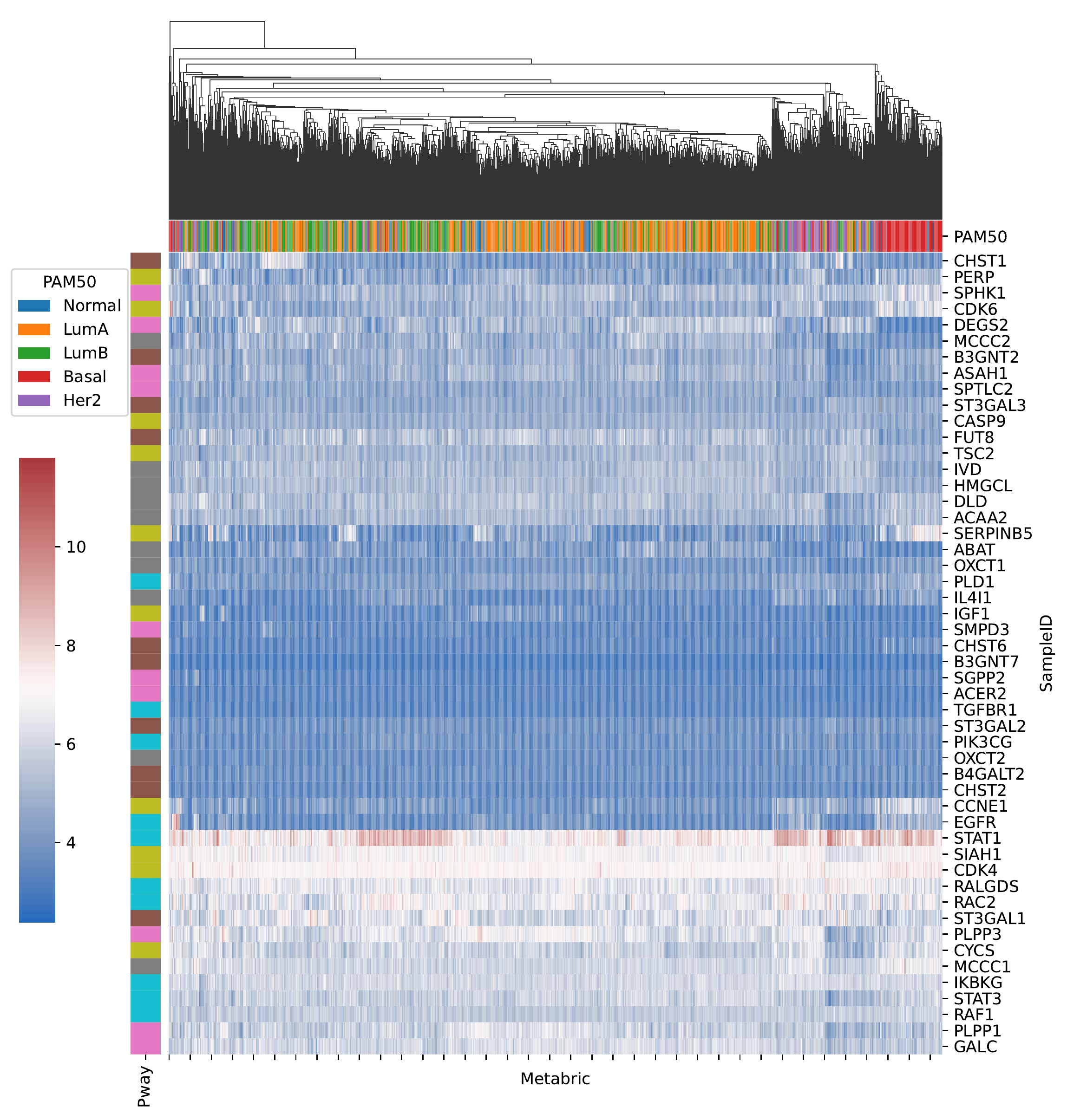}}
    
    \caption{The clustermap using the euclidean distance between samples' log2p1 TPM/IPM gene expression vectors for KEGG PAAE's pathway activity space with the colours marking BRCA's 5-classes: {\color{brca-normal}Normal in blue}, {\color{brca-luma}Luminal A in orange}, {\color{brca-lumb}Luminal B in green}, {\color{brca-basal}Basal in red}, {\color{brca-her2}Her2 in purple}.}
    \label{fig:clustermap-kegg-importantgenes}
\end{figure*}

Plotting the clustermaps of these 50 features in Fig.~\ref{fig:clustermap-kegg-importantgenes}, we can see that there is good separation of almost all classes on both datasets, with Normal class samples not being as separated in the test Metabric dataset as it is on the training TCGA dataset.

\renewcommand*{\thefootnote}{\alph{footnote}}
\begin{table}[]
    \centering
    \scriptsize
    \caption{Top 10 Genes w.r.t ANPW for the 5 pathways with highest MI w.r.t. the classification target for the KEGG pathway set.}
    \label{tab:important-genes-kegg}
    \begin{tabular}{ccccccccccc}
    \toprule
    \multirow{2}{*}{PATHWAY} & \multicolumn{10}{c}{Gene names} \\
    & \multicolumn{10}{c}{Neural Path Weights} \\
    \midrule
    \multirow{2}{*}{GBKS\tablefootnote{GLYCOSAMINOGLYCAN BIOSYNTHESIS KERATAN SULFATE}}
     & ST3GAL3 & FUT8 & B4GALT2 & CHST6 & ST3GAL2 & CHST1 & CHST2 & B3GNT2 & ST3GAL1 & B3GNT7 \\
     & +0.62 & -0.52 & +0.49 & +0.48 & +0.41 & -0.39 & +0.29 & +0.26 & -0.23 & +0.22 \\
    \midrule
    SPHINGOLIPID
     & SPTLC2 & DEGS2 & ACER2 & SMPD3 & PLPP1 & GALC & SGPP2 & PLPP3 & ASAH1 & SPHK1 \\
    METABOLISM
     & +0.40 & +0.39 & +0.32 & +0.30 & +0.30 & +0.30 & -0.29 & +0.27 & +0.26 & -0.24 \\
    \midrule
    VLI
     & MCCC1 & IVD & ABAT & DLD & ACAA2 & OXCT2 & OXCT1 & HMGCL & MCCC2 & IL4I1 \\
    DEG\tablefootnote{VALINE LEUCINE AND ISOLEUCINE
DEGRADATION}
     & +0.40 & -0.35 & -0.30 & +0.29 & +0.25 & +0.24 & +0.23 & -0.23 & -0.23 & +0.22 \\
    \midrule
    P53 SIGNALING
     & CYCS & CASP9 & SIAH1 & TSC2$^{\ast}$ & CDK6$^{\ast}$ & CDK4$^{\ast}$ & PERP & CCNE1$^{\ast}$ & IGF1$^{\ast}$ & SERPINB5 \\
    PATHWAY
     & -0.33 & +0.33 & -0.32 & -0.30 & -0.25 & -0.23 & -0.22 & -0.21 & +0.21 & -0.21 \\
    \midrule
    PANCREATIC
     & PIK3CG$^{\ast}$ & STAT3$^{\ast}$ & RALGDS$^{\ast}$ & RAC2$^{\ast}$ & RAF1$^{\ast}$ & TGFBR1$^{\ast}$ & STAT1$^{\ast}$ & EGFR$^{\ast}$ & PLD1 & IKBKG \\
    CANCER & +0.36 & +0.29 & +0.28 & +0.28 & -0.26 & -0.24 & +0.24 & -0.24 & -0.20 & -0.20 \\
    \midrule
    \midrule
    \multirow{2}{*}{GLIOMA}
     & IGF1R$^{\ast}$ & CDKN1A$^{\ast}$ & IGF1$^{\ast}$ & CAMK2G & CCND1$^{\ast}$ & MDM2$^{\ast}$ & KRAS$^{\ast}$ & PDGFRB$^{\ast}$ & PDGFB$^{\ast}$ & RAF1$^{\ast}$ \\
     & -0.33 & -0.31 & -0.28 & +0.27 & -0.25 & -0.23 & -0.22 & +0.22 & -0.20 & +0.19 \\
    \midrule
    MISMATCH
     & MSH6$^{\ast}$ & RFC5 & EXO1 & POLD1$^{\ast}$ & RFC2 & RPA2 & SSBP1 & MLH3 & RFC1 & RPA3 \\
    REPAIR
     & +0.36 & +0.32 & +0.29 & +0.28 & +0.27 & +0.24 & +0.21 & +0.21 & +0.16 & +0.14 \\
    \bottomrule
    \end{tabular}
\end{table}
\renewcommand*{\thefootnote}{\arabic{footnote}}

\begin{figure*}
    \centering
    \subcaptionbox{TCGA (train) \label{fig:featuremap-tcga:sub:class}}{\includegraphics[width=.3\linewidth]{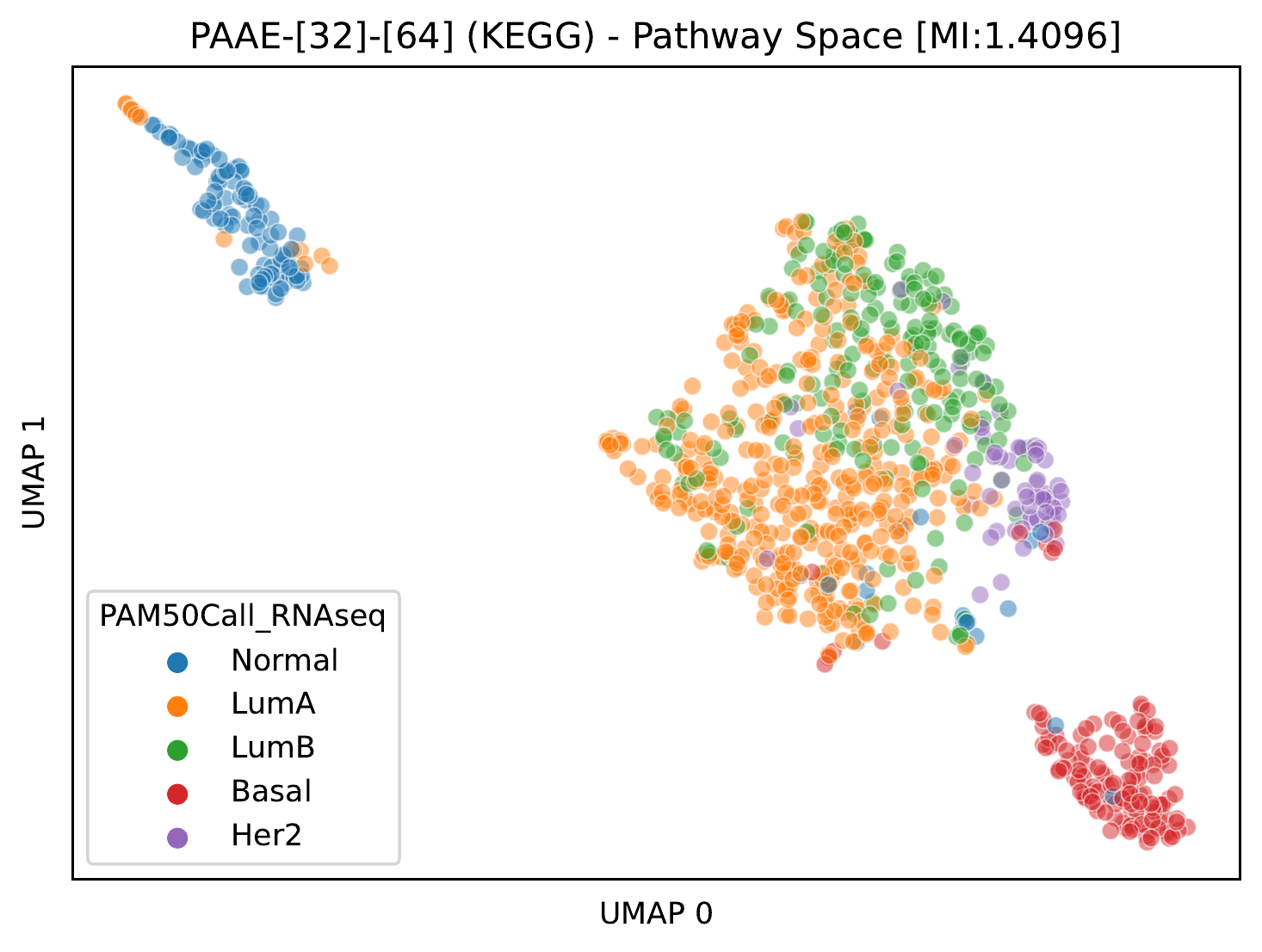}}
    \hfill
    \subcaptionbox{\tiny GLYCOSAMINOGLYCAN\_BIOSYNTHESIS\_KERATAN\_SULFATE  \label{fig:featuremap-tcga:sub:GLYCOSAMINOGLYCAN_BIOSYNTHESIS_KERATAN_SULFATE}}{\includegraphics[width=.3\linewidth]{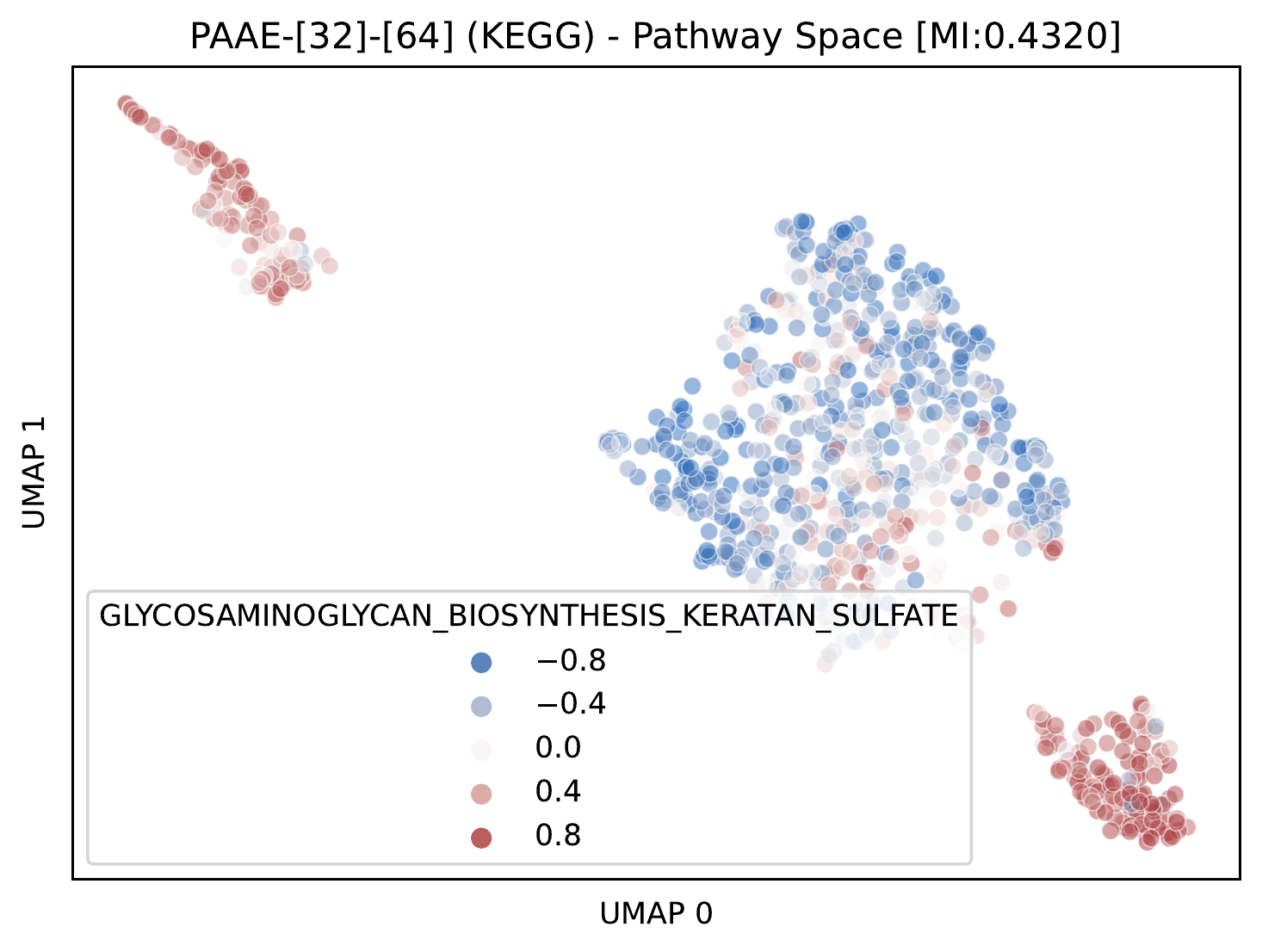}}
    \hfill
    \subcaptionbox{\tiny SPHINGOLIPID\_METABOLISM \label{fig:featuremap-tcga:sub:SPHINGOLIPID_METABOLISM}}{\includegraphics[width=.3\linewidth]{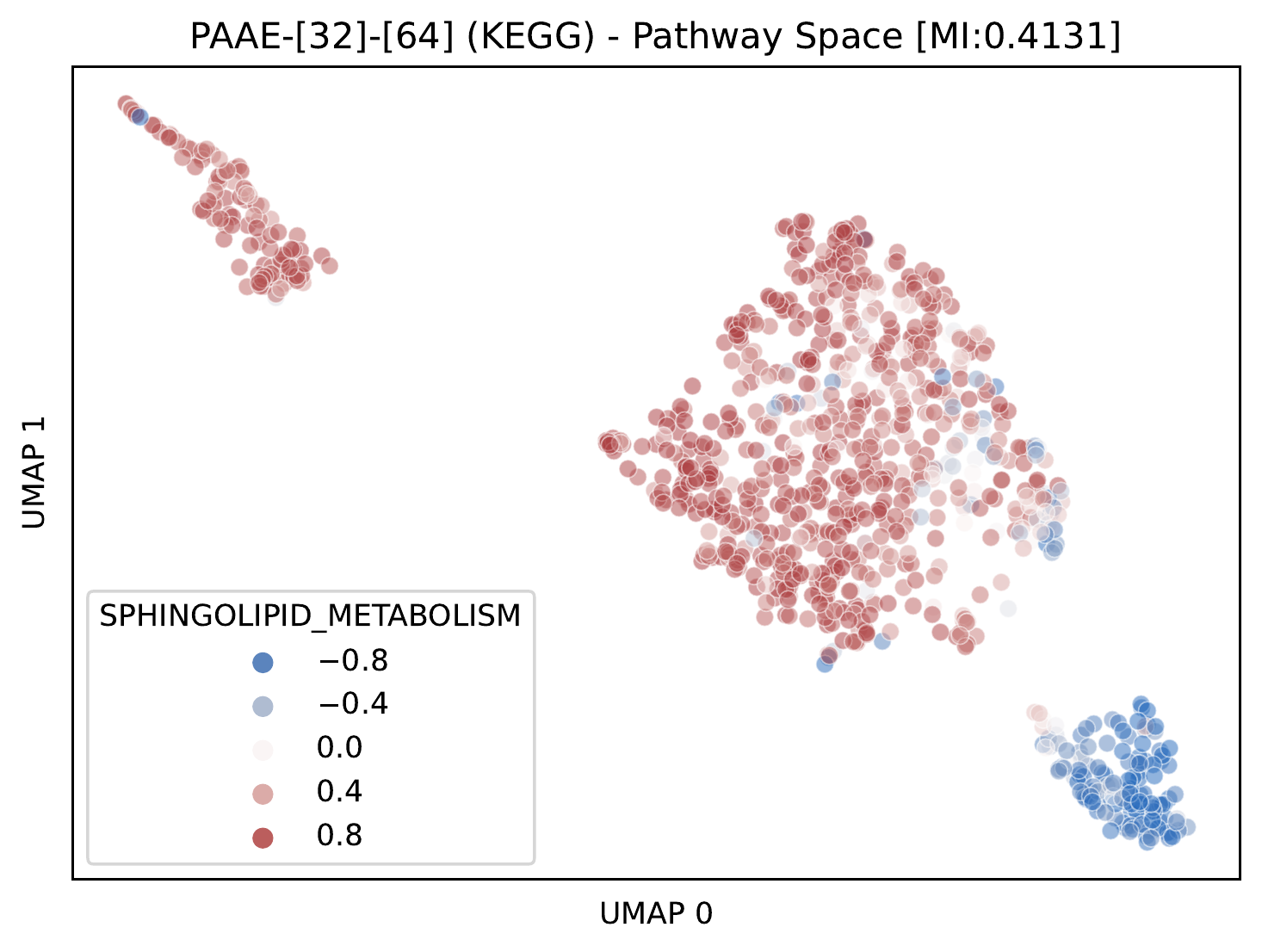}}
    \\
    \subcaptionbox{\tiny VALINE\_LEUCINE\_AND\_ISOLEUCINE\_DEGRADATION \label{fig:featuremap-tcga:sub:VALINE_LEUCINE_AND_ISOLEUCINE_DEGRADATION}}{\includegraphics[width=.3\linewidth]{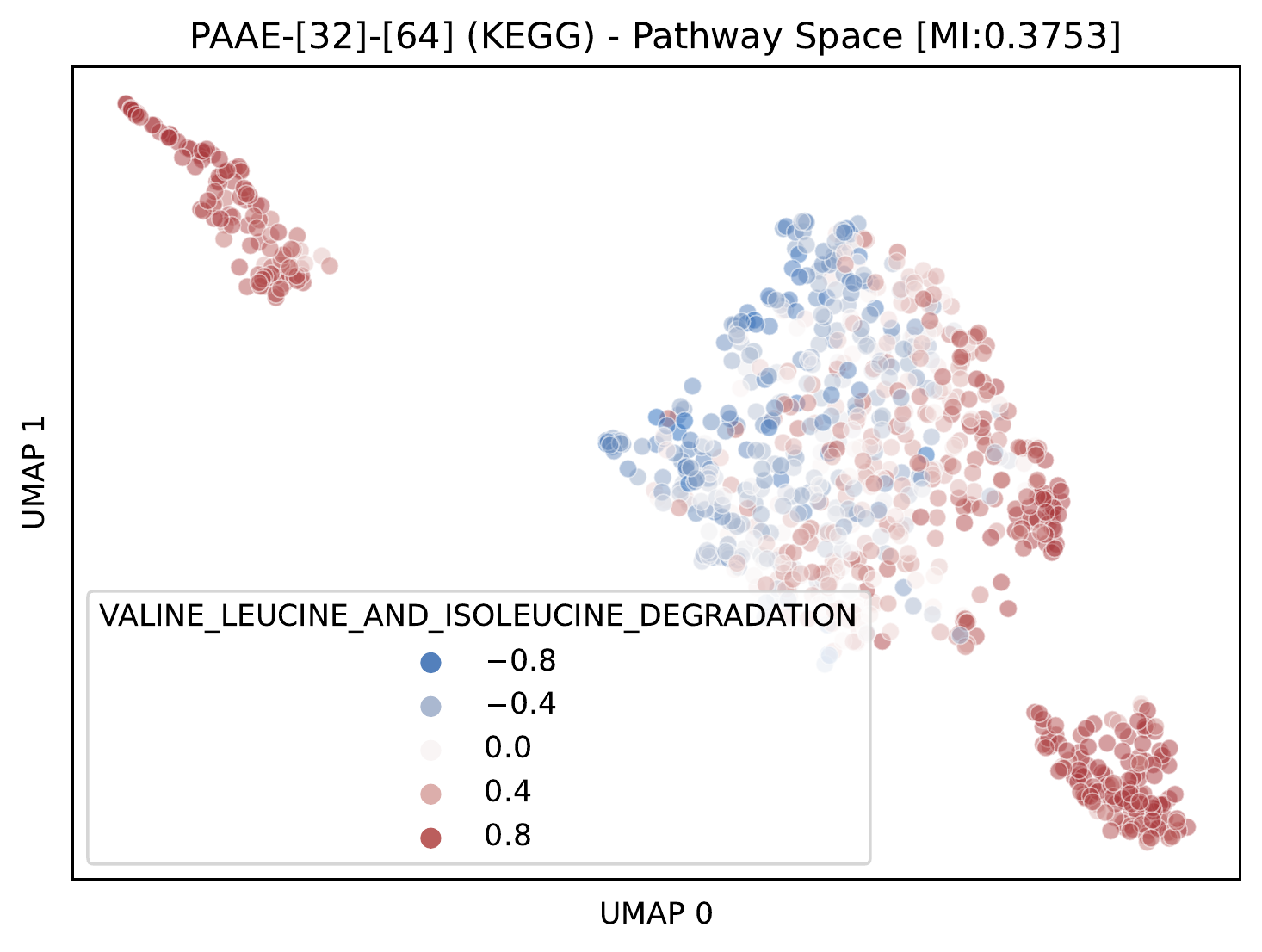}}
    \hfill
    \subcaptionbox{\tiny P53\_SIGNALING\_PATHWAY \label{fig:featuremap-tcga:sub:P53_SIGNALING_PATHWAY}}{\includegraphics[width=.3\linewidth]{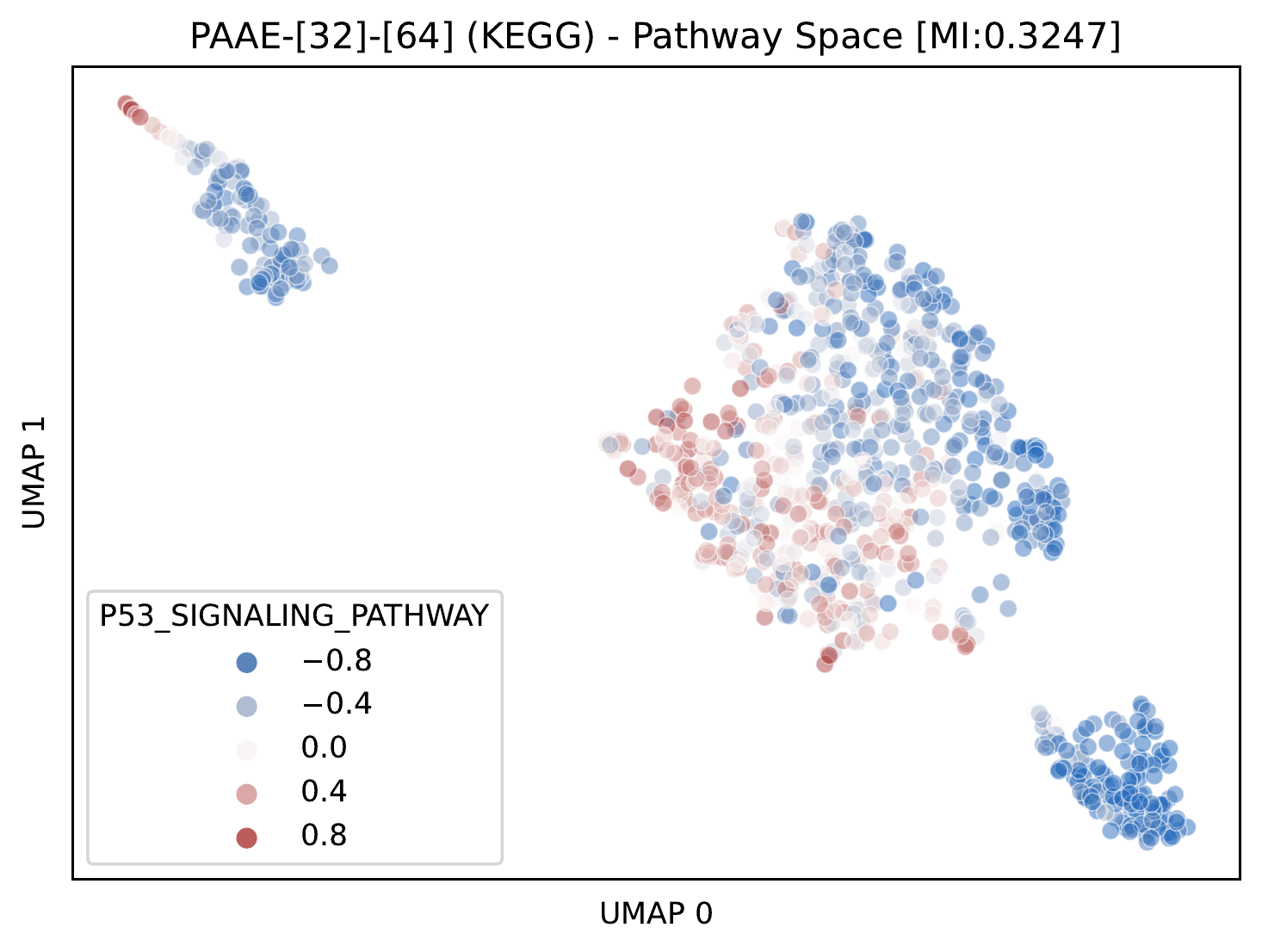}}
    \hfill
    \subcaptionbox{\tiny PANCREATIC\_CANCER \label{fig:featuremap-tcga:sub:PANCREATIC_CANCER}}{\includegraphics[width=.3\linewidth]{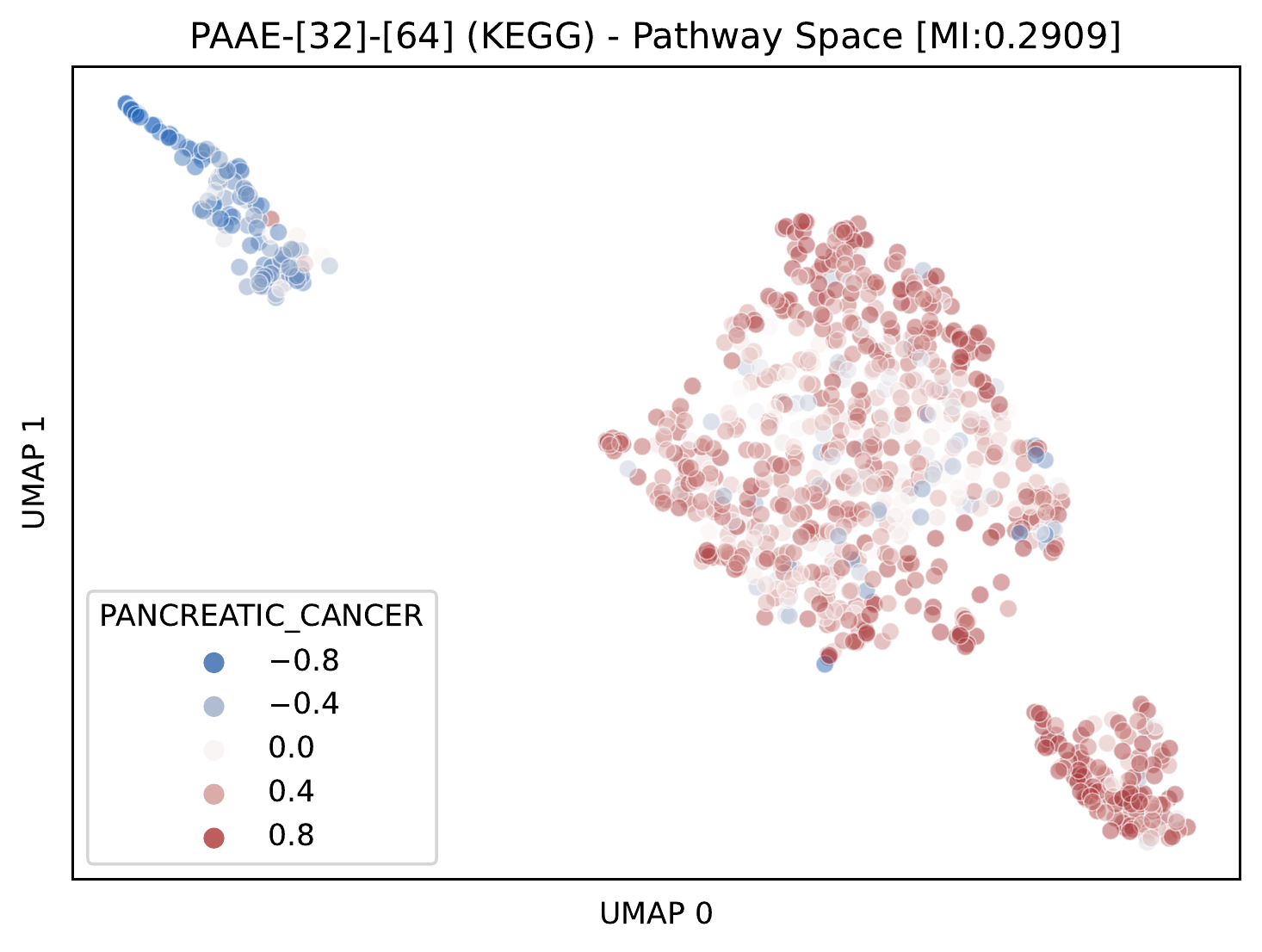}}
    \\
    \subcaptionbox{\tiny GLIOMA \label{fig:featuremap-tcga:sub:GLIOMA}}{\includegraphics[width=.3\linewidth]{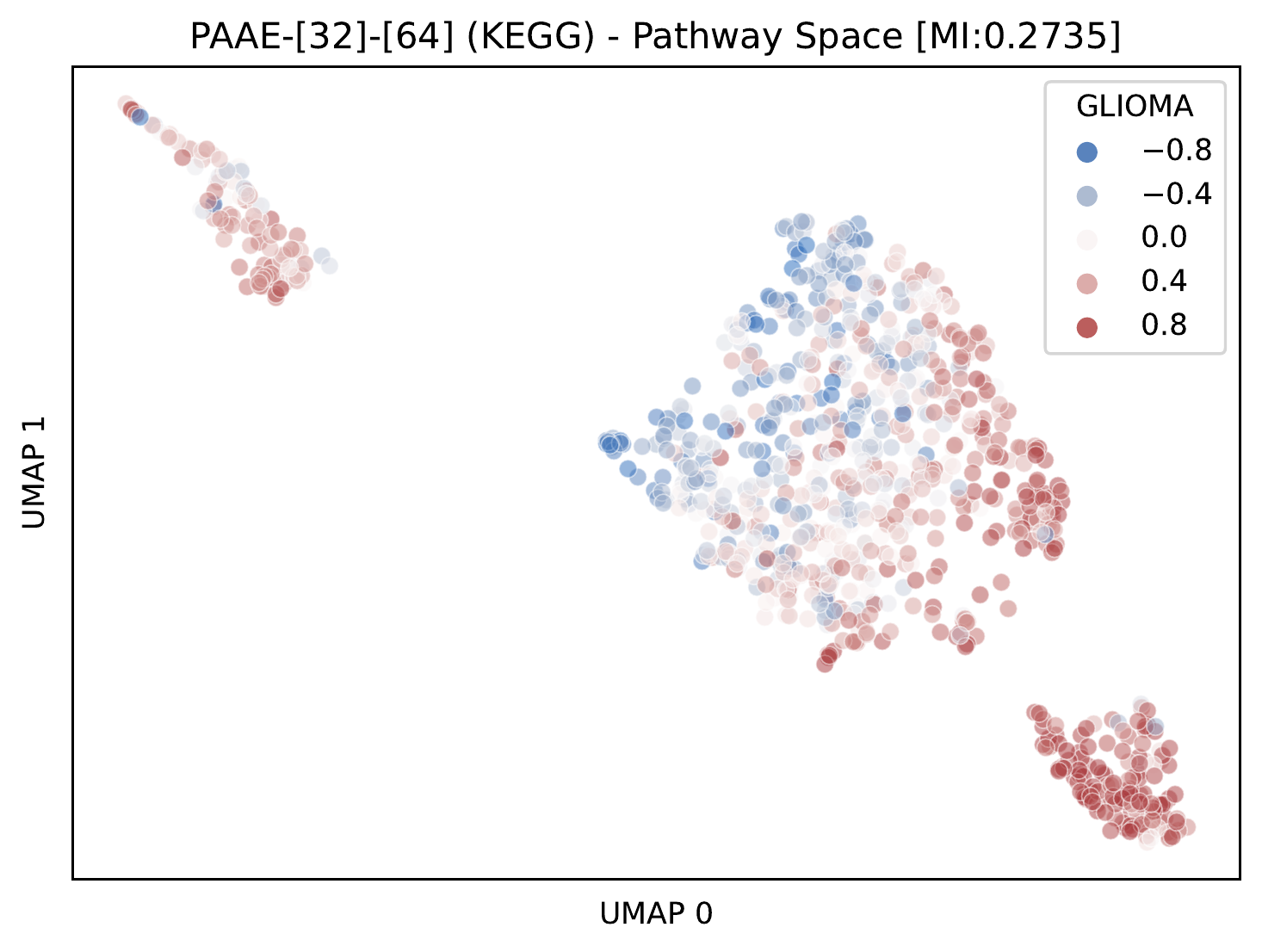}}
    \hfill
    \subcaptionbox{\tiny MISMATCH\_REPAIR \label{fig:featuremap-tcga:sub:MISMATCH_REPAIR}}{\includegraphics[width=.3\linewidth]{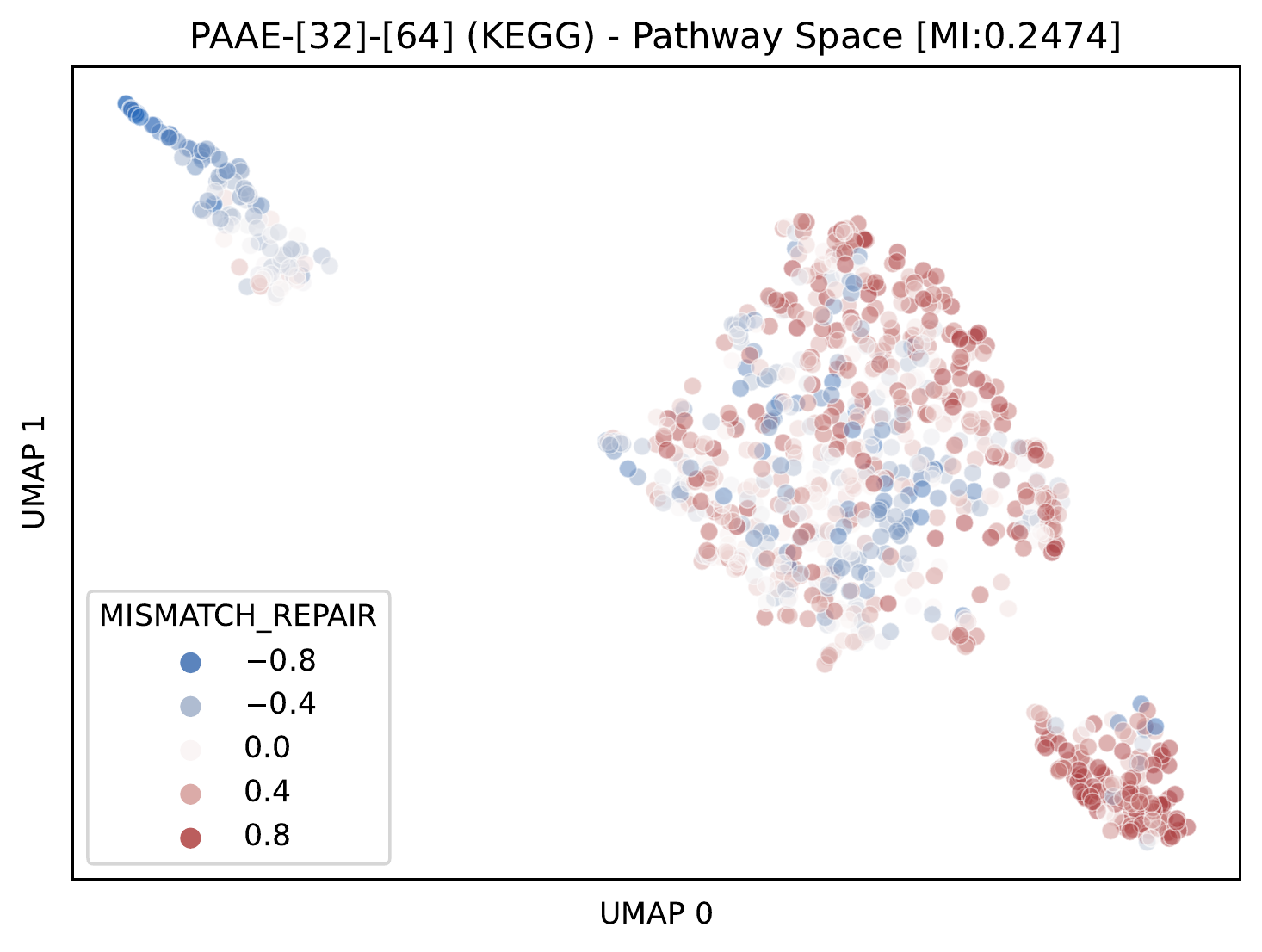}}
    
    \caption{The featuremap showing the intensity of each sample's inferred pathway activity vectors for KEGG PAAE's pathway activity space in the TCGA dataset, as well as the class distribution overlayed on top of the 2-dimensional UMAP reduction: {\color{brca-normal}Normal in blue}, {\color{brca-luma}Luminal A in orange}, {\color{brca-lumb}Luminal B in green}, {\color{brca-basal}Basal in red}, {\color{brca-her2}Her2 in purple}. We only show here pathways that are among the top 5 with highest the most mutual information w.r.t. the classes on either TCGA or Metabric.}
    \label{fig:featuremap-tcga}
\end{figure*}

\begin{figure*}
    \centering
    \subcaptionbox{Metabric (test) \label{fig:featuremap-meta:sub:class}}{\includegraphics[width=.3\linewidth]{figures/paae/interpretation/brca-Metabric-scatter-PAAE_32_64_KEGG__PathwaySpace-UMAP-wide-PAM50.pdf}}
    \hfill
    \subcaptionbox{\tiny GLYCOSAMINOGLYCAN\_BIOSYNTHESIS\_KERATAN\_SULFATE  \label{fig:featuremap-meta:sub:GLYCOSAMINOGLYCAN_BIOSYNTHESIS_KERATAN_SULFATE}}{\includegraphics[width=.3\linewidth]{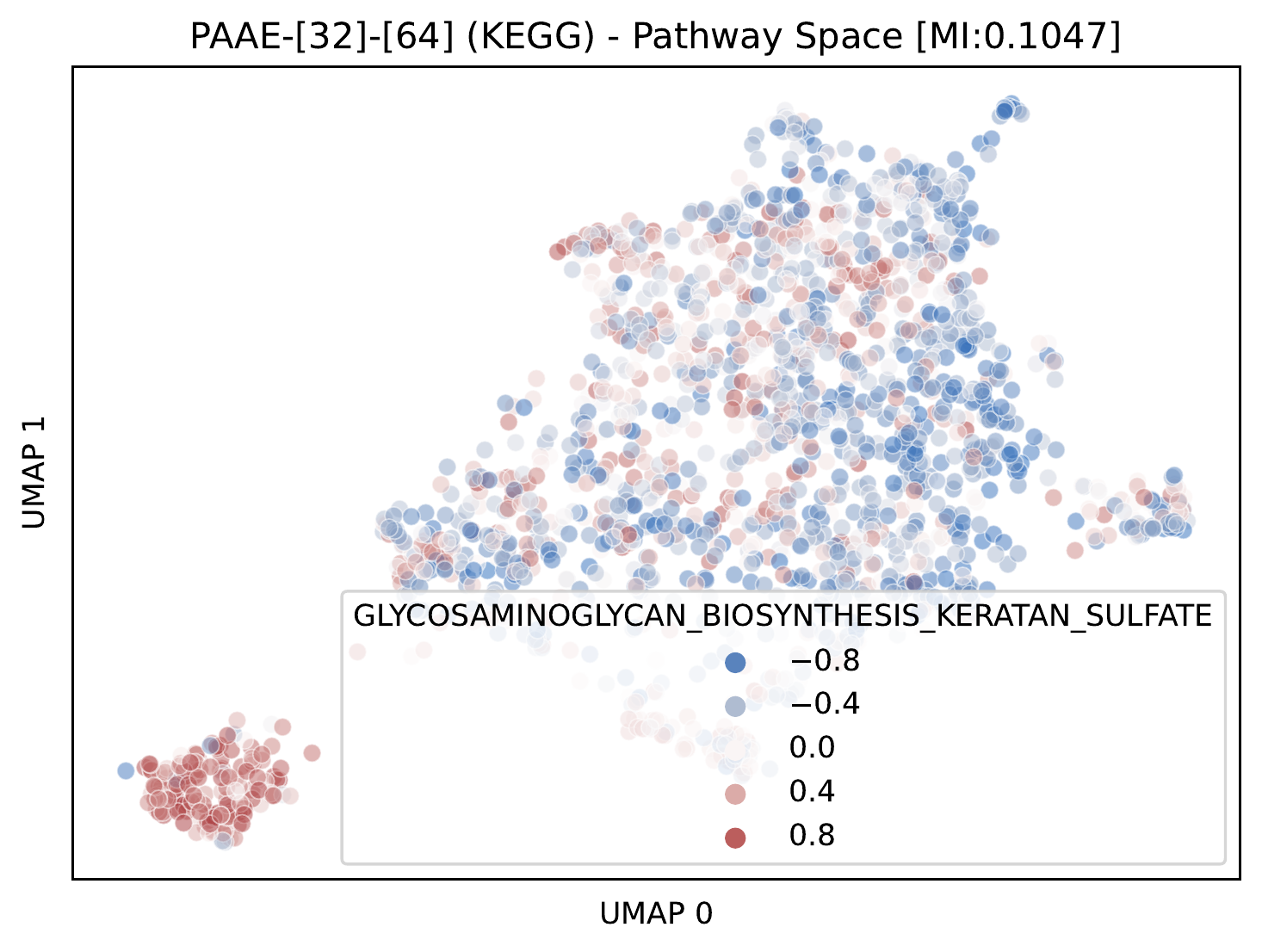}}
    \hfill
    \subcaptionbox{\tiny SPHINGOLIPID\_METABOLISM \label{fig:featuremap-meta:sub:SPHINGOLIPID_METABOLISM}}{\includegraphics[width=.3\linewidth]{figures/paae/interpretation/brca-Metabric-scatter-PAAE_32_64_KEGG__PathwaySpace-UMAP-wide-SPHINGOLIPID_METABOLISM.pdf}}
    \\
    \subcaptionbox{\tiny VALINE\_LEUCINE\_AND\_ISOLEUCINE\_DEGRADATION \label{fig:featuremap-meta:sub:VALINE_LEUCINE_AND_ISOLEUCINE_DEGRADATION}}{\includegraphics[width=.3\linewidth]{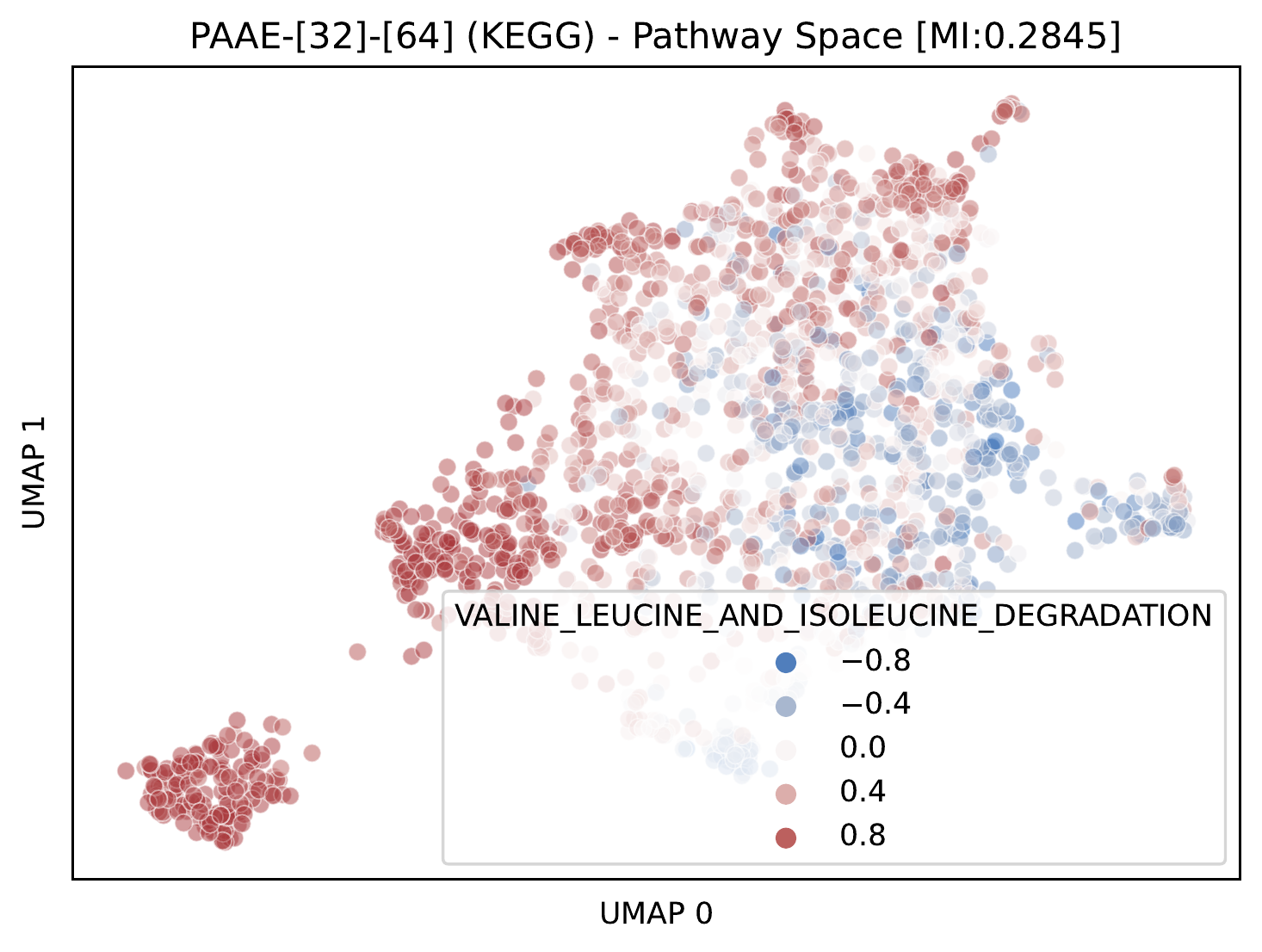}}
    \hfill
    \subcaptionbox{\tiny P53\_SIGNALING\_PATHWAY \label{fig:featuremap-meta:sub:P53_SIGNALING_PATHWAY}}{\includegraphics[width=.3\linewidth]{figures/paae/interpretation/brca-Metabric-scatter-PAAE_32_64_KEGG__PathwaySpace-UMAP-wide-P53_SIGNALING_PATHWAY.pdf}}
    \hfill
    \subcaptionbox{\tiny PANCREATIC\_CANCER \label{fig:featuremap-meta:sub:PANCREATIC_CANCER}}{\includegraphics[width=.3\linewidth]{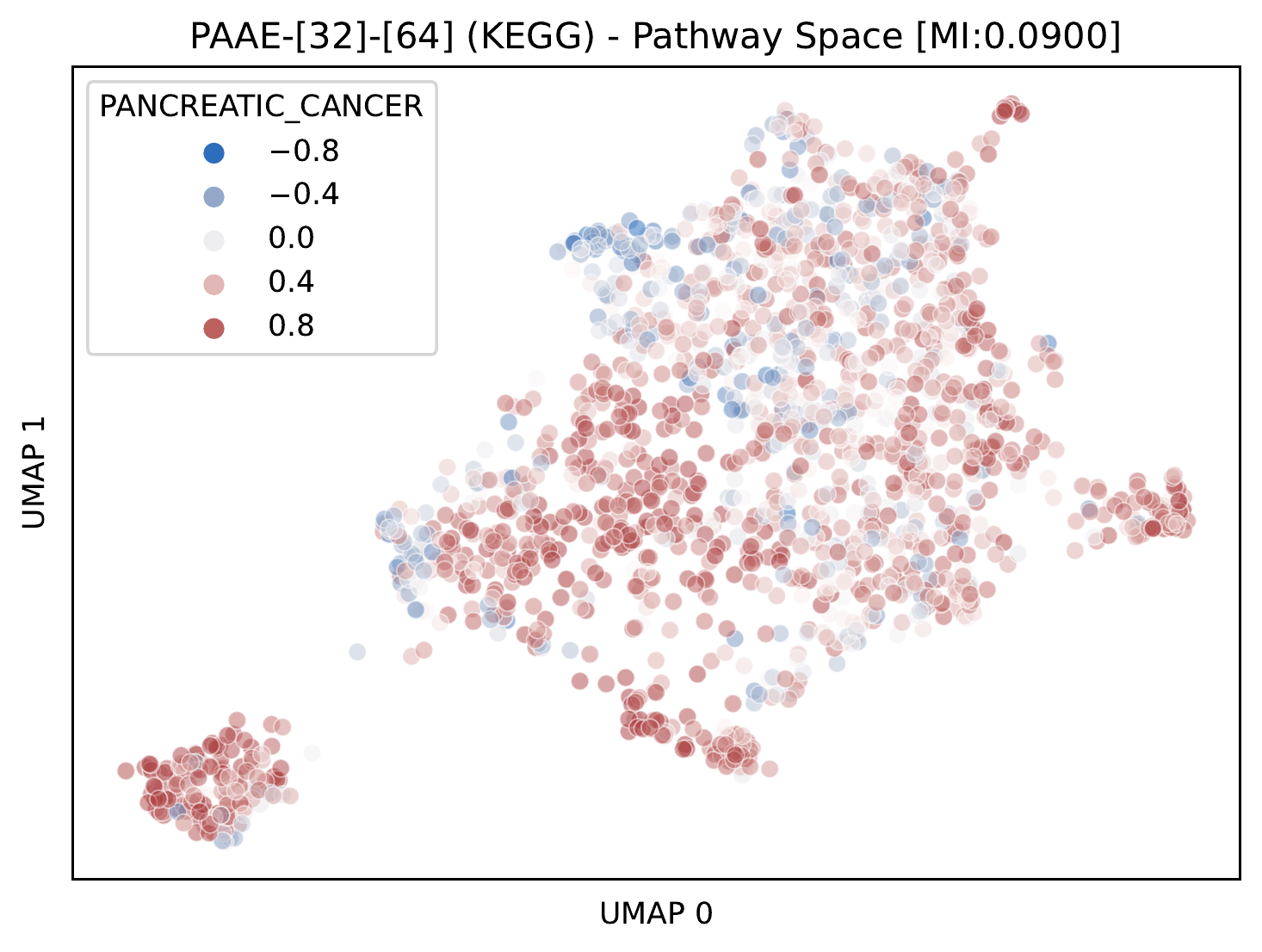}}
    \\
    \subcaptionbox{\tiny GLIOMA \label{fig:featuremap-meta:sub:GLIOMA}}{\includegraphics[width=.3\linewidth]{figures/paae/interpretation/brca-Metabric-scatter-PAAE_32_64_KEGG__PathwaySpace-UMAP-wide-GLIOMA.pdf}}
    \hfill
    \subcaptionbox{\tiny MISMATCH\_REPAIR \label{fig:featuremap-meta:sub:MISMATCH_REPAIR}}{\includegraphics[width=.3\linewidth]{figures/paae/interpretation/brca-Metabric-scatter-PAAE_32_64_KEGG__PathwaySpace-UMAP-wide-MISMATCH_REPAIR.pdf}}
    
    \caption{The featuremap showing the intensity of each sample's inferred pathway activity vectors for KEGG PAAE's pathway activity space in the Metabric dataset, as well as the class distribution overlayed on top of the 2-dimensional UMAP reduction: {\color{brca-lumb}Normal in green}, {\color{brca-normal}Luminal A in blue}, {\color{brca-luma}Luminal B in orange}, {\color{brca-her2}Basal in purple}, {\color{brca-basal}Her2 in red}. We only show here pathways that are among the top 5 with highest the most mutual information w.r.t. the classes on either TCGA or Metabric.}
    \label{fig:featuremap-meta}
\end{figure*}

Furthermore, we can see in Fig.~\ref{fig:clustermap-kegg-importantgenes} that some of these gene combinations already provide good enough clustering separation, and even if one looks at the feature maps in Figs.~\ref{fig:featuremap-meta} and \ref{fig:featuremap-tcga}, some pathway activities already separate enough of the classes even on the training set. 

As a thought-experiment, suppose that someone would use only the 5 strongest pathway activity score predictors in the TCGA dataset as shown in Fig.~\ref{fig:featuremap-tcga}, to manually build a decision tree to classify samples:

\begin{enumerate}[i]
    \item~\label{kegg:tree:normal} Considering first only the \GBKS{} (GBKS) and \SM{} (SM) pathways, if we take values that are positive on both we'd get a decent predictor for the Normal class;
    \item~\label{kegg:tree:basal} if we take high GBKS and low SM values we have the same for the Basal class;
    \item~\label{kegg:tree:else} Anything else can then be considered to be in one of the other three classes:
    \begin{enumerate}[a]
        \item\label{kegg:tree:else:luma} With the rest, we can then see that positive \PSP{} (P53) catch some of the LumA samples;
        \item\label{kegg:tree:else:lumb} Whereas low-positive, but not strongly-positive, \VLID{} (VLID) combined with positive \PC{} (PC) marks some LumB samples;
        \item\label{kegg:tree:else:her2} Finally, strongly-positive VLID marks some Her2 samples, although it wrongly encompasses some LumB samples as well.
    \end{enumerate}
\end{enumerate}
 
Applying this manually-constructed decision tree model on the test dataset, then, we have that: \ref{kegg:tree:normal} correctly classifies some the Normal class, although it seems to also include some LumA samples; \ref{kegg:tree:basal} seems to be mostly correct; \ref{kegg:tree:else:luma} also seem to correctly classify some LumA samples, even catching some of the samples that are more separated from the main LumA body; while \ref{kegg:tree:else:lumb} does capture some LumB samples, but it seems to be one of the weakest leaves; and, finally, although \ref{kegg:tree:else:her2} seem to correctly classify some Her2 samples it misclassifies many others together.

We then see that our manually-constructed tree, that only uses unsupervised learning techniques up to the tree construction, seem to mostly work on 3 out of the 5 final leaves, despite being tested on a completely different cohort with a different technology altogether.

\begin{figure*}
    \centering
    \centering
    \subcaptionbox{\label{fig:survival-kegg-importantgenes:sm:degs2}}{\includegraphics[width=.45\linewidth]{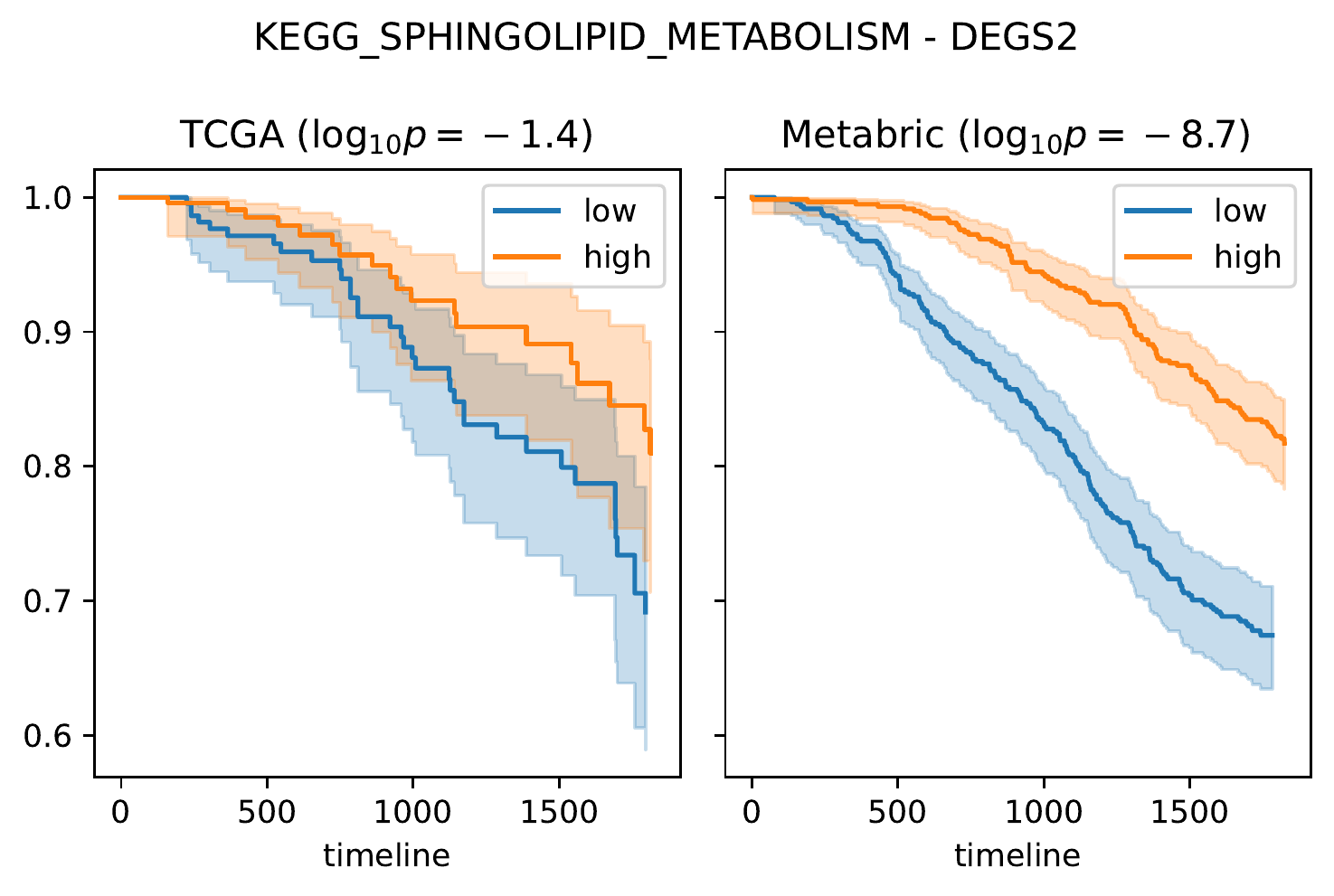}}
    \hfill
    \subcaptionbox{\label{fig:survival-kegg-importantgenes:vlid:mccc1}}{\includegraphics[width=.45\linewidth]{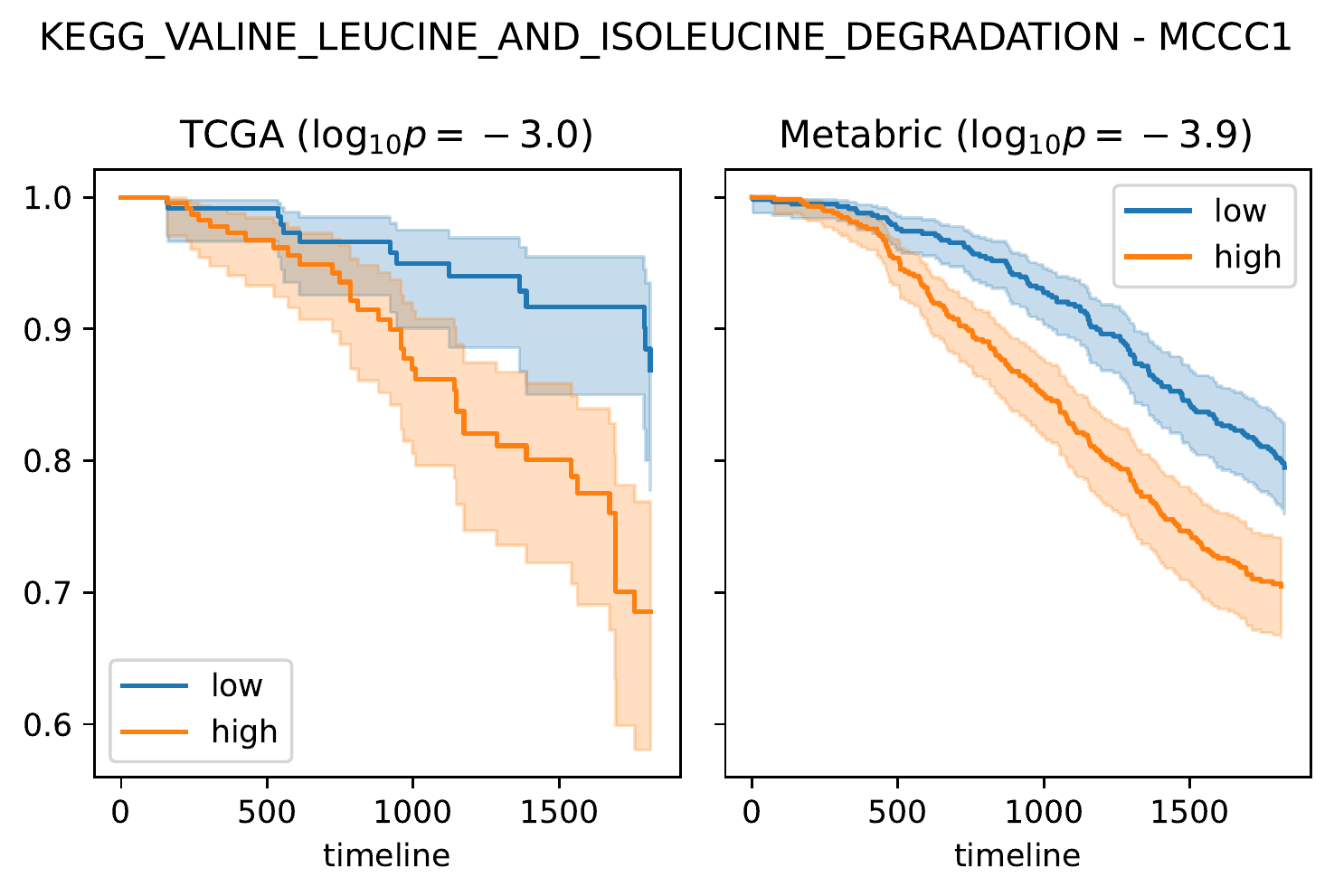}}\\
    \subcaptionbox{\label{fig:survival-kegg-importantgenes:vlid:oxct1}}{\includegraphics[width=.45\linewidth]{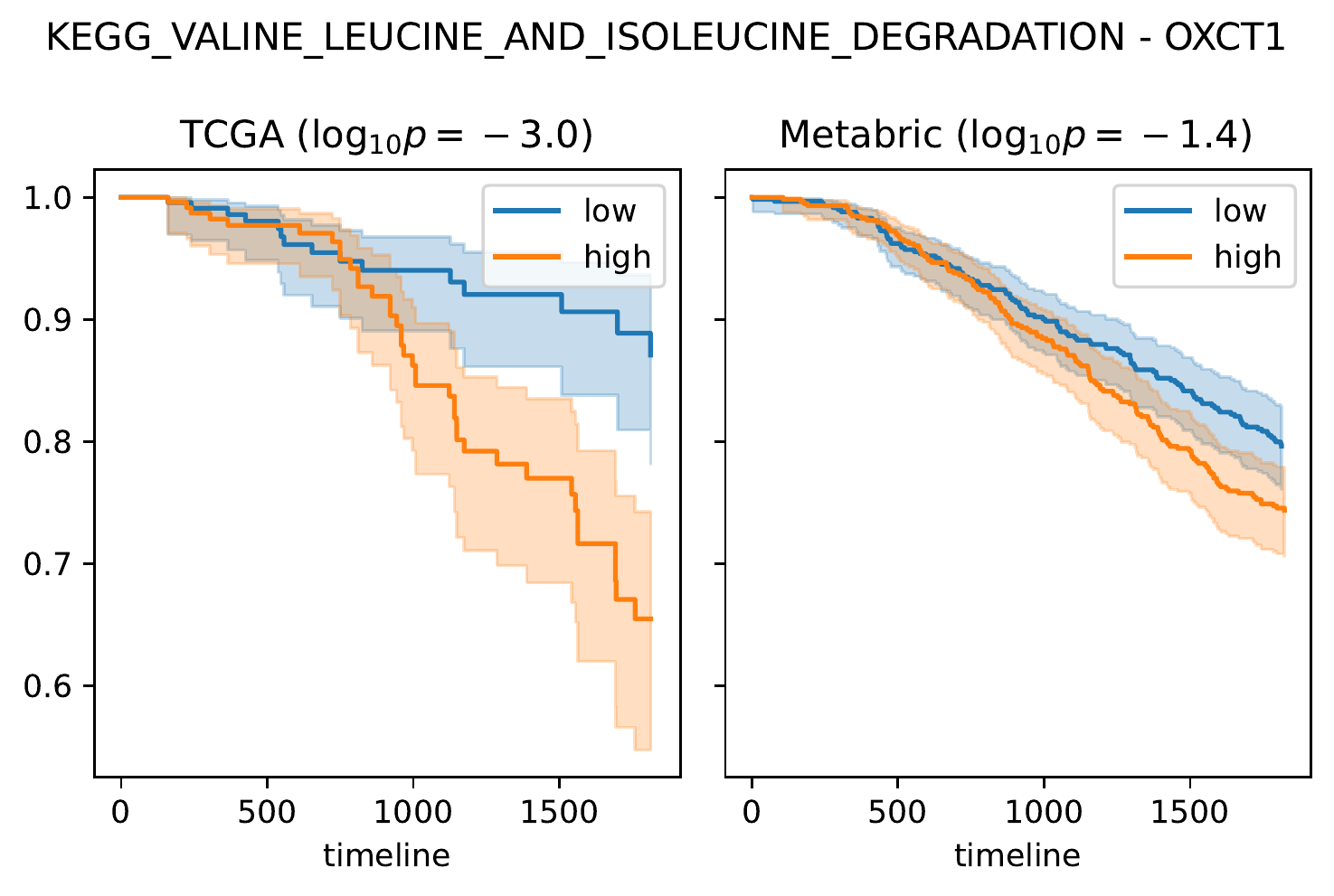}}
    \hfill
    \subcaptionbox{\label{fig:survival-kegg-importantgenes:p53:perp}}{\includegraphics[width=.45\linewidth]{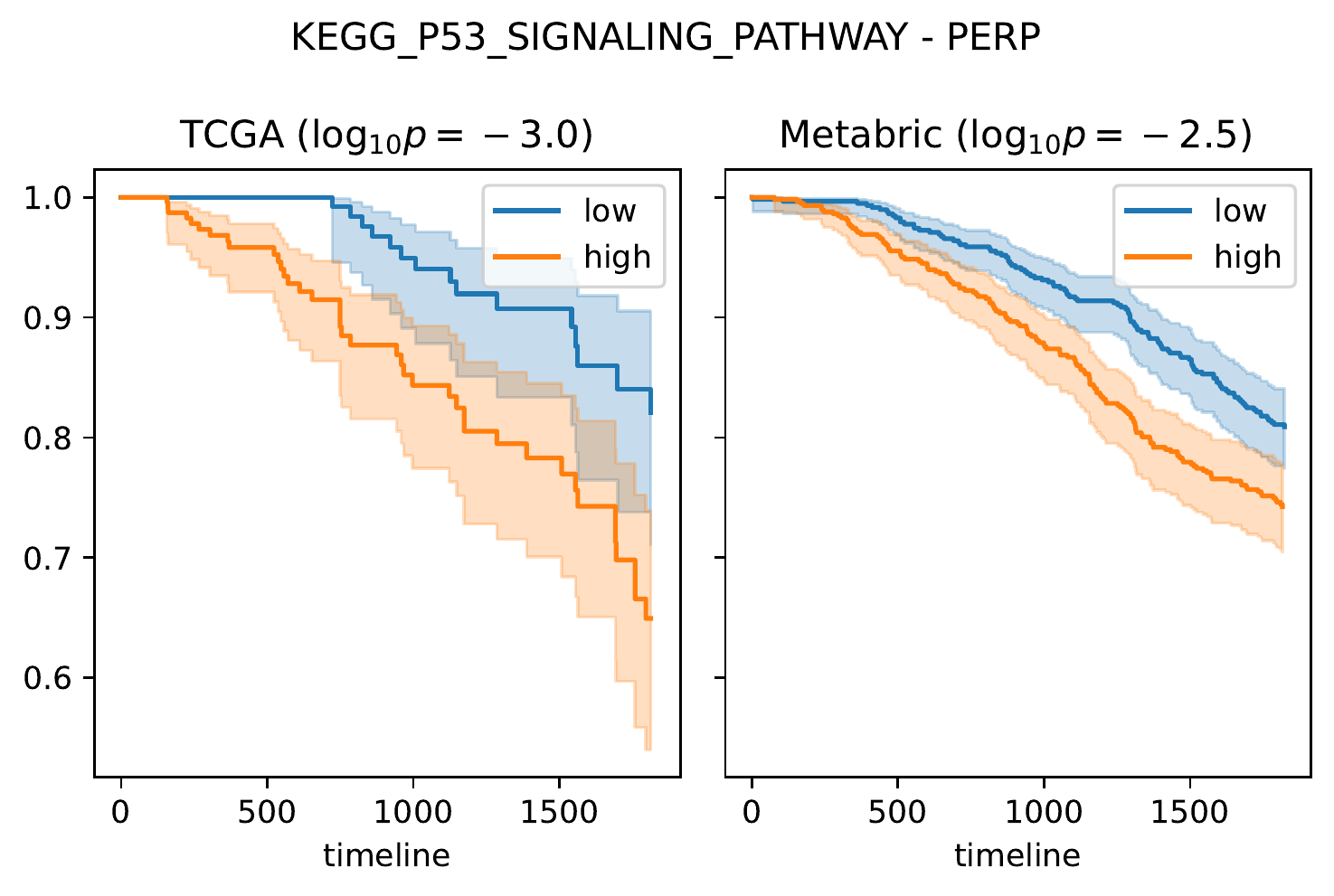}}
    
    \caption{Kaplan-Meier Curves for the gene that had a significant p-value on a logrank test between the upper and lower thirds in expression, on both the TCGA and Metabric datasets, and matched low-high survival sign. These 4 genes are $57.14\%$ of the 7 genes that were significant in the logrank test for the TCGA dataset, considering the 10 most important genes from each of the 5 pathways with highest MI w.r.t. the classification target in TCGA.}
    \label{fig:survival-kegg-importantgenes}
\end{figure*}

We show in Fig.~\ref{fig:survival-kegg-importantgenes} the Kaplan-Meier curves for the upper and lower thirds percentiles of the expressed TPM/IPM values along with the p-values for logrank separation tests. We show the gene that had both significant logrank separation as well as a matching low-high survival sign on the cutoff date.

\subsubsection{Hallmark Genes}

For a neural network we produced when training with the hallmark genes pathway set, we found that the 5 pathways with most mutual information w.r.t. the PAM50 subtype in the TCGA dataset were P53\_PATHWAY, UV\_RESPONSE\_DN, CHOLESTEROL\_HOMEOSTASIS, APICAL\_SURFACE, and MTORC1\_SIGNALING.

For each of these pathways, we calculated the 10 genes with the highest ANPW, which can be seen in Tab.~\ref{tab:important-genes-hallmark}, where $^{\ast}$ indicates that the gene was a match on OncoKB \cite{chakravarty_oncokb_2017}.

\begin{figure*}
    \centering
    \subcaptionbox{TCGA (train)\label{fig:clustermap-hallmark-importantgenes:sub:tcga}}{\includegraphics[width=.49\linewidth]{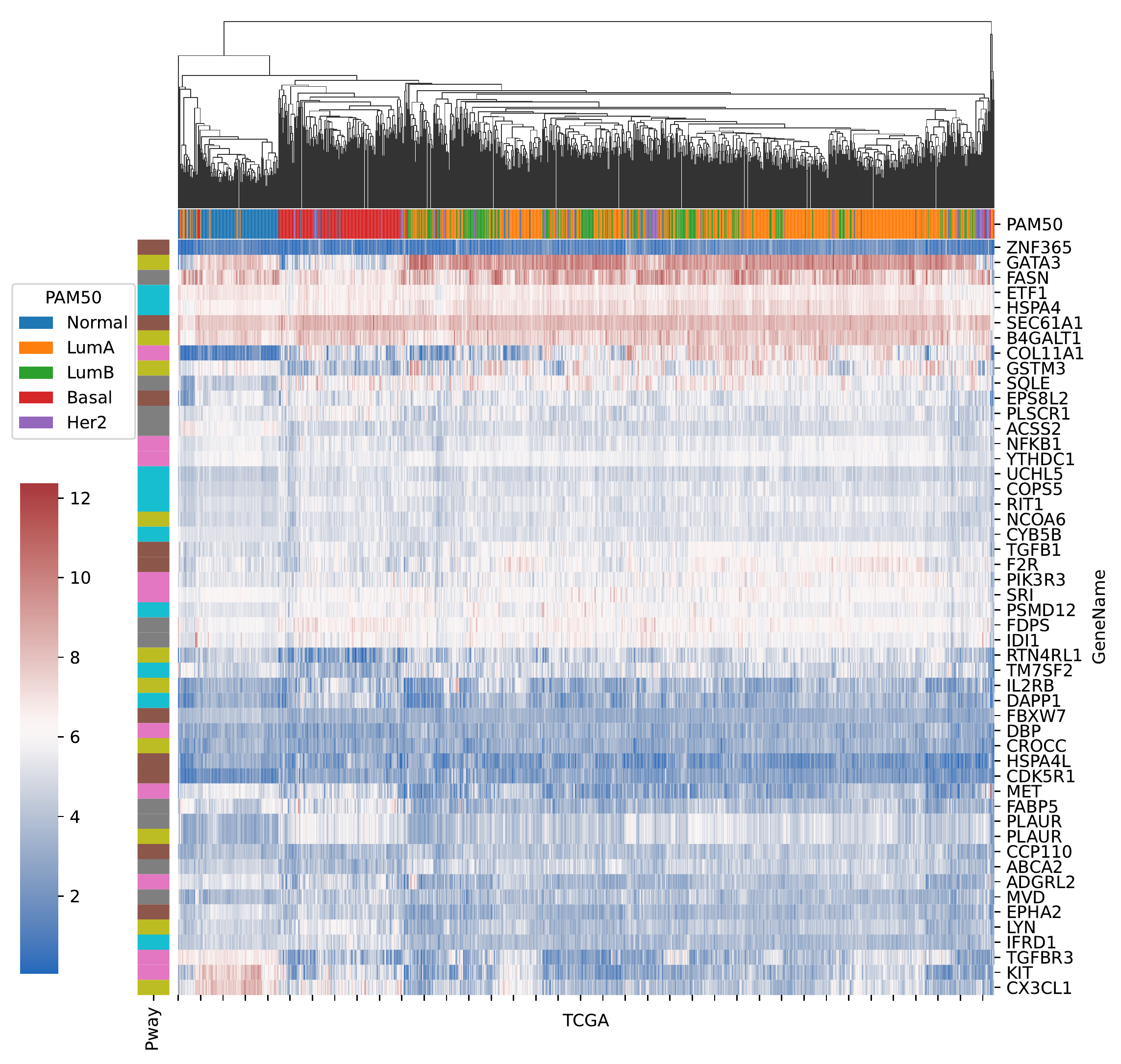}}
    \hfill
    \subcaptionbox{Metabric (test)\label{fig:clustermap-hallmark-importantgenes:sub:meta}}{\includegraphics[width=.49\linewidth]{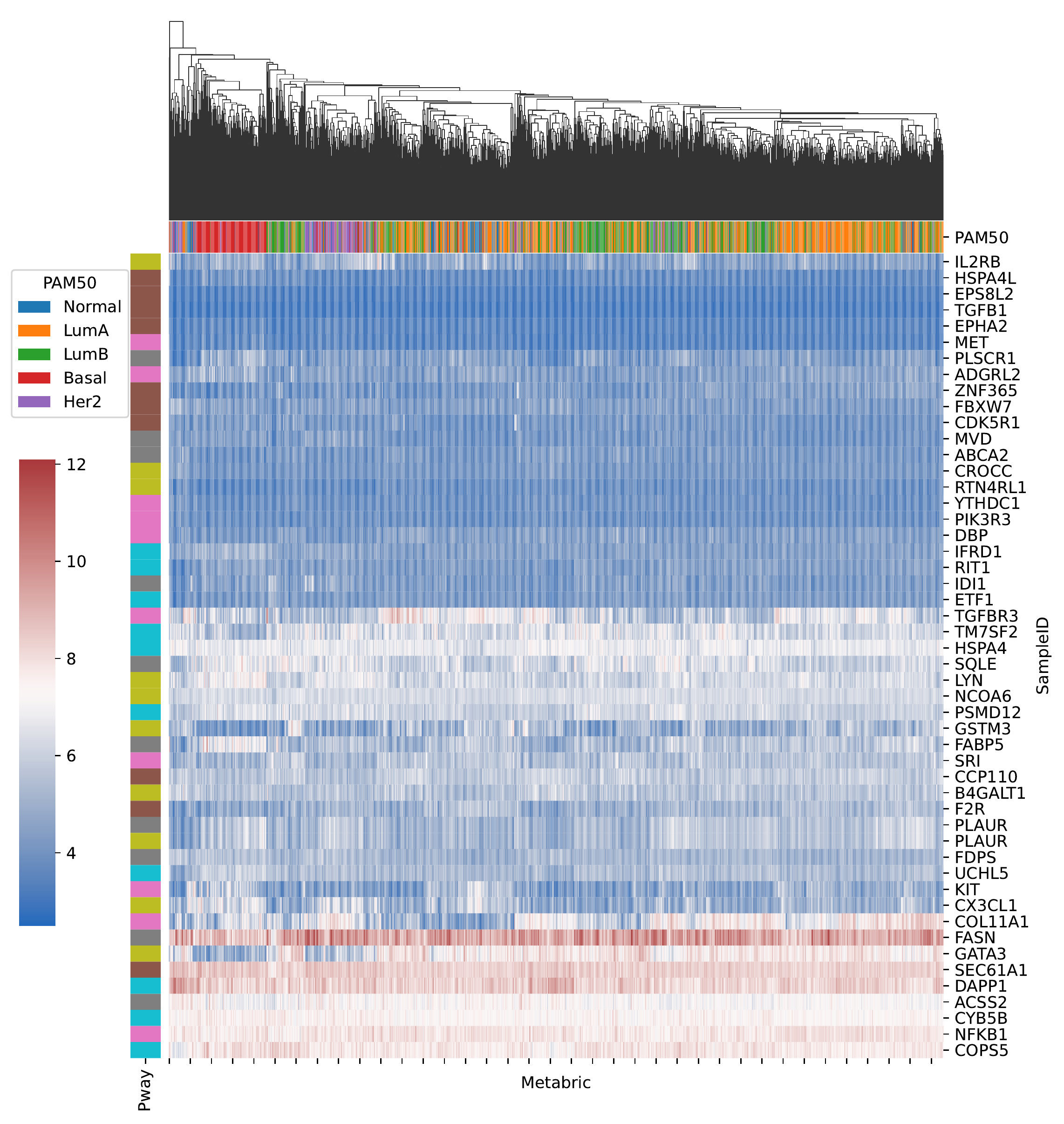}}
    \caption{The clustermap using the euclidean distance between samples' log2p1 TPM/IPM gene expression vectors for Hallmark Genes PAAE's pathway activity space with the colours marking BRCA's 5-classes: {\color{brca-normal}Normal in blue}, {\color{brca-luma}Luminal A in orange}, {\color{brca-lumb}Luminal B in green}, {\color{brca-basal}Basal in red}, {\color{brca-her2}Her2 in purple}.}
    \label{fig:clustermap-hallmark-importantgenes}
\end{figure*}

Plotting the clustermaps of these 50 features in Fig.~\ref{fig:clustermap-hallmark-importantgenes}, we can see that there is good separation of almost all classes on both datasets, with the Normal class samples, again, not being as separated in the test Metabric dataset as it is on the training TCGA dataset.

\begin{table}[]
    \centering
    \scriptsize
    \caption{Top 10 Genes w.r.t ANPW for the 5 pathways with highest MI w.r.t. the classification target for the Hallmark Genes pathway set.}
    \label{tab:important-genes-hallmark}
    \begin{tabular}{ccccccccccc}
    \toprule
    \multirow{2}{*}{PATHWAY} & \multicolumn{10}{c}{Gene names} \\
    & \multicolumn{10}{c}{Neural Path Weights} \\
    \midrule
    P53
     & EPHA2 & EPS8L2 & TGFB1 & FBXW7$^{\ast}$ & SEC61A1 & CCP110 & F2R & HSPA4L & CDK5R1 & ZNF365 \\
    PATHWAY
     & +0.31 & +0.30 & -0.26 & +0.25 & -0.24 & -0.23 & -0.22 & +0.22 & -0.22 & -0.21 \\
    \midrule
    UV RESPONSE
     & KIT$^{\ast}$ & COL11A1 & ADGRL2 & TGFBR3 & NFKB1 & YTHDC1 & PIK3R3$^{\ast}$ & DBP & MET$^{\ast}$ & SRI \\
    DN
     & -0.39 & +0.38 & -0.32 & -0.27 & -0.23 & +0.22 & -0.21 & -0.21 & -0.20 & -0.20 \\
    \midrule
    CHOLESTEROL
     & PLAUR & FDPS & FABP5 & ABCA2 & MVD & PLSCR1 & FASN & IDI1 & SQLE & ACSS2 \\
    HOMEOSTASIS
     & +0.48 & +0.36 & +0.35 & -0.34 & +0.32 & +0.29 & -0.27 & -0.26 & -0.26 & -0.22 \\
    \midrule
    APICAL
     & LYN$^{\ast}$ & CROCC & IL2RB & GATA3$^{\ast}$ & GSTM3 & PLAUR & CX3CL1 & RTN4RL1 & NCOA6 & B4GALT1 \\
    SURFACE
     & -0.49 & +0.45 & -0.41 & +0.40 & +0.39 & +0.33 & -0.32 & +0.28 & +0.25 & -0.25 \\
    \midrule
    MTORC1
     & CYB5B & ETF1 & UCHL5 & DAPP1 & RIT1$^{\ast}$ & PSMD12 & HSPA4 & IFRD1 & TM7SF2 & COPS5 \\
    SIGNALING
     & +0.36 & -0.33 & +0.25 & -0.22 & +0.22 & +0.22 & -0.22 & +0.21 & -0.20 & +0.20 \\
    \bottomrule
    \end{tabular}
\end{table}

\begin{figure*}
    \centering
    \subcaptionbox{\label{fig:survival-hallmark-importantgenes:apical:gata3}}{\includegraphics[width=.45\linewidth]{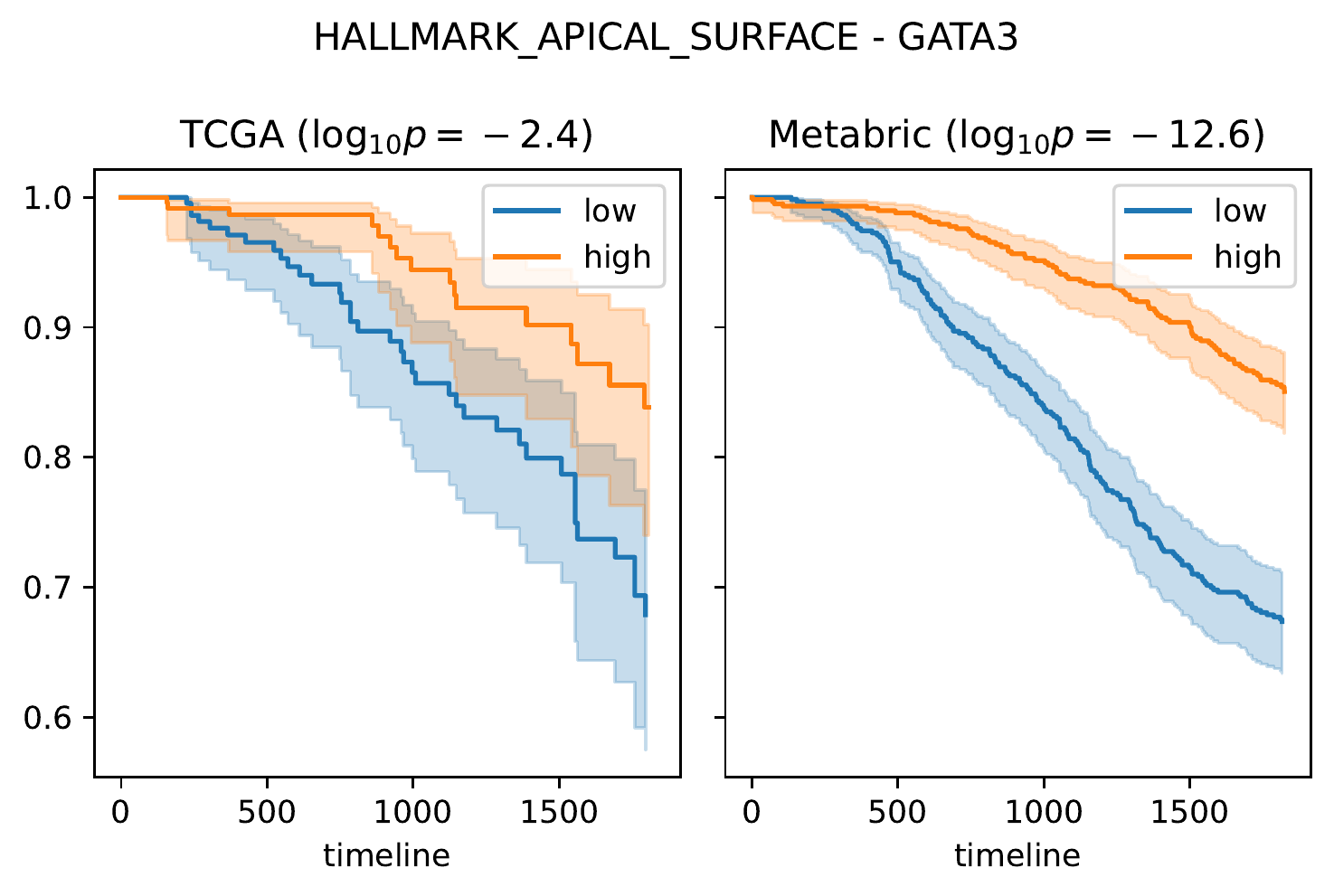}}
    \hfill
    \subcaptionbox{\label{fig:survival-hallmark-importantgenes:mtorc1:ifrd1}}{\includegraphics[width=.45\linewidth]{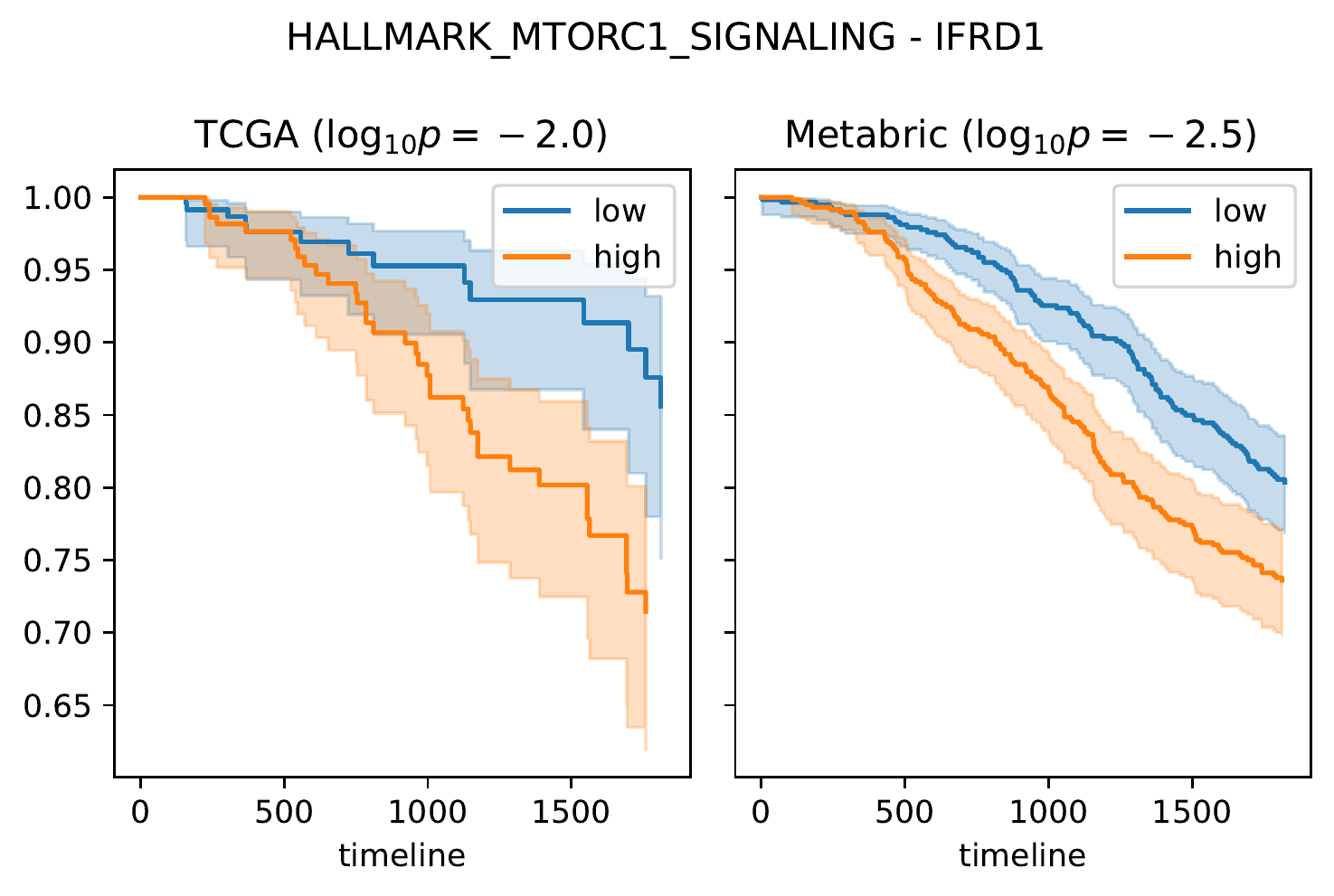}}
    
    \caption{Kaplan-Meier Curves for the 2 genes that had a significant p-value on a logrank test between the upper and lower thirds in expression, on both the TCGA and Metabric datasets, and matched low-high survival sign. These 2 genes are $28.57\%$ of the 7 genes that were significant in the logrank test for the TCGA dataset, considering the 10 most important genes from each of the 5 pathways with highest MI w.r.t. the classification target in TCGA.}
    \label{fig:survival-hallmark-importantgenes}
\end{figure*}

For the Hallmark Genes pathway set we show in Fig.~\ref{fig:survival-hallmark-importantgenes} the Kaplan-Meier curves for the upper and lower thirds percentiles of the expressed TPM/IPM values along with the p-values for logrank separation tests. We show the 2 genes that had both significant logrank separation as well as a matching low-high survival sign on the cutoff date.

\end{document}